\documentclass[lettersize,journal]{IEEEtran}
\ifCLASSOPTIONcompsoc
  \usepackage[nocompress]{cite}
\else
  \usepackage{cite}
\fi

\usepackage{amsmath,amsfonts,amssymb}
\usepackage{algorithmic}
\usepackage{algorithm}
\usepackage{array}
\usepackage[caption=false,font=normalsize,labelfont=sf,textfont=sf]{subfig}
\usepackage{textcomp}
\usepackage{stfloats}
\usepackage{url}
\usepackage{verbatim}
\usepackage{graphicx}
\usepackage{cite}
\usepackage{ulem}

\usepackage{multirow}
\usepackage{booktabs}
\usepackage{pifont}
\usepackage{xcolor} 
\usepackage[breaklinks=true,colorlinks,citecolor=blue,urlcolor=blue,linkcolor=blue,bookmarks=false]{hyperref}
\usepackage{cleveref}
\usepackage{threeparttable}
\usepackage[numbers,sort,compress]{natbib}
\usepackage{pifont}

\newcommand{\etal}{\textit{et al.}}

\newcommand{\xh}[1]{{\textcolor{black}{#1}}}

\newcommand{\rev}[1]{{\textcolor{black}{#1}}}

\usepackage{orcidlink}

\begin{document}

\title{Semantic Correspondence: \\
Unified Benchmarking and a Strong Baseline}

\author{Kaiyan Zhang \orcidlink{0009-0008-3359-4649},
        Xinghui Li \orcidlink{0000-0003-3797-5082},
        Jingyi Lu \orcidlink{0009-0005-1373-231X},
        Kai Han \orcidlink{0000-0002-7995-9999}
        \IEEEcompsocitemizethanks{
        \IEEEcompsocthanksitem Kaiyan Zhang, Jingyi Lu, and Kai Han are with the University of Hong Kong, Hong Kong (e-mail: kaihanx@hku.hk).
        \IEEEcompsocthanksitem Xinghui Li is with the Department of Engineering, University of Oxford, OX1 3PJ Oxford, U.K.
        \IEEEcompsocthanksitem Corresponding author: Kai Han (kaihanx@hku.hk)}}
        % <-this % stops a space

% The paper headers
\markboth{IEEE TRANSACTIONS ON PATTERN ANALYSIS AND MACHINE INTELLIGENCE, VOL. 48, NO. 3, MARCH 2026}
{Zhang \MakeLowercase{\textit{et al.}}: Semantic Correspondence: Unified Benchmarking and a Strong Baseline}

\maketitle

\IEEEtitleabstractindextext{%
\begin{abstract}
Establishing semantic correspondence is a challenging task in computer vision, aiming to match keypoints with the same semantic information across different images. 
Benefiting from the rapid development of deep learning, remarkable progress has been made over the past decade.
However, a comprehensive review and analysis of this task remains absent.
In this paper, we present the first extensive survey of semantic correspondence methods.
We first propose a taxonomy to classify existing methods based on the type of their method designs.
These methods are then categorized accordingly, and we provide a detailed analysis of each approach.
Furthermore, we aggregate and summarize the results of methods in the literature across various benchmarks into a unified comparative table, with detailed configurations to highlight performance variations.
% To further advance the field, we conduct an extensive ablation study of semantic correspondence models, evaluating the effectiveness of commonly used architectures at each stage. 
Additionally, to provide a detailed understanding of existing methods for semantic matching, we thoroughly conduct controlled experiments to analyze the effectiveness of the components of different methods. 
Finally, we propose a simple yet effective baseline that achieves state-of-the-art performance on multiple benchmarks, providing a solid foundation for future research in this field.
We hope this survey serves as a comprehensive reference and consolidated baseline for future development.
Code is publicly available at: \url{https://github.com/Visual-AI/Semantic-Correspondence}.

\end{abstract}

}

\IEEEdisplaynontitleabstractindextext
\IEEEpeerreviewmaketitle

%%%%%%%%%%%%%%%%%%%%%%%%%%%%%%%%%%%%%%%%%%%%%%%%%%%%%%%%%

% !TEX root = ../main.tex 

% \IEEEraisesectionheading{
% \section{Introduction}\label{sec:introduction}
% }

\section{Introduction}\label{sec:introduction}
\IEEEPARstart{S}{emantic} matching establishes keypoint correspondences that share equivalent functional or contextual roles across different images, regardless of variations in appearance, pose, or viewing conditions. For instance, it can identify corresponding points between images such as the tips of animal ears, corners of building windows, or joints of human bodies, even when these elements exhibit different visual characteristics. Semantic correspondence facilitates various computer vision tasks: it enables structural analysis in scene understanding~\cite{SFNet,SFNet2022, SIFTFlow}, ensures semantic consistency of content during style transfer~\cite{lee2020reference, geyer2023tokenflow}, and guides coherent manipulations in image editing~\cite{RegionDrag, ofri2023neural,mou2023dragondiffusion}. By establishing meaningful point-to-point relationships across images, it provides essential semantic guidance for these applications.
% challenge
However, establishing these correspondences remains challenging, as keypoints with the same semantic information may have significantly different appearances. This requires methods to rely on high-level abstract information to find matches.
% handcrafted 
Early methodologies~\cite{SIFTFlow,cho2015unsupervised,HOG} primarily rely on handcrafted features to establish correspondences in image pairs.
However, these are fragile when faced with large appearance variations.
% CNN
More recently, learned features have become prominent, allowing semantic correspondence to achieve significant progress in recent years. 
Recent approaches~\cite{SCNet,FCSS,HyperpixelFlow,DynamicHyperpixelFlow,DCCNet,HCCNet} rely on deep convolutional neural networks (CNNs) to extract semantic features.
% Backbones~\cite{ResNet, VGG} trained on large-scale image classification datasets~\cite{Imagenet} have garnered significant attention for its vast knowledge of different objects.
% Such knowledge is particularly beneficial to semantic correspondence, so most works focus on how to leverage this knowledge and further adapt it to this task. 
% architectures and paradigms
A wide range of network architectures and paradigms have been explored, ranging from CNN architectures~\cite{NC-Net, ANC-Net, CHM, PMNC} to transformer modules~\cite{CATs,CATs++,TransforMatcher}.
% foundation model
More recently, vision foundation models, such as DINOv2~\cite{DINOv2} and Stable Diffusion~\cite{Stable-Diffusion}, have demonstrated remarkable zero-shot semantic correspondence capability~\cite{DIFT, A-Tale-of-Two-Features,DINOv3}, enabling follow-up works~\cite{SD4Match, GeoAware-SC,DistillDIFT, SemAlign3D, mariotti2025jamais, qian2025bridging, SCAC} to achieve state-of-the-art performance.

Semantic matching is a relatively new and distinct task in computer vision, standing apart from established tasks such as geometry matching, stereo matching, and optical flow estimation. 
% Geometry matching
Geometry matching~\cite{hartley2003multiple, ma2021image} primarily involves identifying correspondences between points representing the same physical location across images. It heavily relies on spatial consistency, often constrained by the spatial arrangement of the scene. 
% Stereo matching
Stereo matching \cite{marr1979computational}, on the other hand, estimates depth by matching corresponding pixels in rectified stereo images, leveraging the constraints of epipolar geometry. 
% Optical flow
In contrast, optical flow estimation \cite{horn1981determining} emphasizes the motion of pixels or objects between consecutive frames in a video sequence to capture temporal dynamics. 
% Semantic correspondence
Unlike these tasks, semantic correspondence focuses on aligning semantically similar regions or objects across images, without being bound by spatial and geometric constraints. 
This task is unique in its objectives, methodologies, and application scenarios. 
Therefore, in this paper, we focus primarily on the unique challenges and advances of semantic matching.

% taxonomy
\begin{figure*}[h!]
	\centering
	% \vspace*{-10mm}
	\includegraphics[width=1.0\linewidth]{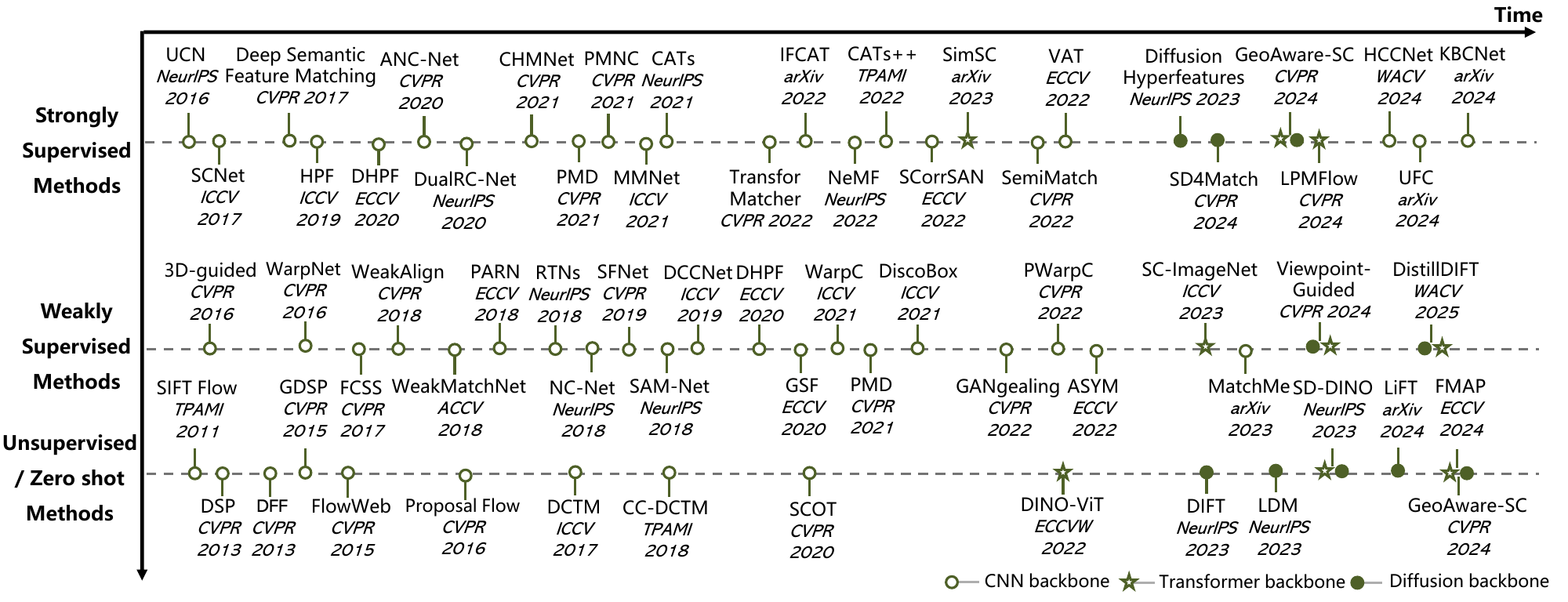}
	\vspace{-1em}
	\caption{\textbf{The development timeline of semantic correspondence methods categorized by their levels of supervision. }
	These categories include strongly supervised, weakly supervised, and zero-shot methods.
	Each method is labeled with its backbone architecture, indicated by distinct symbols: 
	$\bigcirc$ for CNN backbones, 
	\ding{73} for vision transformer backbones, 
	and \raisebox{-1.7mm}{\scalebox{3}{\textbullet}} for stable diffusion backbones. 
	For methods that utilize both vision transformer backbones and stable diffusion  backbones (e.g., DINOv2+SD), we use a combination of a \ding{73} and a \raisebox{-1.7mm}{\scalebox{3}{\textbullet}} to represent their hybrid architecture.
	}
	\label{fig:Timeline}
\vspace{-1em}
\end{figure*}
\begin{figure*}[h!]
	\centering
	%	\vspace*{-25mm}
	\includegraphics[width=0.9\linewidth]{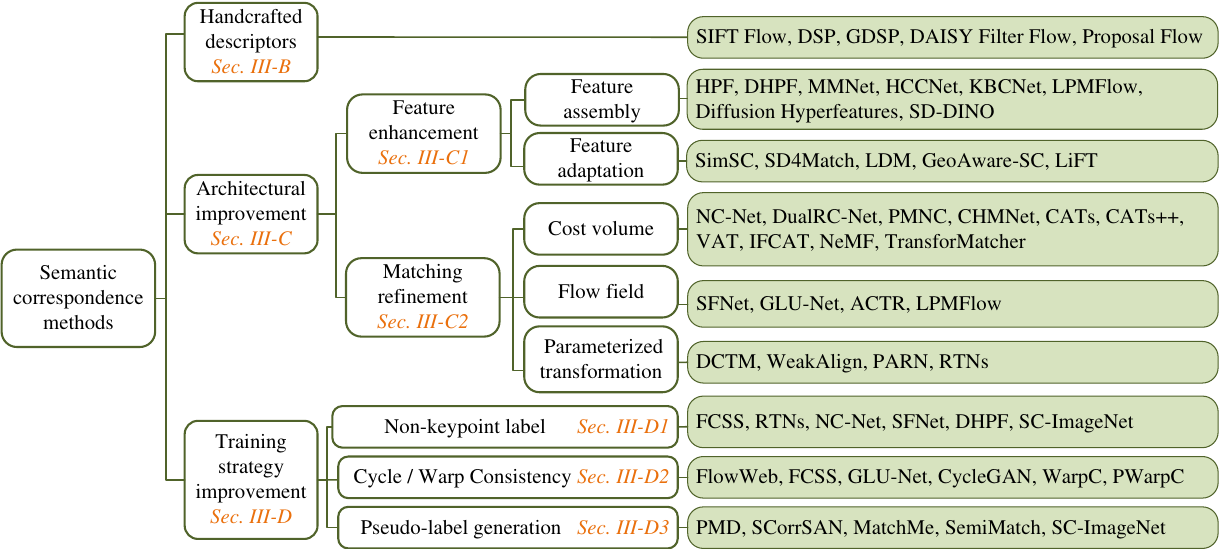}
	% \vspace{-1em}
	\caption{\textbf{Taxonomy of semantic correspondence methods.} 
	This taxonomy provides a comprehensive overview of the diverse approaches to enhance feature quality, matching performance, or training strategies. 
	Only a few representative methods of each category are shown.}
	\label{fig:taxonomy}
% \vspace{-1em}
\end{figure*}

% Gap 
Since the study of semantic matching has evolved significantly, it is critical to track the literature, categorize and analyze existing methods, and reflect on their progress.
\Cref{fig:Timeline} presents a summarized timeline of the development of semantic matching methods, highlighting the rapid progress in this field. 
The diversity and sheer number of emerging methods make it challenging to discern their relationships and developmental trajectories. 
Furthermore, the fragmented nature of existing literature hinders researchers' understanding of the overall impact of these methods on the task, let alone grasping their differences and innovations.
To address this challenge, we propose a taxonomy to classify existing methods based on their general designs, dividing them into three main categories: early attempts, architectural improvements, and training strategy improvements, as shown in \Cref{fig:taxonomy}. 
Early attempts primarily tackle this task using handcrafted features such as SIFT~\cite{sift} or HOG~\cite{HOG}. 
Architectural improvements focus on enhancing representation learning and matching quality, while training strategies aim to improve the matching performance by optimizing how the model learns. 
This taxonomy provides a structured framework for systematically analyzing and comparing different methods. 
It enables researchers to quickly identify the position and contributions of each approach within the field and supports the establishment of unified benchmarking for consistent evaluation.

% benchmark
Existing methods exhibit differences in backbone type, fine-tuning strategy, and model architecture, yet a unified evaluation remains absent. 
This inconsistency hinders objective comparisons, making it difficult to isolate the impact of specific components.
To address this, we conduct extensive experiments across these factors and reveal that fine-tuning a powerful backbone is the most decisive factor affecting performance.
Building on this finding, we propose a benchmark to enable more objective and transparent comparisons.

% baseline
Based on our benchmarking and analysis, we integrate the most effective components from each stage of semantic correspondence methods to propose a simple yet strong baseline. This baseline achieves state-of-the-art performance across multiple public benchmarks, providing a strong foundation for future development.

In summary, we make the following contributions:
\begin{itemize}
    \item We propose a clear taxonomy and provide a comprehensive survey of existing semantic correspondence methods, categorizing them based on their general design and summarizing their key contributions.
    \item We systematically conduct extensive experiments to evaluate the performance of existing methods across various benchmarks, providing insights into their strengths and weaknesses. 
    \item We propose a simple yet effective baseline that achieves state-of-the-art performance on multiple benchmarks, offering a strong foundation for future research in this field.
\end{itemize}
\section{Problem Statement}\label{sec:Problem}
Given a pair of source and target images, $I^s$ and $I^t$, and a set of query points $\mathbf{X}^s = \{\mathbf{x}^{s}_{q} = (x^{s}_{q}, y^{s}_{q}) \mid q = 1, 2, \ldots, n\}$ in $I^s$, semantic matching aims to find corresponding points $\bar{\mathbf{X}}^t = \{\bar{\mathbf{x}}^{t}_{q} = (\bar{x}^{t}_{q}, \bar{y}^{t}_{q}) \mid q = 1, 2, \ldots, n\}$ in $I^t$ that share the same semantic meaning as those in $\mathbf{X}^s$.

Semantic matching methods can be broadly divided into sparse and dense correspondence, depending on the number of matched points. 
Sparse methods estimate correspondences for a few query points, while dense methods aim to predict correspondences for all pixels. 
% Accordingly, their training objectives differ: sparse methods typically use classification losses (e.g., entropy loss), whereas dense methods adopt regression losses such as the Average End-Point Error (AEPE) \cite{Dgc-net}.
A widely adopted pipeline for semantic correspondence involves extracting feature vectors $\{f^s_q \in \mathbb{R}^C \mid q = 1, 2, \ldots, n\}$ for the query points $\mathbf{X}^s$, and a dense feature map $F^t \in \mathbb{R}^{H_t \times W_t \times C}$ for the target image $I^t$, using a feature extractor $f(\cdot)$. Each query feature $f^s_q$ is then compared to all locations in $F^t$, and the most similar feature $\bar{f}^t_q$ is identified. The spatial location of $\bar{f}^t_q$ within $F^t$ is taken as the predicted correspondence.

The most crucial component in this pipeline is the feature extractor $f(\cdot)$, as it forms the foundation for accurate matching. Consequently, designing a discriminative and generalizable feature extractor has become a central research focus in this domain. In addition, the inference stage---beyond simply selecting the most similar feature---can also be improved using more sophisticated algorithms, representing another active research direction.

In the following sections, we present our taxonomy of existing literature and provide a detailed review of each method.

\section{Method Review}\label{sec:METHOD}

\subsection{Overview}
To provide a structured understanding of semantic correspondence methods, we present a taxonomy categorizing approaches into handcrafted descriptors, architectural improvements, and training strategy improvements, as shown in \Cref{fig:taxonomy}. 
This taxonomy traces the evolution from handcrafted methods to advanced deep learning solutions, providing a clear overview of how different approaches enhance feature quality, matching performance, and training strategies.

\subsection{Handcrafted Descriptors}
% Hand-craft 
Early methods establish semantic correspondence by matching two sets of local regions based on handcrafted feature descriptors.
% For example, Scale-Invariant Feature Transform (SIFT)~\cite{sift} and Histograms of Oriented Gradients (HOG)~\cite{HOG} are widely used to extract local features from images.
% SIFT
For example, Lazebnik~\etal~\cite{Ponce2005} utilized Scale-Invariant Feature Transform (SIFT)~\cite{sift} to represent local regions (keypoints) in images, capturing features that correspond to textures or object components.
Building on this, Kushal~\etal~\cite{Ponce2007} leveraged SIFT for appearance-based matching of salient regions. 
\rev{In order to achieve real-time computation, a number of faster methods were developed.
% SURF
Speeded-Up Robust Features (SURF)~\cite{SURF} emerged as an alternative to SIFT, approximating its performance with a substantial speed increase by using integral images and a Hessian matrix-based detector. 
Subsequently, Oriented FAST and Rotated BRIEF (ORB)~\cite{ORB} was introduced as a patent-free and even faster option, combining the FAST keypoint detector with the BRIEF binary descriptor. ORB's high efficiency in both feature extraction and matching made it ideal for resource-constrained environments.
}

% SIFTFlow
\rev{Moving beyond sparse descriptors, another line of work focused on establishing dense, pixel-level correspondence.
A notable example is SIFT Flow~\cite{SIFTFlow}, which adapted optical flow principles to establish a semantically coherent alignment across the entire image.}
\rev{This dense approach was later refined by DSP~\cite{DSP2013}, which introduced a multi-scale regularization scheme to improve matching accuracy.}
% others
Yang~\etal~\cite{DAISY-Filter-Flow} employed DAISY~\cite{DAISY} descriptors to handle non-rigid geometric transformations. 
Meanwhile, Cho~\etal~\cite{cho2015unsupervised} proposed a voting-based algorithm that utilizes region proposals and Histograms of Oriented Gradients (HOG)~\cite{HOG} features for semantic matching and object discovery.
Ham~\etal~\cite{ham2016proposal, ham2018proposal} also utilized HOG features extracted from object proposals to establish region correspondence.

% summary
% \rev{These techniques
% typically —from sparse descriptors, such as SIFT~\cite{bristow2015dense, DSP2013, GDSP, SIFTFlow},  SURF~\cite{SURF}, and ORB~\cite{ORB}  to dense ones like HOG~\cite{ProposalFlow2016, ProposalFlow2018, TSS, OHG} and DAISY~\cite{DAISY-Filter-Flow} —
% formed the bedrock of early semantic matching. 
\rev{Apart from descriptor-oriented methods, other approaches were also explored, such as graph-based matching algorithms~\cite{graph-matching, Cho2012ProgressiveGM}, which leverage flexible graph representations to find category-level feature matches, 
and probabilistic matching algorithms~\cite{cho2015unsupervised, ham2016proposal, ham2018proposal} that model uncertainty in feature correspondence using probabilistic frameworks.}
% But they often struggle with large appearance or viewpoint changes, which limits their robustness in more challenging scenarios.

\rev{However, these methods face inherent limitations when applied to complex computer vision tasks. 
They struggle to scale efficiently to large datasets, and their low-level, gradient-based features lack robustness in complex scenes involving significant geometric distortions, appearance variations, and large intra-class variance. 
This performance gap catalyzed the shift from handcrafted engineering to the learned representations that are the focus of this work.}
With the rise of deep learning, research has moved from handcrafted to learned features, improving reliability and enabling more flexible, scalable solutions for semantic matching.
In this work, we focus on learning-based approaches, which have shown notable effectiveness in addressing these challenges.

\subsection{Architectural Improvements} \label{sec:arch_improve}
\begin{figure*}[h!]
	\centering
	% \vspace*{-5mm}
	\includegraphics[width=0.6\linewidth]{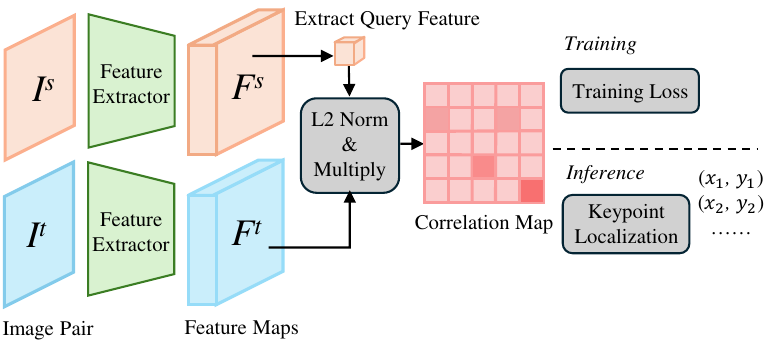}
	\vspace{-1em}
	\caption{
	\textbf{Pipeline for feature enhancement methods.}
	The feature extractor generates feature maps $F^s$ and $F^t$ from the source image $I^s$ and target image $I^t$, respectively.
	After channel-wise L2 normalization, their dot product constructs a cosine similarity matrix (2D correlation map) for each query point in $F^s$.
	The correlation map is transformed into a probability distribution through a localization operation such as soft-argmax, which is then supervised by a ground-truth distribution derived from ground-truth correspondences using the same localization technique.
	During inference, the correlation map localizes the correspondences$(x_1, y_1), (x_2, y_2), \dots$ in $I^t$.
	}
	\label{fig:pipeline_backbone}
\vspace{-1em}
\end{figure*}

Backbones trained on large-scale datasets have shown strong general-purpose feature extraction capabilities.
Consequently, research in this field has increasingly shifted toward how to effectively leverage learned features by innovation in the architecture of the overall pipeline.

Generally, the literature in architectural improvements can be divided into two families: feature enhancement and matching refinement. The former focuses on extracting more discriminative features and could be further categorized into two sub-families: feature assembly and feature adaptation, while the latter aims to refine matches established by feature maps and can be categorized into three classes: cost volume, flow field, and parameterized transformation-based matching refinement. In the following paragraphs, we first introduce feature enhancement and its two sub-families in~\Cref{sec:feat_enhance}, followed by a review of three matching refinement classes in~\Cref{sec:match_filter}.

\subsubsection{Feature Enhancement}\label{sec:feat_enhance}

% early CNN methods
\xh{Early deep learning methods typically learned features from scratch. For instance, Choy~\etal~\cite{UCN} developed a similarity metric using a contrastive loss combined with hard negative mining; Han~\etal~\cite{SCNet} proposed a scoring function using a CNN to rank the similarity between patches identified by region proposals and determined correspondences through probabilistic Hough matching~\cite{HOG}; Kim~\etal~\cite{FCSS} introduced a CNN self-similarity feature, which is utilized to estimate dense affine transformation fields. While these approaches showcase promising results, their performance and generalizability are constrained by the limited scale of training data.}

\xh{Long~\etal~\cite{long2014convnets} have shown that CNNs pre-trained on image classification tasks could extract features for semantic correspondence without additional training. 
The capability to perform matching without task-specific fine-tuning in some cases highlights the strength of object information learned from millions of images. 
This suggests that, instead of training semantic features from scratch, one could develop a more accurate and generalizable matching pipeline by leveraging pre-trained CNNs. 
Such a pipeline takes a straightforward architecture: a dense feature map is first extracted from the input image and the correlation map between query and candidate features is computed. This correlation map is then transformed into correspondences using keypoint localization techniques such as argmax or kernel soft-argmax operations~\cite{SFNet}. We illustrate this pipeline in~\Cref{fig:pipeline_backbone}.} 

\xh{A key component in the pipeline is the feature extractor, as it significantly affects the subsequent correlation map and the accuracy of the final correspondence.  While a pre-trained CNN, as shown by \cite{long2014convnets}, can fulfill this purpose, it can be further enhanced to yield more powerful feature extractors. 
Two commonly used strategies in the literature for improving feature extraction are \textbf{\textit{feature assembly}} and \textbf{\textit{feature adaptation}}. Feature assembly constructs hyper-features by combining multiple complementary features \cite{hariharan2015hypercolumns, HyperpixelFlow,MMNet, HCCNet, KBCNet, LPMFlow, Diffusion-Hyperfeatures, A-Tale-of-Two-Features}, while feature adaptation employs an adaptation module to transfer pre-trained features to the semantic matching task \cite{SimSC, SD4Match, LDM_correspondences, GeoAware-SC, LiFT}. We review both strategies in the following paragraphs. It is worth noting that more recent methods \cite{Diffusion-Hyperfeatures, A-Tale-of-Two-Features, DIFT, GeoAware-SC} switch from CNN backbones to more powerful counterparts such as diffusion models \cite{Stable-Diffusion} or DINOv2 \cite{DINOv2} and significantly improve accuracy. 
The general ideas behind these methods fall into our categorization. Our review is not tied to a specific backbone family but approaches from the perspective of methodology.}

\noindent\textbf{Feature Assembly.} \  
\xh{Semantic matching finds correspondences by overcoming appearance differences brought by intra-category variation and pose changes. 
A good feature should contain both abstract information—such as high-level semantic priors that encode object structure and functional part relationships—and local details to accurately localize the correspondence.}
Consequently, one line of work focuses on assembling features from different layers of CNN to produce a hyper-feature that includes different levels of information about the scene.
% Explanation of CNN Hierarchy
% \xh{It was discovered that CNN backbones pretrained on the image classification task can extract features at different semantic levels in a bottom-up manner~\cite{zeiler2014visualizing}.}
% The features from the bottom convolutional layers, which are close to the input image, act as low-level feature descriptors and are sensitive to colors, edges, textures, and other low-level statistics, \xh{while the top convolutional layers capture high-level semantics. 

% HyperpixelFlow
Hariharan~\etal~\cite{hariharan2015hypercolumns} \xh{showed that combining features from multiple layers of CNN into a ``hypercolumn'' significantly improves feature performance in object detection, segmentation, and part labeling.}
Inspired by this, Min~\etal~\cite{HyperpixelFlow} introduced ``hyperpixels'', which leverages different levels of features from early to late layers of CNNs with beam search~\cite{beamsearch}, enabling the disambiguation of image parts across multiple visual aspects.
% Dynamic Hyperpixel Flow
Further extending this idea, dynamic hyperpixel flow (DHF)~\cite{DynamicHyperpixelFlow} was developed to dynamically compose hypercolumn features by selecting a small number of relevant layers from CNN.
By leveraging these relevant layers, conditioned on the specific images to be matched, DHF~\cite{DynamicHyperpixelFlow} composes effective features on the fly, and the resultant hyperpixels provide both fine-grained and context-aware features, making DHF a powerful approach for feature assembly.
% MMNet
Unlike ``hypercolumns'' and ``hyperpixels'', which are obtained by searching or selecting features from various layers, MMNet~\cite{MMNet} fuses feature maps from all residual blocks within the same group. 
In contrast to FPN~\cite{FPN}, which only utilizes the feature map of the last residual block in each convolutional group, MMNet~\cite{MMNet} captures semantics at different levels.
% HCCNet
Kim~\etal~\cite{HCCNet} proposed a technique called ``feature slicing'' to generate richer multi-scale correlation maps by slicing intermediate feature maps into smaller, equi-channel segments, which maximize the potential of the constructed hypercolumn correlation.
These multi-scale correlation maps are then concatenated along the channel dimension to obtain a hypercolumn correlation.
% KBCNet
Jin~\etal~\cite{KBCNet} proposed the Keypoint Bounding Box-Centered Cropping (KBC) method. 
This approach involves \xh{digitally upsampling} the source image to increase the separation of close keypoints on small objects if necessary, and then cropping it to match the input size based on the keypoint bounding box. 
This technique facilitates the independent learning of these keypoints.
KBCNet~\cite{KBCNet} applies cross-attention to leverage fine-grained and deep semantic features across scales within the same image, thereby enhancing the robustness and accuracy of semantic correspondence.
% LPMFlow
LPMFlow~\cite{LPMFlow} progressively generates feature maps at different scales by employing self-attention and cross-attention mechanisms to enhance and upsample features, producing hybrid enhanced features that incorporate both self-feature guidance and interaction-enhanced super-resolution features, aiming to obtain detailed semantic representations while increasing feature resolution.

% foundation model feature assembly
More recently, features of large foundation models, such as diffusion models \cite{Stable-Diffusion, DIFT} and DINOv2 \cite{DINOv2}, have better performance than those of CNN backbones due to larger training data and attention. 
This encourages research \cite{Diffusion-Hyperfeatures, A-Tale-of-Two-Features, GeoAware-SC} in adopting the idea of feature assembly to foundational models.
For instance, Luo~\etal~\cite{Diffusion-Hyperfeatures} proposed to concatenate all intermediate feature maps from the diffusion model, \xh{across different network layers and diffusion timesteps}, into a single per-pixel descriptor for semantic correspondence. 
Similarly, Zhang~\etal~\cite{A-Tale-of-Two-Features} observed that \xh{features from Stable Diffusion (SD) exhibit complementary properties to those from DINOv2. Therefore, by combining them into a single feature, they achieve stronger zero-shot performance.
}

% summary
\xh{Although feature assembly achieves good results by constructing ``hyper-features'', the dimension may be very large, which increases the memory consumption and slows down the pipeline. 
An alternative is to select a specific layer feature and adapt it to the matching task, which keeps the dimension of the feature relatively low, accelerating matching. 
We review this line of work in the next paragraphs.}

\begin{figure*}[h!]
	\centering
	% \vspace*{-5mm}
	\includegraphics[width=0.7\linewidth]{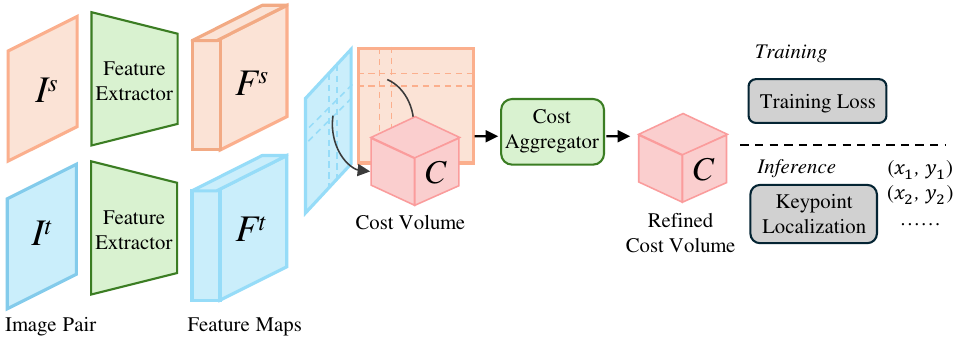}
	\vspace{-1em}
	\caption{
	\textbf{Pipeline for cost volume-based methods.}
	The feature extractor generates feature maps $F^s$ and $F^t$ from the source image $I^s$ and target image $I^t$, respectively.
	After normalization, their dot product constructs a cost volume for each query point in $F^s$, storing the cosine similarity between all possible feature pairs. 
	This cost volume is refined by a cost aggregator and converted into a probability distribution via softmax, supervised by a ground-truth distribution derived from known correspondences. 
	During inference, the correlation map localizes the correspondences$(x_1, y_1), (x_2, y_2), \dots$ in $I^t$.}
	\label{fig:pipeline_cost}
\vspace{-1em}
\end{figure*}

\noindent\textbf{Feature Adaptation.} \  
\xh{Unlike feature assembly, feature adaptation improves the quality of features by adapting them to the semantic matching task. It can be achieved by appending a trainable adaptation module or fine-tuning the feature backbone using matching loss.}
% SimSC
\xh{SimSC~\cite{SimSC} introduces a generic and simple learning pipeline to fine-tune backbones using adaptive temperature learning. It employed a lightweight temperature learning module to find the optimal temperature for the softmax operation to alleviate the gradient suppression caused by L2 normalization during training. 
% SD4Match
Building on top of it, SD4Match~\cite{SD4Match} explores how to optimize the prompt to enhance the performance of Stable Diffusion's feature~\cite{DIFT} in semantic correspondence, and introduces a conditional prompting module to infuse prior knowledge of objects into the diffusion model, increasing the quality of the feature. 
% LDM
LDM~\cite{LDM_correspondences} leverages Stable Diffusion for semantic matching from a different perspective. It borrows the idea of textual inversion~\cite{textual_inversion} and learns an embedding for a query point in the source image. Such an embedding is transferred to the target image to locate the correspondence from the attention map of the target image with the embedding. 
% GeoAware-SC
GeoAware-SC~\cite{GeoAware-SC} employs SD-DINO~\cite{A-Tale-of-Two-Features} as the feature backbone and uses a CNN to adapt the feature to the semantic matching task. Additionally, it improved the geometry-aware ability of the model by data augmentation.
% SemAlign3D 
\rev{Taking a different approach focused on 3D geometry, SemAlign3D \cite{SemAlign3D} builds 3D object class representations from a small set of sparsely annotated RGB images. 
The core of the method is to align the 3D object-class template to each individual image instance to construct scale-invariant geometric features. 
This approach is particularly effective for rigid objects, demonstrating strong robustness under extreme viewpoint changes.}
% LiFT
LiFT~\cite{LiFT} introduces a post-processing method to replace the naive bilinear interpolation when upsampling the ViT feature for better performance. Starting from a low-resolution ViT feature, it learns to reconstruct the high-resolution ViT feature in a self-supervised manner.}
% CleanDIFT
\rev{CleanDIFT \cite{CleanDIFT} employs lightweight fine-tuning and knowledge distillation to create a single, noise-free feature extractor from all diffusion timesteps. 
% introduces a lightweight fine-tuning strategy that adapts a pre-trained diffusion backbone to extract noise-free, timestep-independent features for dense downstream tasks. By distilling knowledge across all diffusion timesteps into a single, task-agnostic feature extractor, it eliminates the need for noisy inputs and per-task timestep tuning while consistently improving semantic correspondence performance.
}
% DiTF
\rev{Complementing these approaches, DiTF~\cite{DiTF} presents a training-free framework that directly extracts discriminative features  by modulating internal activations using the model's AdaLN-zero mechanism, without updating any parameters.
% from Diffusion Transformers by modulating massive activations through the inherent AdaLN-zero mechanism. Without any parameter updates, DiTF employs channel-wise modulation followed by a simple channel-discard strategy.
}
% Summary
\xh{Compared with other methods, feature adaptation is an efficient pipeline as the network focuses solely on extracting features from images. However, the model is prone to overfitting the training dataset, showing inferior generalizability.}

\xh{We have reviewed feature enhancement, which primarily focuses on improving features extracted from the images. The second family of work, matching refinement, aims to refine the initial matches established from the feature backbone. We will review this topic in the next section.}

\subsubsection{Matching Refinement}\label{sec:match_filter}
After obtaining feature maps, matches can be established by comparing the features of the source image with those of the target image. 
% basic matching
The most intuitive approach is to compute the similarity between each pair of features and select the best match. Several keypoint localization techniques have been explored:
\begin{itemize}
    \item \textbf{Nearest Neighbor (NN)} \cite{NC-Net, FMAP,GLU-Net, Deep-vit-features, DIFT, ufer2017deep, A-Tale-of-Two-Features}:  For each feature in the source image, it finds the feature in the target image with the highest similarity score through direct comparison of feature vectors.
    \item \textbf{Bilinear Interpolation} \cite{SimSC, SFNet, SFNet2022, TransforMatcher, LDM_correspondences, IFCAT, WarpC,DualRC-Net, DualRC2020, ACTR, LPMFlow}: It computes the correspondence by taking weighted combinations of neighbouring features, where the weights are determined by the relative position of the query point to its neighbours in a regular grid.
    \item \textbf{Kernel Soft Argmax} \cite{SimSC, GSF, SFNet, SFNet2022,SemiMatch, HCCNet,CATs,CHM, SCorrSAN}: This method extends Soft Argmax by using a kernel function to convert feature distance into similarity. In general, similarity score $s_i$ is first converted into probability by softmax: $p_i = \frac{\exp(s_i)}{\sum_j \exp(s_j)}$, then the expected coordinate is computed as $\sum_i p_i \cdot c_i$, where $c_i$ is the coordinate value.
    For example, with a Gaussian kernel, the similarity is computed as $\exp(-\|f_1-f_2\|^2/\sigma^2)$, where $f_1$ and $f_2$ are feature vectors and $\sigma$ controls the kernel width.
    \item \textbf{Window Soft Argmax} \cite{SAM-Net, GeoAware-SC, DistillDIFT, thai20243}: For each feature point, this method defines a $k \times k$ window centered at that point. The matching score at position $(x,y)$ is computed as the weighted sum of similarities within this window: $\sum_{i,j \in W} w_{ij} \cdot s(f_{x+i,y+j})$, where $W$ is the window region, $w_{ij}$ is the spatial weight, and $s(\cdot)$ is the similarity function.
\end{itemize}
However, these methods heavily rely on the quality of the feature map and consider each match individually. They are prone to outliers and noise in the feature map.

To address this issue, match refinement has been explored to consider the matching globally.
Intuitively, good matches should be spatially continuous, smooth and follow certain patterns, rather than being stochastic with abrupt changes. This indicates that teaching the model to recognize and correct unreliable matches could also improve the matching accuracy. 
Generally, there are three common formats for matches: cost volume~\cite{NC-Net, TransforMatcher, ANC-Net, CHMNet, PMNC}, flow field~\cite{LPMFlow, SFNet, SFNet2022, ACTR} and parameterized transformation~\cite{WeakAlign, DCTM}. Cost volume is a high-dimensional tensor storing the likelihood of all possible matches between two images; flow field is a 2D vector field indicating the correspondence for each spatial position; and parameterized transformation represents matches as the warping between two images using few parameters. The majority of the literature in this direction refines matches based on these three formats. We review methods on each format in the following paragraphs.

\noindent\textbf{Cost Volume.} \ 
% Motivation and Background
The cost volume, also known as the correlation tensor, is a high-dimensional tensor that stores the similarities between all possible feature pairs between two images. Given two image feature maps \(F_{s}\in \mathbb{R}^{H_s\times W_s\times C}\) and \(F_{t}\in \mathbb{R}^{H_t\times W_t\times C}\), the cost volume is normally expressed as a 4D tensor \(C\in\mathbb{R}^{H_s\times W_s\times H_t\times W_t}\) or a 2D matrix \(C\in\mathbb{R}^{H_s W_s\times H_tW_t}\). The simplest way to find the correspondence is by slicing the cost volume at the position of the query point and applying the argmax operation. With a perfect feature extractor, this is good enough to localize accurate correspondence. However, the real feature extractor could be mistaken by the ambiguity or noise within images, leading to incorrect correspondences after the argmax operation. Apart from features, matching patterns provide valuable information as well. Intuitively, good dense matching should be smooth and continuous without abrupt changes. This provides extra constraints on matching, offering an alternative perspective to feature enhancement methods and complementary insights into the matching problem. 
Based on this intuition, the cost volume-based methods apply a learnable cost aggregation module to the cost volume and refine it by recognizing the correct matching pattern. ~\Cref{fig:pipeline_cost} illustrates the general pipeline of this line of work.

\begin{figure*}[h!]
	\centering
	% \vspace*{-5mm}
	\includegraphics[width=0.75\linewidth]{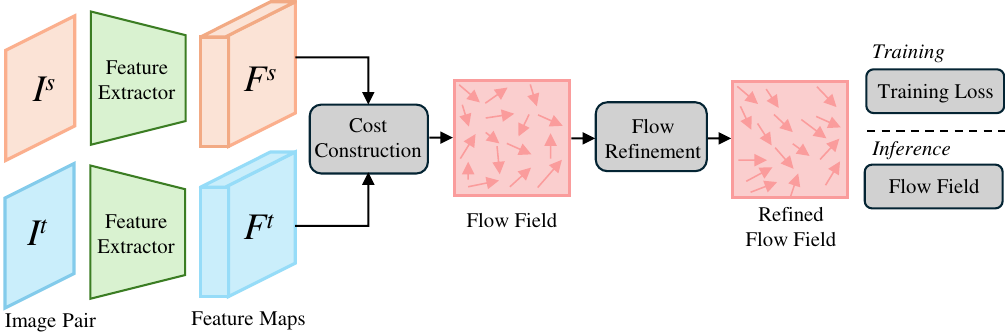}
	\vspace{-1em}
	\caption{
		\textbf{Pipeline for flow field-based methods.}
		Given a pair of images $I^s$ and $I^t$, feature extraction is performed to obtain dense feature maps  $F^s$ and $F^t$ respectively.
		The cost construction step is then applied to derive a cost volume, which is subsequently transformed into a flow field. 
		The flow field represents the correspondence between each pixel in the target image and its corresponding pixel in the source image.}
	\label{fig:pipeline_flow}
\vspace{-1em}
\end{figure*}

One of the most powerful constraints is the neighbourhood consensus constraint, which assumes that correspondences of adjacent query points should follow a similar local pattern to the query points. It has been widely used in geometric matching to filter outliers~\cite{bian2017gms, sattler2009improving, schaffalitzky2002automated}. In the context of semantic correspondence, the seminal work in exploring the neighbourhood consensus constraint is NC-Net~\cite{NC-Net}. It employs a 4D convolutional module, consisting of a series of $5\times 5\times 5\times 5$ convolutional kernels, and applies it to the cost volume to recognize reliable matches from local matching patterns, showing good results. It inspired a line of follow-up works exploring the cost volume for semantic matching: 
% ANC-Net --- adaptive neighbourhood consensus 
ANC-Net~\cite{ANC-Net} proposes an adaptive neighbourhood consensus module that contains a set of non-isotropic 4D convolutional layers, such as kernels with the size of $3\times 3\times 5\times 5$, enabling the model to handle the scale difference between objects.
% DualRC-Net --- dynamic neighbourhood consensus
DualRC-Net~\cite{DualRC2020, DualRC-Net} follows the idea of ANC-Net, and proposes a learning mechanism to dynamically combine different non-isotropic 4D kernels, avoiding manually searching for the optimal weights for combination.
% CHM CHMNet 6D Hough Matching
CHM \cite{CHM,CHMNet} tackles the scale issue by incorporating scale difference into the cost volume. In addition to two spatial dimensions height and width, an extra dimension of scale upgrades the cost volume to 6D. By employing a 6D convolutional module, the method performs position-aware Hough voting in the high-dimensional space, improving the matching accuracy.
% MMNet
MMNet~\cite{MMNet} builds the final cost volume by incorporating information from intermediate features and cost volumes.
% Guided Semantic Flow (GSF)
GSF~\cite{GSF} first employs a learnable pruning module to select reliable initial matches from the cost volume and uses selected matches as the anchor point and refines neighbouring matching scores using the Gaussian process.  
% PMNC -- patch-match
PMNC~\cite{PMNC}, instead of processing the full cost volume, predicts the matching score of a specific correspondence pair by taking the local cost volume centered at the pair and iteratively refining the initial matches using the PatchMatch~\cite{bleyer2011patchmatch} algorithm. As it only processes a fraction of the cost volume at a time, this method is significantly faster than the aforementioned methods.
% NC-Net 系列总结
Although the neighbourhood consensus constraint by high-dimensional convolution is effective in recognizing correct matches, its focus on locality limits the ability to capture global matching patterns. This has motivated a shift towards transformer architecture, which enables direct modeling of long-range dependencies and global matching patterns in the correspondence space.

% Transformer Introduction
Transformer network architecture~\cite{Transformer} has demonstrated remarkable capability across a wide range of computer vision tasks~\cite{ViT,DETR,DeformableDETR,LoFTR}. The success is attributed to the global dependency between features realized by the attention mechanism.
% CATs 系列 CATs -- CATs++ -- VAT -- NeMF --IFCAT -- TransforMatcher
% CATs
CATs~\cite{CATs} is the first to employ a transformer module to do cost aggregation, aiming to overcome the limitations of CNNs' local receptive fields through the global self-attention mechanism of this architecture, thereby demonstrating the feasibility of the transformer as a cost aggregator.
Specifically, CATs transforms the cost volume into a token sequence and incorporates positional embeddings to retain spatial information. 
The tokenized sequence is then processed by a standard transformer module, which performs global aggregation and optimization of the matching scores, producing a refined cost volume.
% CATs++ -- computational cost
CATs++~\cite{CATs++} extends CATs~\cite{CATs} by introducing early convolutions before the transformer cost aggregator. By leveraging both convolution and transformer, it combines the strengths of both techniques and enhances the model's ability through local interactions before globally aggregating pairwise interactions.
% VAT -- inductive bias 
\xh{VAT~\cite{VAT} introduces a 4D Convolutional Swin Transformer, which consists of a high-dimensional Swin Transformer \cite{SwinTransformer} preceded by a series of small-kernel convolutions. This alleviates the local context loss caused by the tokenization of the transformer, improving the capture of local features and overall performance.}
% NeMF inductive bias
\xh{Similarly, NeMF~\cite{NeMF} tackles the local context loss by integrating the transformer architecture with convolutional operators. 
This approach allows for the local and global integration of matching cues by encapsulating local contexts and imparting them to all pixels via self-attention.}
% IFCAT --- feature descriptors 和 cost volume
\xh{In addition to aggregation on the cost volume, IFCAT~\cite{IFCAT} employs a similar aggregation module directly on the feature space, leveraging their complementarity. The aggregation in the feature space helps disambiguate the noise in the cost volume, while the cost volume enhances feature aggregation by introducing matching similarities as a factor for aggregation. }
% TransforMatcher -- patch-to patch - match-to-match
\xh{TransforMatcher~\cite{TransforMatcher}, unlike previous methods that reshape the 4D cost volume into 2D for patch-to-patch attention, treats every match on the correlation map as an input element. It uses multi-level scores as features to perform match-to-match attention, enabling more fine-grained interactions.}

% summary
Transformer cost aggregation methods have introduced a new paradigm to semantic correspondence by leveraging the attention mechanisms of the transformer. Despite the effectiveness of the global self-attention mechanism in integrating global information, computational efficiency becomes a drawback when processing high-resolution images. 
Additionally, due to the data-hungry nature of the transformer architecture, these methods require extensive data augmentation to improve generalizability.

\noindent\textbf{Flow Field.}
In addition to representing the dense matching between two images as a cost volume, a common alternative is the flow field. A flow field is a 2D vector field \(M\in R^{H_{s}\times W_{s}\times 2}\) where each spatial location records the offset to its correspondence in the target image. The flow field facilitates precise semantic matching by providing an accurate and continuous mapping of pixel-level correspondences. Similar to the cost volume-based methods, the flow field-based method employs a flow refinement module, learning to refine the initial flow field based on various types of constraints, such as smoothness or the neighbourhood consensus. An abstract illustration of this line of work is in \Cref{fig:pipeline_flow}.

\begin{figure*}[h!]
	\centering
	% \vspace*{-5mm}
	\includegraphics[width=0.5\linewidth]{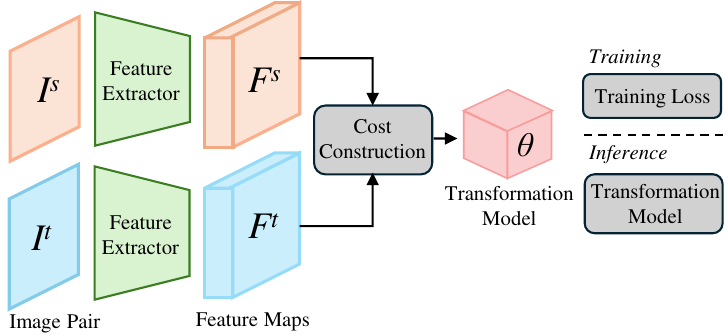}
	\vspace{-1em}
	\caption{
		\textbf{Pipeline for parameterized transformation-based methods.}
		Given a pair of images $I^s$ and $I^t$, feature extraction is performed to obtain dense feature maps  $F^s$ and $F^t$ respectively.
		The cost construction step is then applied to derive a cost volume, which is subsequently used to infer and regress the transformation parameters.}
	\label{fig:pipeline_transformation}
\vspace{-1em}
\end{figure*}
% SFNet
\xh{Several methods have been explored in this direction.}
Proposal flow~\cite{ham2016proposal, ham2018proposal} first introduces object proposals to establish region correspondence. It generates semantic flow between similar images by leveraging both local and geometric consistency constraints among object proposals and demonstrates that proposal flow can effectively be transformed into a dense flow field. 
SFNet~\cite{SFNet} proposes using binary foreground masks as a supervisory signal to establish object-aware semantic flow. By incorporating mask-flow consistency and smoothness terms, SFNet is trained end-to-end to establish object-aware correspondences while filtering distracting details. 
% GLU-Net
GLU-Net~\cite{GLU-Net} integrates global and local correlation layers to effectively manage both large and small displacements. The network processes low-resolution images for global correlation, capturing long-range correspondences, and then refines the flow estimation on high-resolution images through local correlations. 
% ACTR
ACTR~\cite{ACTR} fuses the multi-path coarse flows for refinement to benefit from different matching tensors.
% LPMFlow
LPMFlow~\cite{LPMFlow} refines flow estimation by integrating multi-scale matching flows. Specifically, flows are progressively refined using Swin-attention across various window sizes, enabling the final high-resolution flow to accurately capture fine local details, thereby incorporating correspondence in different ranges to distinguish subtle differences in narrowed pixel regions.

%summary
Due to the similarity between cost aggregation methods, flow field-based methods generally adopt similar network meta-architectures, such as multi-scale exploration and feature fusion, integrating information across scales and domains to enhance the accuracy of
semantic correspondence.

\noindent\textbf{Parameterized Transformation.} 
% introduction
\xh{Apart from cost volume and flow field, a third option is to represent the dense matching as the global warping between two images using a transformation model, such as affine, homography, or thin-plate spline. The benefit of using the transformation model is that it may automatically filter outliers during the model regression process, as the resultant transformation is intrinsically smooth and continuous. \Cref{fig:pipeline_transformation} illustrates the general pipeline of this approach.}

% CNNGeo
\xh{One representative work in this direction is CNNGeo~\cite{CNNGeometric}. It regresses the transformation between images using geometric models such as an affine and thin plate spline (TPS) transformation through a CNN, mimicking the traditional matching pipeline. Once obtaining the parameters of the transformation, the correspondence of a query point can be directly established by applying the transformation to that point.}
% CNNGeo improvements
\xh{Seo~\etal~\cite{paul2018attentive} improved this approach using offset-aware correlation kernels to put more attention to reliable matches to reduce the influence of outliers when regressing the transformation.}
% WeakAlign
\xh{WeakAlign~\cite{WeakAlign} extends this approach by proposing a weakly-supervised training method without using the ground-truth label. In addition, it proposes a soft inlier-scoring module to reduce the influence of the background clutter when estimating the transformation between objects.}

% DCTM
\xh{Unlike previous methods where the transformation between two images is summarized by one set of parameters, DCTM~\cite{DCTM} employs an affine transformation field, where each pixel has its own affine transformation, to describe the relationship between two images, greatly improving the flexibility of the model on the non-rigid deformation. It then alternatively optimizes the parameters on discrete and continuous space.}
% RTN
\xh{RTN~\cite{RTNs} follows suit, but chooses to use a recurrent neural network to gradually refine the transformation field.}
% In~\cite{RTNs, SAM-Net, SAOLD}, locally-varying affine transformation fields are iteratively estimated within locally constrained cost volume.
% PARN
\xh{PARN~\cite{PARN} constructs a network in a hierarchical nature. The bottom layer regresses one set of parameters for the global transformation between images. Going deeper down the network, each layer estimates a transformation field with a finer scale and the pixel-level transformation field is estimated in the last layer of the network. This allows the model to refine the field in a coarse-to-fine manner.}

% summary
\xh{The smooth and continuous nature of the geometric transformation model enables high-quality object warping if the parameters are accurately estimated.
However, these methods are easily distracted by background clutter and occlusion, as the regression of the transformation depends on correlations between feature pairs. Noise and outliers could lead to sub-optimal parameter regression, which could lead to inaccurate matching on a global scale. The smooth and continuous nature amplifies the error in this case.
Although this can be alleviated by using the attention module~\cite{paul2018attentive} or suppressing outlier matches~\cite{WeakAlign}, the problem still persists.}

\subsection{Training Strategy}

\xh{Apart from innovations in the neural network architecture, the other research direction is on the training strategy. The most convenient way to train a semantic matcher is supervised training: the model is trained with ground-truth correspondence labels between two images. 
However, unlike the geometric matching~\cite{LoFTR, DualRC-Net, LiFT, Sparse-NCNet}, where the dense correspondence labels can be easily obtained through multiview geometry, the correspondence labels of semantic matching rely on manual efforts. Despite several semantic matching datasets having been released, the scale of the data and the density of the labels remain relatively small and sparse. This hinders the learning of an accurate and generalizable semantic matcher.}

This prompts the research to reduce the training's reliance on ground-truth keypoint labels. This line of work can be categorized into three classes: non-keypoint-based methods, consistency-based methods, and pseudo-label generation-based methods. 
Non-keypoint-based methods explore training the model with non-keypoint labels, such as segmentation masks or images themselves. Consistency-based methods leverage the warping consistency between images to supervise the training, and pseudo-label generation-based methods generate pseudo labels based on existing sparse labels. 
We introduce each family of methods in the following paragraphs.

\subsubsection{Non-Keypoint Label} 
\xh{An intuitive way to address the reliance on keypoint labels is to use non-keypoint labels to train models. Several options have been explored.} 
Object proposals are originally developed for object detection, where they help reduce the search space and minimize false alarms. 
Proposal flow~\cite{ham2016proposal, ham2018proposal} first introduces object proposals to generate semantic flow between similar images. 
By leveraging multi-scale object proposals, it establishes region correspondences, focusing on prominent objects and parts while suppressing background clutter and distracting scene components.
% FCSS
The Fully Convolutional Self-Similarity (FCSS)~\cite{FCSS} introduces a weakly-supervised learning approach that incorporates object location priors, such as an object bounding box containing the target object to be matched. In this way, the FCSS descriptor can be trained effectively without the need for extensive keypoint annotations. 
% NC-Net DHPF image-pairs
\xh{NC-Net~\cite{NC-Net} and Dynamic Hyperpixel Flow~\cite{DynamicHyperpixelFlow} employ image-level supervision. They maximize the sum of cost volume between image pairs of the same object category and minimize that between those 
 of different categories, implicitly encouraging the network to learn correspondences by distinguishing different objects.}
% SC-ImageNet large-scale ImageNet
SC-ImageNet~\cite{SC-ImageNet} uses categories information of the large-scale ImageNet ILSVRC dataset~\cite{Imagenet} for weak supervision. 
It formulates a contrastive learning task to enhance the distinction between refined cost volumes with and without matching relationship.
% RTNs class 
RTNs \cite{RTNs} introduces a weakly-supervised classification loss to estimate the geometric transformations, based on the principle that the correct transformation for a pixel should maximize the similarity of the source features and transformed target features at that pixel, while other transformations within the search window are treated as negative examples.
% SFNet mask
SFNet \cite{SFNet, SFNet2022} trains a CNN using images with binary foreground masks and synthetic geometric deformations, leveraging high-quality correspondences to segment matched objects. The binary masks serve as strong object-level priors, enabling object-aware correspondence learning without pixel-level ground truth.

% summary
These non-keypoint label-based methods employ various forms of supervision, such as object bounding boxes, image pairs, class labels, and binary masks, to guide the learning process. 
This approach enables effective learning of semantic correspondences while significantly reducing the reliance on costly manual annotations. 
% weakness 
However, most of these methods primarily use image-level loss functions that lack explicit pixel-wise or localized supervision. This coarse-grained optimization objective sometimes leads the model to converge to suboptimal local minima. 
Therefore, more advanced methods are needed to address these limitations, such as those based on consistency constraints.

\subsubsection{Cycle / Warp Consistency} 
Cycle consistency in image matching refers to the concept that if a pixel from image A undergoes a series of correspondence matching through multiple images and eventually returns to its source image, all the matches in this process should form a closed loop and the query point will eventually return to its original position. 
This concept has been adopted by various semantic correspondence methods to train models without explicit keypoint label supervision.

% FlowWeb
FlowWeb~\cite{FlowWeb} models an image collection as a fully connected graph, where nodes represent images and edges denote pairwise correspondence flow fields. By enforcing cycle consistency across all edges, it ensures that the cumulative displacement along any closed loop is zero, establishing globally consistent dense correspondences. 
% FCSS
FCSS \cite{FCSS} enforces correspondence between a source image and a target image by verifying that the matching relationship remains consistent when reversed. 
% This approach selects valid pixel matches as positive samples by restricting correspondence candidates based on object location priors such as bounding boxes.

%  2D Point Set Registration
Laskar~\etal~\cite{laskar2019semantic} adopted a similar approach, formulating the semantic correspondence task as a 2D point set registration problem. By enforcing cycle consistency in both forward and backward transformations, they ensured that keypoints from the source image, when projected onto the target image and back, return to their original positions. 
Similarly, Chen~\etal~\cite{WeakMatchNet} introduced forward-backward consistency as a loss function, enforcing consistency by minimizing the discrepancy between the forward and backward transformation paths.
Shtedritski~\etal~\cite{shtedritski2023learning} combined ViT-DINO with cycle consistency to derive complementary learning signals, distilling a high-quality point matcher. This integration leverages the generalization ability of ViT-DINO with the spatial precision of cycle consistency.
% GLU-Net
GLU-Net~\cite{GLU-Net} incorporates cycle consistency as a post-processing step to filter the output of the global correlation, ensuring reciprocal matching. 

\xh{Instead of relying on the consistency between two images, some methods create a longer consistency path by image augmentation.}
%  WarpC 
WarpC~\cite{WarpC} introduces warp consistency, a weakly-supervised objective for dense flow regression. It enforces flow constraints using a third image, generated by randomly warping one of the original pairs, forming an image triplet to capture flow consistency. 
Furthermore, PWarpC~\cite{PWarpC} extends the warp-supervision constraint introduced in WarpC into a probabilistic form, enhancing robustness and performance. 
% 3d-guided cycle consistency
Zhou~\etal~\cite{3D-guided} incorporated a 3D CAD model to enforce cycle consistency between synthetic and real images. This method forms a 4-cycle flow field, where correspondences between real images and rendered ones act as a supervisory signal for training.
% CycleGAN
CycleGAN~\cite{CycleGAN}, similar to Zhou~\etal~\cite{3D-guided}, combines cycle consistency loss with adversarial loss to learn transformations between source and target domains without requiring paired samples.

% summary
Consistency-based methods offer implicit supervision on local correspondence matching, improving accuracy and smoothness of dense matching between two images when compared with plain image-level supervision introduced in the previous section. However, without keypoint labels, they still underperform supervised methods in terms of correspondence precision. This inspires the pseudo-label generation-based methods that enhance the training signal based on sparse keypoint labels.

\subsubsection{Pseudo-Label Generation}
% pseudo-label
\xh{The majority of methods in~\Cref{sec:arch_improve} are supervised paradigms trained directly on sparse keypoint annotations provided by the dataset. While these methods excel in metrics that measure the precision of sparse keypoint localization, they may suffer from generalization issues due to overfitting to the specific keypoint positions of the training data. To address this limitation, pseudo-label generation methods expand supervision beyond these specific locations by incorporating pseudo labels. These pseudo labels can take various forms, such as cost volumes or flow maps, serving as complementary signals to sparse ground-truth labels. By diversifying the supervision region, they enhance model accuracy and robustness.}

% PMD 
PMD~\cite{PMD} proposes a probabilistic teacher-student framework to perform knowledge distillation. 
It generates pseudo semantic flows on unlabeled real image pairs using a teacher model trained on synthetic data, and then trains a student model with its predictions. 
It demonstrates that the knowledge for semantic correspondence learned from synthetic data can be transferred to real data through this distillation process.
% SCorrSAN
Unlike PMD~\cite{PMD}, SCorrSAN \cite{SCorrSAN} directly learns dense flows from real image pairs labeled with sparse keypoints. It generates dense pseudo-labels for unlabeled regions by expanding the sparse annotations to surrounding regions using spatial-smoothness assumptions, then iteratively selects the most reliable predictions following the small-loss principle \cite{han2018co}.
% MatchMe 
MatchMe \cite{MatchMe} argues that previous approaches attempting to densify keypoints for training may not effectively address the data-hungry nature of these tasks. 
To overcome this limitation, MatchMe \cite{MatchMe} proposes to leverage the rich semantic information available in large corpora of unlabeled image pairs. 
It first trains a teacher model using the labeled data, and this teacher then generates pseudo-correspondences for unlabeled image pairs. The student model is subsequently trained using pseudo-labels as supervision. To progressively enhance label quality, it iteratively updates the teacher with the refined student model through multiple training cycles.
\rev{DIY-SC~\cite{DIY-SC} introduces a self-training pipeline that improves label quality via 3D-aware chaining, relaxed cyclic consistency, and spherical prototype filtering—without requiring any keypoint annotations. A lightweight adapter is then trained on these pseudo-labels to refine features for robust semantic matching.}
%!
\vspace{-2em}
\begin{center}
    \begin{table*}
        \centering
        \caption{Benchmark datasets for semantic correspondence. ``Source datasets'' represent original datasets where images of the dataset are sampled from. 
        % "Annotations" refer to types of annotations the dataset contain.
        }
        \vspace{-5pt}
        \renewcommand{\arraystretch}{1.2} % Increase row spacing
        \scalebox{0.92}{
        \begin{tabular} {m{0.1\textwidth} |c|c|c| m{0.2\textwidth}|m{0.2\textwidth}|m{0.35\textwidth}}
            \toprule
            \noalign{\vskip 0.15cm}
             Dataset Name & Year & Pairs & Class & Source datasets & Annotations & Characteristics \\
            \noalign{\vskip 0.15cm}
            \midrule\midrule
             Caltech-101~\cite{DSP2013} & 2006 & 1,515 &  101  &  Caltech-101~\cite{fei2004learning,li2006one}  & object segmentation & tightly cropped images of objects, little background \\
             \midrule
             PASCAL-PARTS~\cite{FlowWeb} & 2015 & 4700 &   20 &   PASCAL-PARTS~\cite{chen_cvpr14} &  keypoints (0$\sim$12), azimuth, elevation, cyclo-rotation, body part segmentation & tightly cropped images of objects, little background, part and 3D information \\
             \midrule
             TSS~\cite{taniai2016joint} & 2016 & 400 &   9  & FG3DCar, JODS,\newline PASCAL~\cite{lin2014jointly,rubinstein2013unsupervised,hariharan2011semantic} &  flows and foreground masks for image pair & dense flow field annotations obtained by interpolating sparse keypoint matches with additional co-segmentation masks \\
             \midrule
             \midrule
             PF-WILLOW~\cite{ham2018proposal} & 2017 & 900 & 5 &  PASCAL VOC 2007,\newline Caltech-256~\cite{everingham2015pascal, griffin2007caltech,cho2013learning} &  keypoints (10)  & center-aligned images, pairs with the same viewpoint \\
             \midrule
             PF-PASCAL~\cite{ham2018proposal} & 2017 & 1,300 & 20 & PASCAL VOC 2007~\cite{everingham2015pascal} &  keypoints (4$\sim$17), bbox. & pairs with the same viewpoint  \\
             \midrule
             SPair-71k~\cite{SPair-71k, HyperpixelFlow} & 2019 &  72,758 & 18 & PASCAL3D+,\newline PASCAL VOC 2012~\cite{everingham2015pascal,Xiang2014BeyondPA} & keypoints (3$\sim$30), azimuth, view-point diff., scale diff., trunc. diff., occl. diff., object seg., bbox. & large-scale data with diverse variations, rich annotations, clear dataset splits  \\
             \midrule
             AP-10K~\cite{AP-10K} & 2021 & ~29k & 47 &  public datasets focusing on animals~\cite{xian2018zero,africanWildLife,wcats,animal5,animalDCP,animals10,IUCN,endangeredanimals} & keypoints (3$\sim$17), bbox. & challenging wild environments and substantial data across multiple supervision settings \\
            %  \midrule
            %  \midrule
            %  MISC210K~\cite{MISC210K} & 2022 & 218,179 & 34 & MS COCO & keypoint (5$\sim$52), bounding box, instance mask, text description & multiple object, more challenging variations, such as mutual occlusion of multiple objects and perspective distortions in complex scenes \\
            %  \midrule
            %  SC-ImageNet~\cite{SC-ImageNet} & 2023 & 794,612 & 679 & ImageNet & 32 pseudo keypoint pairs for each image pair & can be used for model pre-training, provide pixel-level matching relationships in image pairs \\
             \bottomrule
        \end{tabular}}
        \label{tab:benchmark_datasets}
        \vspace{-1em}
    \end{table*}
\end{center}

Unlike the aforementioned approaches \cite{PMD, SCorrSAN, MatchMe, SC-ImageNet} that use flow maps as pseudo-labels, SemiMatch \cite{SemiMatch} and SC-ImageNet~\cite{SC-ImageNet} utilize the cost volume both as ground-truth correspondences and as the format for pseudo-labels. 
% SemiMatch
In the SemiMatch \cite{SemiMatch} framework, given source image $I^s$ and target image $I^t$, the target image is first augmented into a weakly-augmented version and a strongly-augmented version through photometric and geometric augmentations, forming a triplet of images. 
The model then predicts the cost volume between the source image and the weakly-augmented target image supervised by keypoint annotations. 
This cost volume is then transformed by applying the same geometric warp used for the strong augmentation, and sharpened to generate a pseudo-label. 
The pseudo-label is subsequently used to supervise the generation of the cost volume in the strong-augmented branch.
% SC-ImageNet
Huang \etal~\cite{SC-ImageNet} proposed learning correspondences from image-level annotations and created a new dataset called Semantic Correspondence ImageNet (SC-ImageNet), which contains high-quality pseudo-labels for semantic correspondence.

% summary
These pseudo-label generation-based methods demonstrate unique advantages in semantic correspondence tasks by combining sparse keypoint annotations with generated pseudo-labels. Through the complementary signals provided by pseudo-labels, some of these approaches outperform purely supervised methods in certain scenarios. 
However, these methods have not completely eliminated the dependency on manual annotations—model training still requires high-quality sparse keypoint annotations as initial conditions. Moreover, the quality of pseudo-labels heavily relies on the accuracy of the initial supervisory signals, which may lead to error propagation issues. 
Despite these limitations, pseudo-label generation-based methods represent a research direction that has achieved an effective balance between reducing annotation dependency and enhancing model performance. 
\section{Datasets and Evaluation Metrics}\label{sec:Datasets}
This section reviews benchmark datasets and evaluation metrics for semantic correspondence. We first present prevalent datasets spanning from early small-scale collections to modern large-scale benchmarks, detailing their annotation formats and applications. 
Subsequently, we formalize the standard Percentage of Correct Keypoints (PCK) metric with its dataset-specific variants, clarifying its computation methods and practical implementations across different datasets. 

\subsection{Datasets for Semantic Correspondence}
\xh{Many datasets have been released for benchmarking the semantic matching models. We summarize the year of release, number of image pairs, number of object classes, image sources, and types of annotations in~\Cref{tab:benchmark_datasets}. We also briefly introduce them below.} 

\noindent \textbf{Caltech-101}~\cite{DSP2013} dataset provides binary mask annotations of objects of interest for 1,515 image pairs, enabling evaluations of coarse-level semantic correspondence tasks.
This dataset contains images from 101 distinct object categories, each paired with a ground-truth foreground segmentation mask. 
Although initially designed for image classification, this dataset has been repurposed for evaluating semantic alignment. 
% Specifically, the predicted dense correspondences from the target to the source image are utilized to warp the source's ground-truth segmentation mask towards the target. 
% The overlap between the warped source segmentation mask and the target's ground-truth segmentation mask is then measured, serving as a proxy to assess the quality of the predicted dense correspondences.

\noindent \textbf{PASCAL-PARTS}~\cite{FlowWeb} dataset provides segmentation masks for each body part for 20 object categories. It offers detailed annotations for object components (e.g., a person's head, a car's wheel). This fine-grained supervision makes it a foundational benchmark for evaluating part-level semantic correspondence.

\noindent \textbf{TSS}~\cite{taniai2016joint} dataset uniquely provides dense flow field annotations for the foreground objects in its image pairs. 
It contains 400 image pairs of 7 object categories, divided into three groups based on their source datasets: FG3DCAR \cite{lin2014jointly}, JODS \cite{rubinstein2013unsupervised}, and PASCAL~\cite{hariharan2011semantic}.

\noindent \textbf{PF-WILLOW}~\cite{ham2018proposal} dataset consists of 900 image pairs of five object classes: face, car, motorbike, duck, and wine bottle. 
Each image is annotated with 10 keypoints, making this dataset well-suited for evaluating semantic correspondence methods, particularly those focused on keypoint-level matching between similar object instances.

\noindent \textbf{PF-PASCAL}~\cite{ham2018proposal} dataset consists of 1,300 image pairs drawn from the PASCAL VOC dataset~\cite{everingham2015pascal}, covering 20 object classes.
Each image is annotated with between 4 and 17 keypoints, as well as bounding boxes.
% Compared to other datasets~\cite{ham2016proposal,taniai2016joint}, PF-PASCAL poses a greater challenge for semantic correspondence evaluation due to its inclusion of distinct object instances within the same class, accompanied by substantial variations in appearance, scene layout, clutter, and object scale.

\noindent \textbf{SPair-71k}~\cite{SPair-71k, HyperpixelFlow} is a large-scale and comprehensive benchmark dataset for semantic correspondence, consisting of 70,958 image pairs across 18 object categories and encompassing a broad range of viewpoint and scale variations.
Compared to earlier datasets, SPair-71k~\cite{SPair-71k} notably increases the number of image pairs and provides more accurate and detailed annotations. 
The annotations capture variations in viewpoint, scale, truncation, and occlusion. 
% As a result, this extensive dataset offers a robust testbed for investigating semantic correspondence and is expected to drive further advancements in the field.

\noindent \textbf{AP-10K}~\cite{AP-10K} was originally designed for animal pose estimation, consisting of 10,015 images spanning 23 families and 54 species. 
Zhang~\etal \cite{GeoAware-SC} adapted this dataset into a challenging semantic correspondence benchmark, resulting in 260,950 training pairs, 20,630 testing pairs, and 8,816 validation pairs.
The validation and testing subsets are organized into three distinct settings: an intra-species set, a cross-species set, and a cross-family set.
Overall, this dataset is approximately five times larger than SPair-71k~\cite{SPair-71k, HyperpixelFlow}, offering a broader and more complex evaluation environment.

\begin{table*}[!t]
    \begin{center}
    \caption{Performance comparison of supervised methods on SPair-71k, PF-PASCAL, and PF-WILLOW datasets. The methods are categorized into two groups based on the backbone used: CNN (e.g., ResNet-101) and foundation model (e.g., DINO, SD). 
    \newline FT.: Fine-tune, ML.: Multi-layer Feature, Aug.: Data Augmentation, Reso.: Resolution.
    }
    \vspace{-10pt}
    \label{tab:1_supervised_performance}
    \renewcommand{\arraystretch}{1.2} % 行距
    \scalebox{0.9}{
    \begin{tabular}{
         l
        |l
        |>{\centering\arraybackslash}p{0.5cm}
        |>{\centering\arraybackslash}p{0.5cm}
        |>{\centering\arraybackslash}p{0.5cm}
        |>{\centering\arraybackslash}p{0.8cm}
        |ccc|ccc|ccc|ccc}
        \toprule
        \multirow{3}{*}{Methods} & \multirow{3}{*}{Backbone} & 
        \multirow{3}{*}{\shortstack{FT.}} & \multirow{3}{*}{ML.} & \multirow{3}{*}{\shortstack{Aug.}}
        & \multirow{3}{*}{Reso.} & 
        \multicolumn{3}{c|}{SPair-71k} & \multicolumn{3}{c|}{PF-PASCAL} & 
        \multicolumn{6}{c}{PF-WILLOW} \\
        & & & & & & \multicolumn{3}{c|}{PCK @ $\alpha_{\text{bbox}}$}
                  & \multicolumn{3}{c|}{PCK @ $\alpha_{\text{img}}$}
                  & \multicolumn{3}{c}{PCK @ $\alpha_{\text{bbox}}$}
                  & \multicolumn{3}{c}{PCK @ $\alpha_{\text{bbox-kp}}$} \\
        & & & & & & 0.05 & 0.1 & 0.15 & 0.05 & 0.1 & 0.15 & 0.05 & 0.1 & 0.15 & 0.05 & 0.1 & 0.15\\
        \midrule % 水平线
        SCNet~\cite{SCNet} & VGG16 & $\times$ & $\times$ & $\times$ & 224 & - & - & - & 36.2 & 72.2 & 82.0 & 38.6 & 70.4 & 85.3 & - & - & - \\
        NC-Net~\cite{NC-Net} & ResNet-101 & \checkmark & $\times$ & $\times$ & - & - & 20.1 & - & 54.3 & 78.9 & 86.0 & - & - & - & 33.8 & 67.0 & - \\
        CNNGeo~\cite{CNNGeometric} & ResNet-101 & $\times$ & $\times$ & $\times$ & 227 & - & 20.6 & - & 41.0 & 69.5 & 80.4 & - & - & - & 36.9 & 69.2 & - \\
        WeakAlign~\cite{WeakAlign} & ResNet-101 & $\times$ & $\times$ & $\times$ & - & - & 20.9 & - & 49.0 & 74.8 & 84.0 & - & - & - & 37.0 & 70.2 & - \\
        A2Net~\cite{paul2018attentive} & ResNet-101 & $\times$ & $\times$ & $\times$ & - & - & 22.3 & - & 42.8 & 70.8 & 83.3 & - & - & - & 36.3 & 68.8 & - \\
        HPF~\cite{HyperpixelFlow} & ResNet-101 & $\times$ & \checkmark & $\times$ & - & - & 28.2 & - & 60.1 & 84.8 & 92.7 & 45.9 & 74.4 & 85.6 & 45.9 & 74.4 & 85.6 \\
        ANC-Net~\cite{ANC-Net} & ResNeXt-101 & $\times$ & $\times$ & $\times$ & - & - & 30.1 & - & - & 88.7 & - & - & - & - & - & - & - \\
        DHPF~\cite{DynamicHyperpixelFlow} & ResNet-101 & $\times$ & \checkmark & $\times$ & 240 & 20.9 & 37.3 & 47.5 & 75.7 & 90.7 & 95.0 & 49.5 & 77.6 & 89.1 & - & 71.0 & - \\

        PMD~\cite{PMD} & ResNet-101 & \checkmark & - & - & - & - & 37.4 & - & - & 90.7 & - & - & 75.6 & - & - & - & - \\
        CHM~\cite{CHM} & ResNet-101 & \checkmark & $\times$ & $\times$ & 240 & 27.2 & 46.3 & 57.5 & 80.1 & 91.6 & 94.9 & 52.7 & 79.4 & 87.5 & - & 69.6 & - \\
        CATs~\cite{CATs} & ResNet-101 & \checkmark & \checkmark & \checkmark & 256 & 27.7 & 49.9 & 61.7 & 75.4 & 92.6 & 96.4 & 50.3 & 79.2 & 90.3 & 40.7 & 69.0 & - \\
        PMNC~\cite{PMNC} & ResNet-101 & \checkmark & \checkmark & $\times$ & 400 & - & 50.4 & - & 82.4 & 90.6 & - & - & - & - & - & - & - \\
        MMNet-FCN~\cite{MMNet} & ResNet-101 & \checkmark & \checkmark & $\times$ & 224 & 33.3 & 50.4 & 61.2 & 81.1 & 91.6 & 95.9 & - & - & - & - & - & - \\
        SemiMatch~\cite{SemiMatch} & ResNet-101 & \checkmark & - & \checkmark & 256 & - & 50.7 & - & 80.1 & 93.5 & 96.6 & 54.0 & 82.1 & 92.1 & - & - & - \\
        CHMNet~\cite{CHMNet} & ResNet-101 & \checkmark & \checkmark & \checkmark & 240 & - & 51.3 & - & 83.1 & 92.9 & - & 53.8 & 79.3 & - & - & 69.3 & - \\
        PWarpC-NC-Net~\cite{PWarpC} & ResNet-101 & - & - & - & Ori & 31.6 & 52.0 & - & 67.8 & 82.3 & 86.9 & - & - & - & 46.1 & 72.6 & 82.7 \\
        NeMF~\cite{NeMF} & ResNet-101 & - & - & - & Ori & 34.2 & 53.6 & - & 80.6 & 93.6 & - & - & - & - & 60.8 & 75.0 & - \\
        TransforMatcher~\cite{TransforMatcher} & ResNet-101 & \checkmark & \checkmark & \checkmark & 240 & 32.4 & 53.7 & - & 80.8 & 91.8 & - & - & 76.0 & - & - & 65.3 & - \\
        HCCNet~\cite{HCCNet} & ResNet-101 & \checkmark & \checkmark & \checkmark & 240 & 35.8 & 54.8 & - & 80.2 & 92.4 & - & - & 74.5 & - & - & 65.5 & - \\
        SCorrSAN~\cite{SCorrSAN} & ResNet-101 & \checkmark & $\times$ & \checkmark & 256 & - & 55.3 & - & 81.5 & 93.3 & 96.6 & 54.1 & 80.0 & 89.8 & - & - & - \\
        VAT~\cite{VAT} & ResNet-101 & \checkmark & \checkmark & \checkmark & - & 35.0 & 55.5 & 65.1 & 78.2 & 92.3 & 96.2 & 52.8 & 81.6 & 91.4 & 42.3 & 71.3 & - \\
        KBCNet~\cite{KBCNet} & ResNet-101 & \checkmark & - & - & 256 & 39.1 & 59.1 & 68.2 & 78.1 & 93.8 & 97.2 & 56.4 & 84.7 & 93.5 & - & - & - \\
        CATs++~\cite{CATs++} & ResNet-101 & \checkmark & \checkmark & \checkmark & 512 & 40.7 & 59.8 & 68.5 & 84.9 & 93.8 & 96.8 & 56.7 & 81.2 & - & 47.0 & 72.6 & - \\
        MatchMe-CATs++ \cite{MatchMe} & ResNet-101 & \checkmark & \checkmark & \checkmark & - & - & 62.0 & - & 84.9 & 94.3 & 96.7 & 59.6 & 83.6 & 92.9 & - & - & - \\
        UFC~\cite{UFC2024} & ResNet-101 & $\times$ & \checkmark & \checkmark & Ori & 48.5 & 64.4 & 72.1 & 88.0 & 94.8 & 97.9 & 58.6 & 81.2 & - & 50.4 & 74.2 & - \\
        \midrule 
        DINO-ViT~\cite{DINO-ViT} & DINO-ViT & - & \checkmark & \checkmark & - & - & 61.4 & - & - & - & - & - & - & - & - & - & - \\
        ACTR~\cite{ACTR} & iBOT & - & $\times$ & - & - & 42.0 & 62.1 & - & 81.2 & 94.0 & 97.0 & - & 87.2 & - & - & 79.9 & - \\
        SimSC-iBOT~\cite{SimSC} & iBOT & \checkmark & $\times$ & $\times$ & 256 & 43.0 & 63.5 & - &  88.4 & 95.6 & 97.3 & - & - & - & 44.9 & 71.4 & 84.5 \\
        DHF~\cite{Diffusion-Hyperfeatures} & SD+DINOv2 & \checkmark & \checkmark & - & Ori & - & 64.6 & - & - & 86.7 & - & - & 78.0 & - & - & - & - \\
        LPMFlow~\cite{LPMFlow} & ViT-B/16 & \checkmark & \checkmark & - & 256 & 46.7 & 65.6 & - & 82.4 & 94.3 & 97.2 & - & 87.6 & - & - & 81.0 & - \\
        % SD-DINO~\cite{A-Tale-of-Two-Features} & SD+DINOv2 & \checkmark & \checkmark & - & 960, 840  & - & 74.6 & - & 80.9 & 93.6 & 96.9 & - & - & - & - & - & - \\
        SD4Match~\cite{SD4Match} & SD+DINOv2 & $\times$ & $\times$ & $\times$ & 768 & 59.5 & 75.5 & - & 84.4 & 95.2 & 97.5 & - & - & - & 52.1 & 80.4 & 91.2 \\
        GeoAware-SC~\cite{GeoAware-SC} & SD+DINOv2 & - & \checkmark & $\times$ & {\fontsize{6pt}{10pt}\selectfont 960, 840} & 72.6 & 82.9 & - & 85.5 &  95.1 & 97.4 & - & - & - & - & - & - \\
        \bottomrule
    \end{tabular}
    }
    \end{center}
    \vspace{-15pt}
\end{table*}
\begin{table*}[!t]
    \begin{center}
    \caption{Performance comparison of weakly supervised methods on SPair-71k, PF-PASCAL, and PF-WILLOW datasets.
    \newline FT.: Fine-tune, ML.: Multi-layer Feature, Aug.: Data Augmentation, Reso.: Resolution.}
    \vspace{-10pt}
    \label{tab:2_unsupervised_performance}
    \renewcommand{\arraystretch}{1.2} % 行距
    \scalebox{0.9}{
    \begin{tabular}{l|l
        |>{\centering\arraybackslash}p{0.5cm}
        |>{\centering\arraybackslash}p{0.5cm}
        |>{\centering\arraybackslash}p{0.5cm}
        |>{\centering\arraybackslash}p{0.6cm}
        |ccc|ccc|ccc|ccc}
        \toprule
        \multirow{3}{*}{Methods} & \multirow{3}{*}{Backbone} & 
        \multirow{3}{*}{\shortstack{FT.}} & \multirow{3}{*}{ML.} & \multirow{3}{*}{\shortstack{Aug.}}
        & \multirow{3}{*}{Reso.} & 
        \multicolumn{3}{c|}{SPair-71k} & \multicolumn{3}{c|}{PF-PASCAL} & 
        \multicolumn{6}{c}{PF-WILLOW} \\
        & & & & & & \multicolumn{3}{c|}{PCK @ $\alpha_{\text{bbox}}$}
                  & \multicolumn{3}{c|}{PCK @ $\alpha_{\text{img}}$}
                  & \multicolumn{3}{c}{PCK @ $\alpha_{\text{bbox}}$}
                  & \multicolumn{3}{c}{PCK @ $\alpha_{\text{bbox-kp}}$} \\
        & & & & & & 0.05 & 0.1 & 0.15 & 0.05 & 0.1 & 0.15 & 0.05 & 0.1 & 0.15 & 0.05 & 0.1 & 0.15\\
        \midrule % 水平线
        Semantic-GLU-Net \cite{GLU-Net} & VGG-16 & - & \checkmark & - & - & - & 14.3 & - & 46.0 & 70.6 & - & - & - & - & - & - & - \\
        SAM-Net \cite{SAM-Net} & VGG-19 & \checkmark & - & - & - & - & - & - & 60.1 & 80.2 & 86.9 & - & - & - & - & - & - \\
        WarpC \cite{WarpC} & VGG-16 & $\times$ & \checkmark & \checkmark & - & - & 23.5 & - & 62.1 & 81.7 & 89.7 & 49.0 & 75.1 & 86.9 & - & - & - \\
        \midrule
        PARN \cite{PARN} & ResNet-101 & \checkmark & \checkmark & \checkmark & - & - & - & - & 26.8 & 49.1 & 66.2 & - & - & - & - & - & - \\
        WeakMatchNet \cite{WeakMatchNet} & ResNet-101 & \checkmark & $\times$ & \checkmark & - & - & - & - & - & 78.0 & - & 49.1 & 81.9 & 92.2 & - & - & - \\
        DCCNet \cite{DCCNet} & ResNet-101 & $\times$ & $\times$ & $\times$ & 400 & - & - & - & 55.6 & 82.3 & 90.5 & 43.6 & 73.8 & 86.5 & - & - & - \\
        CNNGeo \cite{CNNGeometric} & ResNet-101 & - & $\times$ & - & - & - & 18.1 & - & - & 71.9 & - & 36.9 & 69.2 & 77.8 & - & - & - \\
        A2Net \cite{paul2018attentive} & ResNet-101 & - & $\times$ & - & - & - & 20.1 & - & 42.8 & 70.8 & 83.3 & 36.3 & 68.8 & 84.4 & - & - & - \\
        WeakAlign \cite{WeakAlign} & ResNet-101 & \checkmark & $\times$ & \checkmark & - & - & 21.1 & - & - & 75.8 & - & 37.0 & 70.2 & 79.9 & - & - & - \\
        RTNs \cite{RTNs} & ResNet & $\times$ & $\times$ & $\times$ & - & - & 25.7 & - & 55.2 & 75.9 & 85.2 & - & - & - & 41.3 & 71.9 & 86.2 \\
        SFNet \cite{SFNet, SFNet2022} & ResNet-101 & $\times$ & \checkmark & $\times$ & - & - & 26.3 & - & 53.6 & 81.9 & 90.6 & - & 73.5 & - & 46.3 & 74.0 & - \\
        NC-Net \cite{NC-Net} & ResNet-101 & \checkmark & $\times$ & $\times$ & 400 & - & 26.4 & - & 54.3 & 78.9 & 86.0 & 44.0 & 72.7 & 85.4 & - & - & - \\
        PMD \cite{PMD} & ResNet-101 & \checkmark & - & - & 320 & - & 26.5 & - & - & 81.2 & - & - & 74.7 & - & - & - & - \\
        DHPF \cite{DynamicHyperpixelFlow} & ResNet-101 & $\times$ & \checkmark & $\times$ & 240 & - & 27.7 & - & 56.1 & 82.1 & 91.1 & 50.2 & 80.2 & 91.1 & - & - & - \\
        ASYM \cite{ASYM} & ResNet-50 & $\times$ & - & - & - & - & 34.0 & - & - & - & - & - & - & - & - & - & - \\
        GSF \cite{GSF} & ResNet-101 & $\times$ & \checkmark & $\times$ & - & - & 36.1 & - & 65.6 & 87.8 & 95.9 & 47.0 & 75.8 & 88.9 & 49.1 & 78.7 & - \\
        PWarpC-NC-Net \cite{PWarpC} & ResNet-101 & \checkmark & \checkmark & \checkmark & Ori & 18.5 & 38.0 & - & 61.7 & 82.6 & 88.5 & - & - & - & 43.6 & 74.6 & 86.9 \\
        \midrule
        ASIC \cite{gupta2023asic} & ViT-S/8 & $\times$ & - & - & - & - & 36.9 & - & - & - & - & 53.0 & 76.3 & - & - & - & - \\
        SC-ImageNet \cite{SC-ImageNet} & iBOT-B & $\times$ & - & - & - & - & 60.3 & - & 80.5 & 93.0 & 96.7 & 57.1 & 85.1 & 94.1 & - & - & - \\
        % DistillDIFT \cite{DistillDIFT} & - & - & - & - & - & - & - & - & - & - & - & - & - & - & - & - & - \\
        \bottomrule
    \end{tabular}
    }
    \end{center}
    \vspace{-15pt}
\end{table*}

\begin{table*}[!t] 
    \begin{center}
    \caption{Performance comparison of zero-shot methods on SPair-71k, PF-PASCAL, and PF-WILLOW datasets.}
    \vspace{-10pt}
    \label{tab:3_zero_shot_performance}
    \renewcommand{\arraystretch}{1.2} % 行距
    \scalebox{0.95}{
    \begin{tabular}{l|l|>{\centering\arraybackslash}p{1.5cm}|ccc|ccc|ccc|ccc}
        \toprule
        \multirow{3}{*}{Methods} & \multirow{3}{*}{Backbone} & \multirow{3}{*}{Reso.} & 
        \multicolumn{3}{c|}{SPair-71k} & \multicolumn{3}{c|}{PF-PASCAL} & 
        \multicolumn{6}{c}{PF-WILLOW} \\
        & & & \multicolumn{3}{c|}{PCK @ $\alpha_{\text{bbox}}$}
            & \multicolumn{3}{c|}{PCK @ $\alpha_{\text{img}}$}
            & \multicolumn{3}{c}{PCK @ $\alpha_{\text{bbox}}$}
            & \multicolumn{3}{c}{PCK @ $\alpha_{\text{bbox-kp}}$} \\
        & & & 0.05 & 0.1 & 0.15 & 0.05 & 0.1 & 0.15 & 0.05 & 0.1 & 0.15 & 0.05 & 0.1 & 0.15 \\
        \midrule % 水平线
        SCOT \cite{SCOT} & ResNet-101 & Max 300 & - & 35.6 & - & 63.1 & 85.4 & 92.7 & 47.8 & 76.0 & 87.1 & 47.8 & 76.0 & - \\
        DINO+LiFT \cite{LiFT} & DINO(ViT-S/16) & - & 14.7 & 28.7 & - & - & - & - & - & - & - & - & - & - \\
        LDM \cite{LDM_correspondences} & SD & - & 28.9 & 45.4 & - & - & - & - & 53.0 & 84.3 & - & - & - & - \\
        DIFTsd \cite{DIFT} & SD & 768 & - & 52.9 & - & - & - & - & - & - & - & - & - & - \\
        SD-DINO \cite{A-Tale-of-Two-Features} & SD+DINOv2 & 960, 840\tnote{1}
        & - & 59.3 & - & 73.0 & 86.1 & 91.1 & - & - & - & - & - & - \\
        GeoAware-SC \cite{GeoAware-SC} & SD+DINOv2 & 960, 840 & 45.3 & 61.3 & - & 74.0 & 86.2 & 90.7 & - & - & - & - & - & - \\
        \bottomrule
        % \textbf{Ours} & SD+DINOv2 & 960, 840 & \textbf{47.9} & \{60.4} & \textbf{67.0} & 72.9 & \textbf{88.3} & \textbf{93.4} & - & - & - & \{46.6} & \{72.8} & \textbf{85.8}  \\
        % \midrule
    \end{tabular}
    }
    \begin{tablenotes}
        \item[1] 960, 840 represent image resolutions: 960$\times$960 for SD and 840$\times$840 for DINOv2, respectively.
    \end{tablenotes}
    \end{center}
    \vspace{-15pt}
\end{table*}

\subsection{Evaluation Metrics}
% PCK
The Percentage of Correct Keypoints (PCK) metric serves as a standard measure for evaluating the performance of keypoint matching in semantic correspondence tasks. Given an image pair $(I^s, I^t)$ and its associated correspondence set $\mathcal{X} = \{(\mathbf{x}^s_q, \mathbf{x}^t_q) \mid q = 1, 2, \dots, n\}$, where $\mathbf{x} = (x, y)$, $\mathbf{x}^s_q$ is the query point on the source image $I^s$ and $\mathbf{x}^t_q$ is the ground-truth correspondence on the target image $I^t$, we predict the correspondence $\{\bar{\mathbf{x}}^t_q \mid q = 1, 2, \dots, n\}$ and calculate PCK for the image pair by:
$$
    PCK(I^s, I^t) = \frac{1}{n} \sum_{q=1}^n \mathbb{I}(\|\bar{\mathbf{x}}^t_q - \mathbf{x}^t_q\| \leq \alpha * \theta),
$$
where $\theta$ is a base threshold, $\alpha$ is a value less than 1, and $\mathbb{I}(\cdot)$ is the binary indicator function with $\mathbb{I}(\text{true}) = 1$ and $\mathbb{I}(\text{false}) = 0$. The widely used values for $\alpha$ include 0.05, 0.1, or 0.15. \rev{For the base threshold $\theta$, there are three common types:}
\begin{itemize}
    \item \rev{\textbf{Bounding Box Size}: The base threshold is defined by the object's bounding box dimensions:
    $\theta_{\text{bbox}} = \max(h_{\text{bbox}}, w_{\text{bbox}})$,
    where $h_{\text{bbox}}$ and $w_{\text{bbox}}$ denote the height and width of the bounding box.}

    \item \rev{\textbf{Image Dimension:} The base threshold is determined by the image dimensions:
    $\theta_{\text{img}} = \max(h_{\text{img}}, w_{\text{img}})$.}

    \item \rev{\textbf{Keypoint:} The base threshold uses the keypoint-based bounding box (bbox-kp) PCK:
    $\theta_{\text{kps}} = \max(\max_q(x^t_q) - \min_q(x^t_q), \max_q(y^t_q) - \min_q(y^t_q))$,
    where the bounding box is approximated from the keypoint positions, resulting in a tighter fit around the object.} 
\end{itemize}
There are two ways to report PCK values. 
% per image vs per point
One common practice is to calculate a PCK value for each image and then average it across the dataset or across each category split. 
The other approach calculates PCK by dividing the total number of correctly predicted points over the entire dataset (or each category split) by the total number of predicted points. 
We refer to the former as ``PCK per image'' and the latter as ``PCK per point''. 
In this paper, we align with the majority of previous studies by adopting ``PCK per image'' as our evaluation metric.
\section{Benchmark of Existing Methods}
\label{sec:Benchmark}

% Performance Comparison
We compare the performance of existing methods on the SPair-71k, PF-PASCAL, and PF-WILLOW datasets and categorize them into three groups: 
supervised methods, weakly supervised methods, and zero-shot methods. 
Methods that utilize keypoint annotations are classified as supervised methods, and their performance is summarized in Table~\ref{tab:1_supervised_performance}. 
Methods that do not use keypoint annotations but use other auxiliary annotations, such as class labels, image pairs, or masks, are categorized as weakly supervised methods, and their performance is presented in Table~\ref{tab:2_unsupervised_performance}.
Finally, methods without any fine-tuning are referred to as zero-shot methods, and their performance is shown in Table~\ref{tab:3_zero_shot_performance}.

The quantitative comparison result reveals that supervised methods significantly outperform weakly supervised and zero-shot approaches, as they leverage keypoint annotations for fine-tuning. 
Within supervised methods, vision transformer foundation models (e.g., DINO) and Stable Diffusion backbones demonstrate superior performance compared to CNN backbones like ResNet. This advantage stems from transformers' global interaction capabilities and their exposure to larger-scale pre-training datasets, which provide stronger initialization. 
% Additionally, the choice of cost aggregation strategies substantially impacts final matching accuracy. 
\xh{Due to the diversity of the backbones, matching modules, and inconsistent image resolution, it is hard to conclude the most effective choice for each component in semantic matching. Therefore, we conduct extensive experiments in the next section to make fair and controlled evaluations of each stage in the matching pipeline and find out the most effective design of the model.}

% We conducted extensive experiments on some representative methods, our focus was primarily on supervised methods. 
% Given that most weakly supervised approaches exhibit suboptimal performance and the implementations of better-performing methods are not publicly available, we opted not to pursue further experiments in this category.
% Compared to PF-PASCAL and PF-WILLOW datasets, performance on the SPair-71k dataset is far from saturation. Therefore, we selected this dataset for conducting subsequent benchmark experiments, which is currently the most widely used benchmark for semantic correspondence tasks.
\section{Experiments and Analysis}
\label{sec:Experiments}

\xh{This section evaluates each stage of semantic correspondence methods through detailed experiments. We begin by evaluating commonly used feature backbones in~\Cref{sec:feature_backbone}. We then pick the most effective one to serve as the foundation and evaluate popular matching modules, including both feature enhancement (\Cref{sec:feature_enhancement}) and matching refinement (\Cref{sec:cost_aggregation}) techniques, by appending them to it under standardized settings.
Based on the experiment results, we present a strong and effective baseline in~\Cref{sec:strong_baseline} that achieves state-of-the-art performance in various benchmarks.}

\subsection{Feature Backbone Evaluation}
\label{sec:feature_backbone}

\begin{table}[t]
    \begin{center}
    \caption{Evaluation of zero-shot feature backbones on the SPair-71k dataset. Reso.: Image Resolution, Feature Map Size.}
    \vspace{-7pt}
    \label{tab:4_zero_backbone}
    \renewcommand{\arraystretch}{1.2} % Row spacing
    \scalebox{1.0}{
    \begin{tabular}{c
        |>{\centering\arraybackslash}p{1.7cm}
        |cccc}
        \toprule
        \multirow{3}{*}{Backbone} & 
        \multirow{3}{*}{Reso.} & 
        \multicolumn{4}{c}{SPair-71k} \\
        & & \multicolumn{4}{c}{PCK @ $\alpha_{\text{bbox}}$} \\
        & & 0.01 & 0.05 & 0.1 & 0.15 \\
        \midrule
        CLIP & 224, 16 & 0.5 & 9.5 & 22.8 & 33.2 \\
        \midrule
        ResNet-101 & 960, 60 & 3.3 & 18.6 & 26.5 & 32.2 \\
        ResNet-101 & 512, 32 & 1.6 & 16.5 & 26.8 & 33.1 \\
        ResNet-101 & 256, 16 & 0.5 & 10.3 & 23.0 & 31.7 \\
        \midrule
        DINOv1 & 960, 60 & 1.7 & 14.0 & 25.6 & 35.3 \\
        DINOv1 & 512, 32 & 1.4 & 15.8 & 30.2 & 40.2 \\
        DINOv1 & 256, 16 & 0.6 & 12.2 & 27.8 & 38.8 \\
        \midrule
        iBOT & 960, 60 & 2.2 & 18.7 & 33.8 & 44.9 \\
        iBOT & 512, 32 & 1.8 & 21.4 & 38.9 & 50.3 \\
        iBOT & 256, 16 & 0.9 & 16.2 & 35.1 & 47.7 \\
        \midrule
        SD2-1 & 960, 60 & \underline{7.5} & 40.0 & 50.7 & 56.7 \\
        SD2-1 & 512, 32 & 3.2 & 29.4 & 41.8 & 48.0 \\
        SD2-1 & 256, 16 & 0.5 & 9.7 & 21.0 & 28.7 \\
        \midrule
        DINOv2 & 840, 60 & 7.3 & 40.0 & 54.4 & 62.8 \\
        DINOv2 & 448, 32 & 3.7 & 35.4 & 52.9 & 62.2 \\
        DINOv2 & 224, 16 & 1.1 & 20.9 & 43.1 & 56.2 \\
        \midrule
        SD2-1+DINOv2 & (960,840), 60 & \textbf{9.7} & \textbf{47.9} & \textbf{60.4} & \textbf{67.0} \\
        SD2-1+DINOv2 & (512,448), 32 & 4.7 & \underline{40.3} & \underline{56.5} & \underline{64.1} \\
        SD2-1+DINOv2 & (256,224), 16 & 1.3 & 21.4 & 41.9 & 53.1 \\
        \bottomrule
    \end{tabular}
    }
    \end{center}
    \vspace{-15pt}
\end{table}
\begin{figure}[!t]
  \begin{center}
    \includegraphics[width=0.45\textwidth]{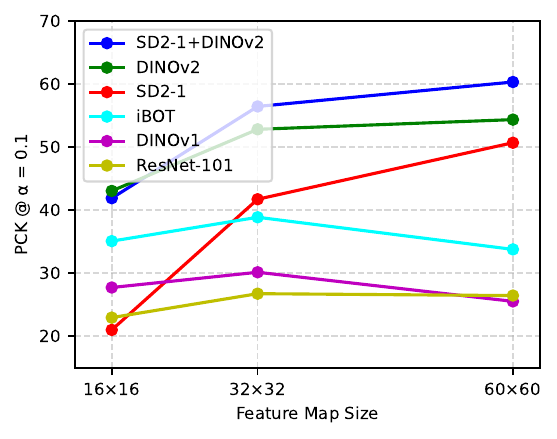}
  \end{center}
  \vspace{-3mm}
    \caption{Performance of different feature backbones on zero-shot evaluation across varying feature map sizes (16$\times$16, 32$\times$32, and 60$\times$60).} 
  \label{fig:Zero}
  \vspace{-5.0mm}
\end{figure}

Due to the absence of systematic evaluation on the performance of commonly used feature backbones, in this section, we evaluate the performance of various feature backbones in a uniform setting to find out the most effective one. We select ResNet-101 \cite{ResNet}, CLIP \cite{CLIP}, iBOT \cite{iBOT}, DINO \cite{DINO-ViT}, DINOv2 \cite{DINOv2}, and Stable Diffusion \cite{Stable-Diffusion} as our choices for the backbone.
% \rev{
% We also analyze DINOv3 \cite{DINOv3}, released very recently (Aug. 2025), with its zero-shot and fine-tuned performance detailed in Appendix A.
% Since the majority of recent methods \cite{A-Tale-of-Two-Features, GeoAware-SC, SemAlign3D} are built upon the DINOv2 backbone, we do not include DINOv3 in the main experiments to maintain fairness when comparing against other methods.}

Specifically, for ResNet-101, we truncate it before layer 4 following the literature convention~\cite{NC-Net, SimSC, ANC-Net}. 
For transformer backbones CLIP, DINO, iBOT, and DINOv2, we select CLIP-ViT-L/14, DINO-ViT-B/16, iBOT-ViT-B/16, and DINOv2-ViT-B/14 configurations, respectively. 
For Stable Diffusion-based backbone, we follow the setup in SD4Match~\cite{SD4Match} and use Stable Diffusion 2-1 as the model. 
Moreover, we evaluate the combined backbone SD+DINO~\cite{A-Tale-of-Two-Features} by concatenating the feature maps from SD2-1 and DINOv2 along the channel dimension. 
Due to the inconsistent image-to-feature ratio across different backbones, we evaluate all backbones on three commonly adopted feature map resolutions: $16 \times 16$, $32 \times 32$, and $60 \times 60$.

\subsubsection{Zero-Shot Evaluation}

We conduct zero-shot evaluations to assess the performance of different feature backbones across varying feature map sizes. 
\rev{
The results are presented in~\Cref{tab:4_zero_backbone}. 
As shown, SD2-1, DINOv2, and their combined variant (SD2-1+DINOv2) significantly outperform CLIP, ResNet-101, DINOv1, and iBOT. 
% DINOv2
Within the Transformer family, performance varies substantially with respect to pre-training scale and training objectives. 
For instance, the success of DINOv2 underscores the importance of large-scale self-supervised pre-training for learning generalizable features suitable for zero-shot evaluations.
% SD
Our experiments also demonstrate that features derived from generative models like SD2-1 are highly effective, confirming the findings of DIFT~\cite{DIFT}.
% SD+DINOv2
The combination of SD2-1 and DINOv2 yields the highest performance, which aligns with SD-DINO~\cite{A-Tale-of-Two-Features} and suggests that representations learned from these distinct paradigms are complementary.
}

To further analyze resolution impacts, we plot performance against feature map sizes in~\Cref{fig:Zero}. 
We find that ResNet-101, DINOv1, and iBOT exhibit minimal or negative gains when the resolution increases to 60 $\times$ 60 from 32 $\times$ 32. 
In contrast, SD2-1 and DINOv2 show consistent improvements with higher resolutions. 
We hypothesize this stems from two factors: 
(1) ResNet's local receptive fields limit global context capture, and 
(2) fixed positional encodings in DINOv1 and iBOT hinder adaptation when interpolated to larger spatial dimensions.

% Our zero-shot evaluation of backbones --- categorized as CNNs (ResNet-101), Transformers (e.g., DINO series), and generative models (SD2-1) --- shows that modern foundation models significantly outperform traditional CNNs. 
% Transformer
% This has led to a trend: a powerful approach is to fuse features from these different models. Combining the complementary strengths of models from different training paradigms (e.g., self-supervised and generative) may be a promising direction for solving complex vision tasks.

\subsubsection{Strongly Supervised Evaluation}
\begin{table}[!t]
    \begin{center}
    \caption{Evaluation of fine-tuning feature backbones on the SPair-71k dataset. Reso.: Image Resolution, Feature Map Size.}
    \vspace{-10pt}
    \label{tab:5_finetune_backbone}
    \renewcommand{\arraystretch}{1.2} % Row spacing
    \scalebox{1.0}{
    \begin{tabular}{c|c|cccc}
        \toprule
        \multirow{3}{*}{Backbone} & 
        \multirow{3}{*}{Reso.} & 
        \multicolumn{4}{c}{SPair-71k} \\
        & & \multicolumn{4}{c}{PCK @ $\alpha_{\text{bbox}}$} \\
        & & 0.01 & 0.05 & 0.1 & 0.15 \\
        \midrule
        SD-CPM & 960, 60 & 11.2 & \underline{62.0} & 75.4 & 80.6 \\
        SD-CPM & 512, 32 & 4.4 & 46.6 & 69.3 & 77.9 \\
        SD-CPM & 256, 16 & 0.4 & 7.5 & 22.1 & 35.4 \\
        \midrule
        SD-CPM+DINOv2 & (960,840), 60 & \underline{12.4} & \underline{62.0} & 76.0 & 81.2 \\
        SD-CPM+DINOv2 & (512,448), 32 & 5.5 & 49.4 & 70.4 & 78.7 \\
        SD-CPM+DINOv2 & (256,224), 16 & 1.6 & 25.9 & 51.7 & 65.2 \\
        \midrule
        DINOv2 & 840, 60 & \textbf{15.0} & \textbf{67.4} & \textbf{81.7} & \textbf{87.1} \\
        DINOv2 & 448, 32 & 7.1 & 56.2 & \underline{76.6} & \underline{84.3} \\
        DINOv2 & 224, 16 & 1.9 & 30.5 & 58.2 & 72.1 \\
        \bottomrule
    \end{tabular}
    }
    \end{center}
    \vspace{-10pt}
\end{table}
\begin{table}[!t]
    \begin{center}
    \caption{Evaluation of fine-tuning different stages of DINOv2 on the SPair-71k dataset. FF.: Fine-tune From, Reso.: Image Resolution, Feature Map Size.}
    \vspace{-10pt}
    \label{tab:5_finetune_DINOv2}
    \renewcommand{\arraystretch}{1.2} % Row spacing
    \scalebox{1.0}{
    \begin{tabular}{c|c|c|cccc}
        \toprule
        \multirow{3}{*}{Backbone} & 
        \multirow{3}{*}{FF.} & 
        \multirow{3}{*}{Reso.} & 
        \multicolumn{4}{c}{SPair-71k} \\
        & & & \multicolumn{4}{c}{PCK @ $\alpha_{\text{bbox}}$} \\
        & & & 0.01 & 0.05 & 0.1 & 0.15 \\
        \midrule
        \textcolor{gray}{DINOv2} & \textcolor{gray}{-} & \textcolor{gray}{840, 60} & \textcolor{gray}{7.3} & \textcolor{gray}{40.0} & \textcolor{gray}{54.4} & \textcolor{gray}{62.8} \\
        DINOv2 & stage12 & 840, 60 & 15.0 & 67.4 & 81.7 & 87.1 \\
        DINOv2 & stage11 & 840, 60 & \underline{17.5} & \textbf{73.3} & \textbf{85.1} & \textbf{89.3} \\
        DINOv2 & stage10 & 840, 60 & 16.6 & \underline{73.2} & \underline{84.5} & \underline{88.2} \\
        DINOv2 & stage9 & 840, 60 & \textbf{17.8} & 73.1 & 84.0 & 87.8 \\
        DINOv2 & stage8 & 840, 60 & 16.7 & 70.9 & 82.2 & 86.3 \\     
        \bottomrule
    \end{tabular}
    }
    \end{center}
    \vspace{-15pt}
\end{table}

In order to systematically analyze the adaptability of these backbones to semantic correspondence tasks, we also explore how fine-tuning improves backbone performance.
We select top-performing backbones in the zero-shot evaluation, SD2-1, DINOv2, and SD2-1+DINOv2, for this experiment. 

Specifically, for SD2-1, we adopt the SD4Match~\cite{SD4Match} framework by fine-tuning its conditional prompting module (SD-CPM). 
For DINOv2, we fine-tune the last stage of the model. 
For the combination of SD2-1 and DINOv2, we concatenate the feature maps from them along the channel dimension, and jointly optimize the conditional prompting module of SD2-1 and the final stage of DINOv2. 
We follow the learning framework in SimSC~\cite{SimSC}, using the cross-entropy loss between the query point and the entire target feature map as the training loss.

\begin{table*}[!t]
    \begin{center}
    \caption{Evaluation of different feature enhancement modules on the SPair-71k dataset. 
    The methods are categorized into two types: frozen DINOv2 backbone and joint training of the last layer of DINOv2 and the feature enhancement module.
    The highest PCK values are highlighted in bold, while baseline methods (without feature enhancement module) are indicated in gray. 
    FT.: Fine-tune. Reso.: Image Resolution, Feature Map Size.}
    \vspace{-10pt}
    \label{tab:6_bottleneck}
    \renewcommand{\arraystretch}{1.2} % Row spacing
    \scalebox{1.0}{
    \begin{tabular}{c|c|c|c|cccc|c}
        \toprule
        \multirow{3}{*}{Backbone} & 
        \multirow{3}{*}{FT.} & 
        \multirow{3}{*}{Bottleneck} & 
        \multirow{3}{*}{Reso.} & 
        \multicolumn{4}{c|}{SPair-71k} & \\ 
        & & & & \multicolumn{4}{c|}{PCK @ $\alpha_{\text{bbox}}$} & \\ 
        & & & & 0.01 & 0.05 & 0.1 & 0.15 & $\Delta$PCK@0.1 \\ 
        \midrule
        \textcolor{gray}{DINOv2} & \textcolor{gray}{$\times$} & \textcolor{gray}{-} & \textcolor{gray}{840, 60} & \textcolor{gray}{7.3} & \textcolor{gray}{40.0} & \textcolor{gray}{54.4} & \textcolor{gray}{62.8} & \textcolor{gray}{--} \\ 
        DINOv2 & $\times$ & ResNet Bottleneck Block & 840, 60 & 10.3 & 60.2 & 75.2 & 81.3 & +20.8 \\ 
        DINOv2 & $\times$ & Self-Attention & 840, 60 & 11.8 & 60.8 & 76.5 & 83.1 & +22.1 \\ 
        DINOv2 & $\times$ & Self+Cross Attention & 840, 60 & 11.5 & 60.6 & 76.3 & 83.2 & +21.9 \\ 
        \midrule
        \textcolor{gray}{DINOv2} & \textcolor{gray}{\checkmark} & \textcolor{gray}{-} & \textcolor{gray}{840, 60} & \textcolor{gray}{15.0} & \textcolor{gray}{67.4} & \textcolor{gray}{81.7} & \textcolor{gray}{87.1} & \textcolor{gray}{--} \\ 
        DINOv2 & \checkmark & ResNet Bottleneck Block & 840, 60 & \textbf{15.1} & \textbf{71.1} & \textbf{84.2} & \textbf{88.9} & +2.4 \\ 
        DINOv2 & \checkmark & Self-Attention & 840, 60 & 13.9 & \underline{66.4} & \underline{80.9} & \underline{87.1} & -0.8 \\ 
        DINOv2 & \checkmark & Self+Cross Attention & 840, 60 & \underline{14.5} & 66.3 & \underline{80.9} & 87.0 & -0.8 \\
        \bottomrule
    \end{tabular}
    }
    \end{center}
    \vspace{-15pt}
\end{table*}

% conclusion
The strongly supervised evaluation results are presented in \Cref{tab:5_finetune_backbone}.
% Fine-tuning the final stage of DINOv2 at an input resolution of $840 \times 840$ achieves a PCK of $81.74$ at $\alpha_{\text{bbox}} = 0.1$, demonstrating the strong representational power of its pre-trained features. 
Fine-tuning the final stage of DINOv2 at an input resolution of $840 \times 840$ yields superior performance across all thresholds, achieving a PCK of $81.74$ at $\alpha_{\text{bbox}} = 0.1$. 
Notably, while the combination of SD2-1 and DINOv2 showed complementary advantages in the zero-shot setting, DINOv2 alone outperforms this combination under fine-tuning. 
% reason 
We speculate that during the fine-tuning process, the complementary nature of these features might be diminished, with SD2-1 features possibly becoming the dominant influence, thus limiting the overall performance. 

% DINOv2
We further explore fine-tuning more stages of DINOv2. The results are summarized in~\Cref{tab:5_finetune_DINOv2}. Notably, we observe that the performance peaks when fine-tuning up to stage 11, achieving the highest PCK scores, but declines when deeper layers are fine-tuned. 
This phenomenon could be attributed to the increased number of trainable parameters in deeper layers. While these parameters help the model better fit the matching task, they also increase the risk of overfitting to training data, thereby compromising the model's generalization ability and resulting in decreased performance on unseen data.
\\

\noindent\textbf{\underline{Takeaways:}} 
\rev{\textbf{DINOv2 demonstrates superior performance for semantic correspondence.}}
According to~\Cref{tab:5_finetune_backbone}, DINOv2 exhibits superior performance across different resolutions, demonstrating its versatility in adapting to varying input sizes while maintaining high performance. 
Its self-supervised pre-training strategy enables the learning of highly generalizable feature representations, making it particularly beneficial for semantic correspondence. 
\textbf{Fine-tuning feature backbones plays a crucial role in performance enhancement.} 
As demonstrated in \Cref{tab:5_finetune_DINOv2}, task-specific adaptation of DINOv2 significantly improves performance: fine-tuning its last two transformer blocks at $840 \times 840$ resolution achieves a PCK@0.1 score of 85.1\%, marking a 30\% improvement over the zero-shot performance. 
\textbf{For DINOv2, larger image sizes improve semantic correspondence performance.}
Results from \Cref{tab:4_zero_backbone} and \Cref{tab:5_finetune_backbone} suggest that higher image resolutions may enhance performance by preserving finer spatial details, which could enable more precise semantic correspondence.
\subsection{Feature Enhancement Evaluation} 
\label{sec:feature_enhancement}

After fine-tuning the feature backbone, we explore how feature enhancement techniques and matching refinement modules may further improve matching accuracy. We select DINOv2 as the feature backbone and investigate two commonly used feature enhancement architectures: CNN architecture~\cite{GeoAware-SC, SFNet, SFNet2022, SCorrSAN} and Transformer architecture~\cite{LoFTR}.
% resnet bottleneck
For the CNN approach, we adopt the standard ResNet Bottleneck Block architecture with 768 input/output channels and a bottleneck dimension of 192 channels (1/4 reduction ratio).
% transformer bottleneck
Similarly, for the transformer approach, we adopt a transformer block with 768 feature dimensions and a linear attention mechanism for efficient computation. 
The block consists of standard multi-head self-attention layers followed by a two-layer MLP with 1536 hidden dimensions.
The block operates in two modes: 
self-attention (intra-image feature interaction) and 
self+cross-attention (intra-image self-focus followed by cross-image interaction).
% self-attention where features attend to their own spatial locations, and self+cross-attention where features from one image attend to their own positions then interact with cross-image features.

We evaluate these enhancements under two training strategies: frozen backbone and end-to-end joint training. In the latter case, only the final stage of DINOv2 is fine-tuned.
% frozen vs joint training
The results are summarized in Table~\ref{tab:6_bottleneck}, with additional experiments across different input resolutions presented in Appendix Table A1.

% frozen
\noindent\textbf{Frozen Backbone:} 
With the DINOv2 backbone frozen, training the feature enhancement layers yields substantial performance gains, with PCK@0.1 improving by more than 20\%. 
The performance variations across different enhancement techniques are marginal,
suggesting that feature enhancement layers consistently boost representation quality regardless of their specific architecture when working with a frozen backbone.

% joint training
\noindent\textbf{Joint Training:} 
When jointly fine-tuning the backbone and enhancement module, the performance gains brought by the enhancement module are limited.
The ResNet Bottleneck Block achieves a modest gain of 2.42\% in PCK@0.1, while the Attention Block shows no advantage and slightly degrades performance. 
This phenomenon may be attributed to the backbone's final layer already serving as an effective feature enhancer and having largely saturated the potential for further improvement. Therefore, it leaves minimal room for additional enhancement modules to provide further gains.
\\

\noindent\textbf{\underline{Takeaway:}}
\textbf{The efficacy of feature enhancement depends on the training paradigm:} 
Table~\ref{tab:6_bottleneck} shows that with frozen backbones, feature enhancement modules boost PCK@0.1 by over 20\%. 
During joint training, gains diminish sharply (ResNet: +2.4\%; attention: performance degradation), revealing architectural limitations in such settings.

\begin{table*}[!t]
    \begin{center}
    \caption{Evaluation of cost aggregation module on the SPair-71k dataset. 
    The methods are categorized into two types: frozen DINOv2 backbone and joint training of the last layer of DINOv2 and the cost aggregator.
    The baseline methods (without cost aggregator) are indicated in gray.  FT.: Fine-tune. Reso.: Image Resolution, Feature Map Size.}
    \vspace{-10pt}
    \label{tab:9_Cost_Aggragation}
    \renewcommand{\arraystretch}{1.2} % Row spacing
    \scalebox{1.0}{
    \begin{tabular}{c|c|c|c|cccc|c}
        \toprule
        \multirow{3}{*}{Backbone} & 
        \multirow{3}{*}{FT.} & 
        \multirow{3}{*}{Cost Aggregator} & 
        \multirow{3}{*}{Reso.} & 
        \multicolumn{4}{c|}{SPair-71k} & \\ 
        & & & & \multicolumn{4}{c|}{PCK @ $\alpha_{\text{bbox}}$} & \\ 
        & & & & 0.01 & 0.05 & 0.1 & 0.15 & $\Delta$PCK@0.1 \\ 
        \midrule
        \textcolor{gray}{DINOv2} & \textcolor{gray}{$\times$} & \textcolor{gray}{-} & \textcolor{gray}{840, 60} & \textcolor{gray}{7.3} & \textcolor{gray}{40.0} & \textcolor{gray}{54.4} & \textcolor{gray}{62.8} & \textcolor{gray}{--} \\ 
        DINOv2 & $\times$ & Match2Match & 840, 60 & 6.8 & 39.5 & 54.1 & 62.6 & -0.3 \\ 
        DINOv2 & $\times$ & NeighConsensus & 840, 60 & 6.2 & 45.8 & 61.9 & 69.3 & +7.5 \\ 
        \midrule
        % 448
        \textcolor{gray}{DINOv2} & \textcolor{gray}{$\times$} & \textcolor{gray}{-} & \textcolor{gray}{$448,32$} & \textcolor{gray}{3.7} & \textcolor{gray}{35.4} & \textcolor{gray}{52.9} & \textcolor{gray}{62.2} & \textcolor{gray}{--} \\ 
        DINOv2 & $\times$ & CATs & 448, 32 & 2.6 & 30.8 & 49.5 & 59.7 & -3.4 \\ 
        DINOv2 & $\times$ & Match2Match & 448, 32 & 3.6 & 34.8 & 52.2 & 61.7 & -0.7 \\ 
        DINOv2 & $\times$ & NeighConsensus & 448, 32 & 2.8 & 33.4 & 56.5 & 67.7 & +3.6 \\ 
        \midrule
        \textcolor{gray}{DINOv2} & \textcolor{gray}{\checkmark} & \textcolor{gray}{-} & \textcolor{gray}{840, 60} & \textcolor{gray}{15.0} & \textcolor{gray}{67.4} & \textcolor{gray}{81.7} & \textcolor{gray}{87.1} & \textcolor{gray}{--} \\ 
        DINOv2 & \checkmark & Match2Match & 840, 60 & \textbf{13.9} & \underline{68.3} & \underline{83.3} & \underline{88.6} & +1.6 \\ 
        DINOv2 & \checkmark & NeighConsensus & 840, 60 & \underline{12.7} & \textbf{70.2} & \textbf{85.2} & \textbf{89.6} & +3.5 \\ 
        \midrule
        \textcolor{gray}{DINOv2} & \textcolor{gray}{\checkmark} & \textcolor{gray}{-} &\textcolor{gray}{448, 32} & \textcolor{gray}{7.1} & \textcolor{gray}{56.2} & \textcolor{gray}{76.6} & \textcolor{gray}{84.3} & \textcolor{gray}{--} \\ 
        DINOv2 & \checkmark & CATs & 448, 32 & 3.2 & 42.6 & 71.4 & 82.4 & -5.2 \\ 
        DINOv2 & \checkmark & Match2Match & 448, 32 & 6.3 & 55.4 & 77.8 & 85.9 & +1.2 \\ 
        DINOv2 & \checkmark & NeighConsensus & 448, 32 & 5.0 & 54.7 & 79.4 & 87.3 & +2.8 \\
        \bottomrule
    \end{tabular}
    }
    \end{center}
    \vspace{-10pt}
\end{table*}

\subsection{Matching Refinement Evaluation}
\label{sec:cost_aggregation}

After evaluating the feature enhancement modules, we proceed to analyze matching refinement methods. 
Prior research has predominantly focused on refining cost volume, as described in \Cref{sec:match_filter}. 
These methods can be largely categorized into CNN \cite{NC-Net, ANC-Net, DualRC-Net, CHM, CHMNet} and Transformer \cite{CATs, CATs++, VAT, NeMF, IFCAT, TransforMatcher} approaches. 
Therefore, we select the Neighbourhood Consensus Aggregator \cite{NC-Net} for CNN methods, and both the CATs Aggregator \cite{CATs} and Match-to-Match Aggregator \cite{TransforMatcher} for Transformer approaches, as they represent the majority of the cost volume-based methods. 

\noindent\textbf{Neighbourhood Consensus Aggregator.} 
While following the neighbourhood consensus module from NC-Net \cite{NC-Net}, our implementation differs in two aspects: we employ $3 \times 3 \times 3 \times 3$ filters instead of the original $5 \times 5 \times 5 \times 5$ filters, and halve the number of feature channels in the intermediate convolutional layers from 16 to 8. 
These modifications reduce the computational cost while maintaining the essential functionality of neighbourhood consensus filtering for match refinement.

\noindent\textbf{CATs Aggregator.} \ 
We adopt the cost aggregator architecture from CATs \cite{CATs} but implement a simplified version, reducing it from six layers to a single layer. 
Given the significant computational demands of Transformer operations at high resolutions, we restrict our experiments to 448 and 224 resolutions for this module.
%~\ref{tab:appendix_cost}.
\\
\noindent\textbf{Match-to-Match Aggregator.} \ 
Another Transformer approach is the match-to-match attention module \cite{TransforMatcher}, which treats each spatial match as an attention unit. 
We follow the original experimental configuration in TransforMatcher \cite{TransforMatcher}.

The evaluation results are summarized in Table~\ref{tab:9_Cost_Aggragation}, with additional experiments across different input resolutions presented in Appendix Table A2.
Neighbourhood consensus module consistently outperforms other cost aggregators on the SPair-71k dataset, demonstrating its superior capability in refining the cost volume and enhancing semantic matching performance. 
\\

\noindent\textbf{\underline{Takeaway:}}
\textbf{CNN cost aggregators are more effective than Transformer aggregators.} 
CNN aggregators, such as Neighbourhood Consensus, consistently outperform Transformer counterparts as shown in Table~\ref{tab:9_Cost_Aggragation}, particularly at lower resolutions (e.g., $224\times224$), as shown in Appendix Table A2.
% ~\ref{tab:appendix_cost}.
This suggests that the local receptive field and hierarchical feature processing of CNN architectures are more suitable for cost volume refinement compared to the global attention mechanism of transformers.
\subsection{Experimental Results Discussion}

% \begin{figure}[htb]
%     \centering
%     \includegraphics[width=0.45\textwidth]{fig_exp/Discussion.pdf}
%     \vspace{-2mm}
%     \caption{Performance of ResNet bottleneck and neighbourhood consensus aggregator for frozen and fine-tuned DINOv2 with a feature map size of 60$\times$60.} 
%     \label{fig:Discussion}
%     \vspace{-2mm}
% \end{figure}

\begin{table*}[!t]
    \begin{center}
    \caption{Performance comparison of zero-shot methods on SPair-71k, PF-PASCAL, and PF-WILLOW datasets.
    The highest PCK values are emphasized in bold, whereas the second highest are underlined.}
    \vspace{-5pt}
    \label{tab:baseline_zero}
    \renewcommand{\arraystretch}{1.2} % 行距
    \scalebox{1.0}{
    \begin{tabular}{l|l|>{\centering\arraybackslash}p{1.2cm}|ccc|ccc|ccc}
        \toprule
        \multirow{3}{*}{Methods} & \multirow{3}{*}{Backbone} & \multirow{3}{*}{Reso.} & 
        \multicolumn{3}{c|}{SPair-71k} & \multicolumn{3}{c|}{PF-PASCAL} & 
        \multicolumn{3}{c}{PF-WILLOW} \\
        & & & \multicolumn{3}{c|}{PCK @ $\alpha_{\text{bbox}}$}
            & \multicolumn{3}{c|}{PCK @ $\alpha_{\text{img}}$}
            & \multicolumn{3}{c}{PCK @ $\alpha_{\text{bbox-kp}}$} \\
        & & & 0.05 & 0.1 & 0.15 & 0.05 & 0.1 & 0.15 & 0.05 & 0.1 & 0.15 \\
        \midrule % 水平线
        LDM \cite{LDM_correspondences} & SD & - & 28.9 & 45.4 & - & - & - & - & - & - & - \\
        DIFTsd \cite{DIFT} & SD & 768 & - & 52.9 & - & - & - & - & - & - & - \\
        SD-DINO \cite{A-Tale-of-Two-Features} & SD+DINOv2 & 960, 840\tnote{1}
        & - & 59.3 & - & \underline{73.0} & 86.1 & \underline{91.1} & - & - & - \\
        GeoAware-SC \cite{GeoAware-SC} & SD+DINOv2 & 960, 840 & \underline{45.3} & \textbf{61.3} & - & \textbf{74.0} & \underline{86.2} & 90.7 & - & - & - \\
        \midrule
        \textbf{Ours} & SD+DINOv2 & 960, 840 & \textbf{47.9} & \underline{60.4} & \textbf{67.0} & 72.9 & \textbf{88.3} & \textbf{93.4} & 46.6 & 72.8 & 85.8  \\
        \bottomrule
    \end{tabular}
    }
    \begin{tablenotes}
        \item[1] 960, 840 represent image resolutions: 960$\times$960 for SD and 840$\times$840 for DINOv2, respectively.
    \end{tablenotes}
    \end{center}
    \vspace{-10pt}
\end{table*}
\begin{table*}[!t]
    \begin{center}
    \caption{Performance comparison of supervised methods on SPair-71k, PF-PASCAL, and PF-WILLOW datasets. 
    The highest PCK values are emphasized in bold, whereas the second highest are underlined. 
    Reso.: Resolution. 
    DINOv2 + ResNet refers to the DINOv2 backbone with additional ResNet bottleneck.
    DINOv2 + NC refers to the DINOv2 backbone with Neighbourhood Consensus cost aggregator.
    }
    \vspace{-5pt}
    \label{tab:baseline_supervised}
    \renewcommand{\arraystretch}{1.2} % 行距
    \scalebox{1.0}{
    \begin{tabular}{
         l
        |l
        |>{\centering\arraybackslash}p{1.2cm}
        |ccc|ccc|ccc}
        \toprule
        \multirow{3}{*}{Methods} & \multirow{3}{*}{Backbone} & 
        \multirow{3}{*}{Reso.} & 
        \multicolumn{3}{c|}{SPair-71k} & \multicolumn{3}{c|}{PF-PASCAL} & 
        \multicolumn{3}{c}{PF-WILLOW} \\
        & & & \multicolumn{3}{c|}{PCK @ $\alpha_{\text{bbox}}$}
                  & \multicolumn{3}{c|}{PCK @ $\alpha_{\text{img}}$}
                  & \multicolumn{3}{c}{PCK @ $\alpha_{\text{bbox-kp}}$} \\
        & & & 0.05 & 0.1 & 0.15 & 0.05 & 0.1 & 0.15 & 0.05 & 0.1 & 0.15 \\
        \midrule
        SimSC-iBOT~\cite{SimSC} & iBOT & 256 & 43.0 & 63.5 & - &  \textbf{88.4} & \underline{95.6} & 97.3 &  44.9 & 71.4 & 84.5 \\
        DHF~\cite{Diffusion-Hyperfeatures} & SD+DINOv2 & Ori & - & 64.6 & - & - & 86.7 & - & - & - & - \\
        LPMFlow~\cite{LPMFlow} & ViT-B/16 & 256 & 46.7 & 65.6 & - & 82.4 & 94.3 & 97.2 & - & \textbf{81.0} & - \\
        SD-DINO~\cite{A-Tale-of-Two-Features} & SD+DINOv2 & 960, 840  & - & 74.6 & - & 80.9 & 93.6 & 96.9 & - & - & - \\
        SD4Match~\cite{SD4Match} & SD+DINOv2 & 768 & 59.5 & 75.5 & - & 84.4 & 95.2 & \underline{97.5} & \textbf{52.1} & \underline{80.4} & \textbf{91.2} \\
        GeoAware-SC~\cite{GeoAware-SC} & SD+DINOv2 & 960, 840 & 72.6 & 82.9 & - & 85.5 &  95.1 & 97.4 & - & - & - \\
        \midrule
        \textbf{Ours(DINOv2)} & DINOv2 & 840 & \textbf{73.3} & \underline{85.1} & 89.3 & \underline{87.6} & \textbf{95.8} & \textbf{98.2} & \underline{48.8} & 73.7 & 85.8 \\
        \textbf{Ours(DINOv2+ResNet)} & DINOv2 & 840 & \underline{72.7} & \textbf{85.2} & \underline{89.3} & 80.3 & 90.8 & 94.9 & 46.6 & 74.1 & \underline{87.1} \\
        \textbf{Ours(DINOv2+NC)} & DINOv2 & 840 & 70.2 & \textbf{85.2} & \textbf{89.6} & 80.6 & 91.5 & 95.0 & 39.8 & 62.3 & 74.9\\
        \bottomrule
    \end{tabular}
    }
    \end{center}
    \vspace{-15pt}
\end{table*}
\begin{table}[!t]
    \begin{center}
    \caption{Performance comparison of supervised methods on the AP-10K Dataset. Reso.: Image Resolution}
    \vspace{-5pt}
    \label{tab:ap10k}
    \renewcommand{\arraystretch}{1.2} % Row spacing
    \scalebox{1.0}{
    \begin{tabular}{c|c
        |cccc}
        \toprule
        \multirow{3}{*}{Methods} & 
        \multirow{3}{*}{Reso.} & 
        \multicolumn{4}{c}{AP-10K} \\
        & & \multicolumn{4}{c}{PCK @ $\alpha_{\text{bbox}}$} \\
        & &  0.01 & 0.05 & 0.1 & 0.15 \\
        \midrule
        GeoAware-SC~\cite{GeoAware-SC} & 960,840 & 23.1 & 73.0 & \textbf{87.5} & - \\
        \midrule
        \textbf{Ours(DINOv2)}  & 840 & \textbf{24.5} & \textbf{74.3} & 87.4 & 92.2 \\
        \textbf{Ours(DINOv2+ResNet)} & 840 & \underline{24.0} & \underline{73.1} & 86.7 & \underline{92.3} \\
        \textbf{Ours(DINOv2+NC)} & 840 & 21.1 & 72.8 & \underline{87.4} & \textbf{92.6} \\
        \bottomrule
    \end{tabular}
    }
    \end{center}
    \vspace{-15pt}
\end{table}

\noindent\textbf{Feature adaptation is more impactful than cost aggregation in the absence of backbone fine-tuning.}
% We compare effectiveness of ResNet feature enhancement module and neighbourhood consensus cost aggregator on DINOv2 backbone in \Cref{fig:Discussion}. 
% Results show that with a frozen backbone, ResNet feature enhancement brings larger performance gains than neighbourhood consensus cost aggregation.
\rev{As observed in \Cref{tab:6_bottleneck} and \Cref{tab:9_Cost_Aggragation}, with a frozen backbone, ResNet feature enhancement brings a greater performance gain than Neighbourhood Consensus cost aggregation.}

\noindent\textbf{The backbone dominates the fundamental performance characteristics.} 
The experimental results presented in \Cref{sec:feature_enhancement} and \Cref{sec:cost_aggregation} demonstrate that joint training of DINOv2 with either bottlenecks or cost aggregators yields relatively modest performance improvements. 
These empirical findings indicate that the backbone's feature extraction capability serves as a critical constraint on overall matching performance. 
This can be attributed to the backbone's critical role in constructing the initial feature representations, which serve as a solid foundation for subsequent matching quality. 
\subsection{A Strong Baseline}\label{sec:strong_baseline}

Based on our extensive experiments and analysis, we have constructed an effective semantic matching framework in both zero-shot and supervised settings. 
% zero-shot
For zero-shot evaluation, we compare the best performing zero-shot model, SD2-1+DINOv2, against the literature in Table~\ref{tab:baseline_zero}. As demonstrated, despite the simpler architecture, our SD2-1+DINOv2 achieves comparable results to GeoAware-SC~\cite{GeoAware-SC}, showing the redundancy in the more complex designs in the literature.

% supervised
In the supervised setting, we compare the best-performing backbone (DINOv2), feature enhancement design (DINOv2+ResNet), and matching filtering design (DINOv2+NC) against other methods in~\Cref{tab:baseline_supervised}.
Notably, by fine-tuning the last two layers of DINOv2 alone, we achieve the second-best performance on SPair-71k. 
When incorporating the ResNet bottleneck and the Neighbourhood Consensus cost aggregator, our approach attains state-of-the-art results on both SPair-71k and PF-PASCAL datasets.

% % AP-10K 
Beyond these \rev{commonly used} benchmarks, we further validate our approach on the AP-10K dataset (Table~\ref{tab:ap10k}). Despite a simple architecture and straightforward training strategy, our method achieves comparable state-of-the-art results as well.
\section{Conclusion and Discussion}
\label{sec:Conclusion}
% taxonomy
In this paper, we first present a comprehensive review of semantic correspondence methods and introduce a structured taxonomy that systematically categorizes existing approaches according to their architectural designs and training strategies. 
% experiments
Through extensive controlled experiments across all stages of the pipeline, we then identify critical insights at each stage of the semantic matching model. 
% baseline
Finally, based on these findings, we propose a simple yet effective baseline that achieves state-of-the-art performance on multiple benchmark datasets, including SPair-71k, PF-PASCAL, and AP-10K, establishing a solid foundation for future research. 

% future research
While our work demonstrates the effectiveness of DINOv2's feature extraction capabilities through fine-tuning, the full potential of foundation models remains largely untapped. Future research could explore more effective adaptation techniques to better leverage foundation model features. 
In terms of training strategies, one possible direction is to further investigate pseudo-label generation techniques, so as to alleviate the reliance on large-scale manual annotations. 
These directions may advance semantic matching from both architectural and training perspectives.
% \newpage

\section*{Acknowledgements}
This work is supported by the National Natural Science Foundation of China under Grant 62306251, the Hong Kong Research Grants Council - Early Career Scheme under Grant 27208022, the Hong Kong Research Grants Council - General Research Fund under Grant 17211024, and the HKU Seed Fund for Basic Research.
We are grateful to Jean Ponce for his valuable comments on an early version of this manuscript.

\ifCLASSOPTIONcaptionsoff
  \newpage
\fi

\begin{footnotesize}
  {
  \bibliographystyle{IEEEtran}
  \normalem
  % argument is your BibTeX string definitions and bibliography database(s)
  \bibliography{ref}

@string{BMVC = {Proc. British Machine Vision Conference (BMVC)}}

@string{CVIU = {Computer Vision and Image Understanding (CVIU)}}

@string{CVPR = {Proc. IEEE Conference on Computer Vision and Pattern Recognition (CVPR)}}

@string{CVPRW = {Proc. IEEE Conference on Computer Vision and Pattern Recognition Workshops (CVPRW)}}

@string{ECCV = {Proc. European Conference on Computer Vision (ECCV)}}

@string{ECCVW = {Proc. European Conference on Computer Vision Workshops (ECCVW)}}

@string{ICCV = {Proc. IEEE International Conference on Computer Vision (ICCV)}}

@string{ICLR = {International Conference on Learning Representations (ICLR)}}

@string{ICML = {Proc. International Conference on Machine Learning (ICML)}}

@string{WACV = {Proc. Winter Conference on Applications of Computer Vision (WACV)}}

@string{NIPS = {Advances in Neural Information Processing Systems (NeurIPS)}}

@string{IJCV = {International Journal of Computer Vision (IJCV)}}

@string{TPAMI = {IEEE Transactions on Pattern Analysis and Machine Intelligence (TPAMI)}}

@string{BMVC = {BMVC}}

@string{CVIU = {CVIU}}

@string{CVPR = {CVPR}}

@string{CVPRW = {CVPRW}}

@string{ECCV = {ECCV}}

@string{ECCVW = {ECCVW}}

@string{ICCV = {ICCV}}

@string{ICLR = {ICLR}}

@string{ICML = {ICML}}

@string{WACV = {WACV}}

@string{NIPS = {NeurIPS}}

@string{IJCV = {IJCV}}

@string{TPAMI = {TPAMI}}

@inproceedings{lee2020reference,
  title={Reference-based sketch image colorization using augmented-self reference and dense semantic correspondence},
  author={Lee, Junsoo and Kim, Eungyeup and Lee, Yunsung and Kim, Dongjun and Chang, Jaehyuk and Choo, Jaegul},
  booktitle=CVPR,
  pages={5801--5810},
  year={2020}
}

@article{geyer2023tokenflow,
  title={Tokenflow: Consistent diffusion features for consistent video editing},
  author={Geyer, Michal and Bar-Tal, Omer and Bagon, Shai and Dekel, Tali},
  booktitle= ICLR,
  pages={20637--20650},
  year={2024}
}

@inproceedings{gupta2023asic,
  title={ASIC: Aligning sparse in-the-wild image collections},
  author={Gupta, Kamal and Jampani, Varun and Esteves, Carlos and Shrivastava, Abhinav and Makadia, Ameesh and Snavely, Noah and Kar, Abhishek},
  booktitle=ICCV,
  pages={4134--4145},
  year={2023}
}

@inproceedings{ofri2023neural,
  title={Neural congealing: Aligning images to a joint semantic atlas},
  author={Ofri-Amar, Dolev and Geyer, Michal and Kasten, Yoni and Dekel, Tali},
  booktitle=CVPR,
  pages={19403--19412},
  year={2023}
}

@inproceedings{ASYM,
  title={Demystifying unsupervised semantic correspondence estimation},
  author={Ayg{\"u}n, Mehmet and Mac Aodha, Oisin},
  booktitle=ECCV,
  pages={125--142},
  year={2022}
}

@inproceedings{bleyer2011patchmatch,
  title={Patchmatch stereo-stereo matching with slanted support windows},
  author={Bleyer, Michael and Rhemann, Christoph and Rother, Carsten},
  booktitle=BMVC,
  volume={11},
  pages={1--11},
  year={2011}
}

@article{mou2023dragondiffusion,
  title={Dragon{D}iffusion: Enabling drag-style manipulation on diffusion models},
  author={Mou, Chong and Wang, Xintao and Song, Jiechong and Shan, Ying and Zhang, Jian},
  booktitle = ICLR,
  pages={8098--8110},
  year={2024}
}

@inproceedings{RegionDrag,
        author = {Jingyi Lu and Xinghui Li and Kai Han},
        title = {RegionDrag: Fast Region-Based Image Editing with Diffusion Models},
        booktitle = ECCV,
        pages={231--246},
        year      = {2024}
}

@inproceedings{cho2015unsupervised,
  title={Unsupervised object discovery and localization in the wild: Part-based matching with bottom-up region proposals},
  author={Cho, Minsu and Kwak, Suha and Schmid, Cordelia and Ponce, Jean},
  booktitle=CVPR,
  pages={1201--1210},
  year={2015}
}

@book{hartley2003multiple,
  title={Multiple view geometry in computer vision},
  author={Hartley, Richard and Zisserman, Andrew},
  year={2003},
  publisher={Cambridge university press}
}

@article{sift,
  title={Distinctive image features from scale-invariant keypoints},
  author={Lowe, David G},
  journal= IJCV,
  volume={60},
  number={2},
  pages={91--110},
  year={2004},
  publisher={Springer}
}

@INPROCEEDINGS{Ponce2005,
  author={Lazebnik, S. and Schmid, C. and Ponce, J.},
  booktitle={ICCV'05}, 
  title={A maximum entropy framework for part-based texture and object recognition}, 
  pages={832--838},
  year={2005}}

@INPROCEEDINGS{Ponce2007,
  author={Kushal, Akash and Schmid, Cordelia and Ponce, Jean},
  booktitle=CVPR, 
  title={Flexible Object Models for Category-Level 3D Object Recognition}, 
  pages={1-8},
  year={2007}}

@ARTICLE{DAISY,
  author={Tola, Engin and Lepetit, Vincent and Fua, Pascal},
  journal=TPAMI, 
  title={DAISY: An Efficient Dense Descriptor Applied to Wide-Baseline Stereo}, 
  year={2010},
  volume={32},
  number={5},
  pages={815-830}
  }

@ARTICLE{SIFTFlow,
  author={Liu, Ce and Yuen, Jenny and Torralba, Antonio},
  journal=TPAMI, 
  title={SIFT Flow: Dense Correspondence across Scenes and Its Applications}, 
  year={2011},
  volume={33},
  number={5},
  pages={978-994}
}

@inproceedings{HOG,
  title={Histograms of oriented gradients for human detection},
  author={Dalal, Navneet and Triggs, Bill},
  booktitle=CVPR,
  pages={886--893},
  year={2005}
}

@inproceedings{DAISY-Filter-Flow,
  title={DAISY Filter Flow: A Generalized Discrete Approach to Dense Correspondences},
  author={Hongsheng Yang and Wen-Yan Lin and Jiangbo Lu},
  booktitle=CVPR,
  pages={3406--3413},
  year={2014},
}

@InProceedings{long2014convnets,
    title     = {Do convnets learn correspondence?},
    author    = {Long, Jonathan L and Zhang, Ning and Darrell, Trevor},
    booktitle = NIPS,
    pages     = {1601--1609},
    year      = {2014}
}

@InProceedings{UCN,
    title     = {Universal correspondence network},
    author    = {Choy, Christopher and Gwak, JunYoung and Savarese, Silvio and Chandraker, Manmohan},
    booktitle = NIPS,
    volume={29},
    year      = {2016}
}

@INPROCEEDINGS{graph-matching,
  author={Duchenne, Olivier and Joulin, Armand and Ponce, Jean},
  booktitle= ICCV, 
  title={A graph-matching kernel for object categorization}, 
  pages={1792--1799},
  year={2011}
  }

@inproceedings{Cho2012ProgressiveGM,
  title={Progressive graph matching: Making a move of graphs via probabilistic voting},
  author={Minsu Cho and Kyoung Mu Lee},
  booktitle = CVPR,
  pages={398--405},
  year={2012},
}

@InProceedings{SCNet,
    title     = {SCNet: Learning semantic correspondence},
    author    = {Han, Kai and Rezende, Rafael S and Ham, Bumsub and Wong, Kwan-Yee K and Cho, Minsu and Schmid, Cordelia and Ponce, Jean},
    booktitle = ICCV,
    pages={1849-1858},
    year      = {2017}
}

@InProceedings{FCSS,
    title     = {FCSS: Fully Convolutional Self-Similarity for Dense Semantic Correspondence},
    author    = {Kim, Seungryong and Min, Dongbo and Ham, Bumsub and Jeon, Sangryul and Lin, Stephen and Sohn, Kwanghoon},
    booktitle=CVPR,
    pages     = {6560--6569},
    year      = {2017}
}

@InProceedings{DCTM,
    title     = {DCTM: Discrete-Continuous Transformation Matching for Semantic Flow},
    author    = {Kim, Seungryong and Min, Dongbo and Lin, Stephen and Sohn, Kwanghoon},
    booktitle = ICCV,
    pages     = {4529--4538},
    year      = {2017}
}

@InProceedings{HyperpixelFlow, 
   title     = {Hyperpixel Flow: Semantic Correspondence with Multi-layer Neural Features},
   author    = {Juhong Min and Jongmin Lee and Jean Ponce and Minsu Cho},
   booktitle = ICCV,
  pages     = {3395--3404},
   year      = {2019}
}

@InProceedings{DynamicHyperpixelFlow, 
    title={Learning to Compose Hypercolumns for Visual Correspondence},
    author={Juhong Min and Jongmin Lee and Jean Ponce and Minsu Cho},
    booktitle=ECCV,
    pages={346--363},
    year={2020}
}

@inproceedings{LoFTR,
  title={LoFTR: Detector-free local feature matching with transformers},
  author={Sun, Jiaming and Shen, Zehong and Wang, Yuang and Bao, Hujun and Zhou, Xiaowei},
  booktitle=CVPR,
  pages={8922--8931},
  year={2021}
}

@inproceedings{HCCNet,
  title={Efficient Semantic Matching with Hypercolumn Correlation},
  author={Kim, Seungwook and Min, Juhong and Cho, Minsu},
  booktitle= WACV,
  pages={139--148},
  year={2024}
}

@article{KBCNet,
  title={KBCNet: Independently keypoint learning for Small Object Semantic Correspondence},
  author={Jin, Hailong and Li, Huiying},
  journal={Expert Systems with Applications},
  volume={261},
  pages={125476},
  year={2025},
  publisher={Elsevier}
}

@inproceedings{LPMFlow,
  title={Pixel-level Semantic Correspondence through Layout-aware Representation Learning and Multi-scale Matching Integration},
  author={Sun, Yixuan and Yin, Zhangyue and Wang, Haibo and Wang, Yan and Qiu, Xipeng and Ge, Weifeng and Zhang, Wenqiang},
  booktitle=CVPR,
  pages={17047--17056},
  year={2024}
}

@InProceedings{hariharan2015hypercolumns,
    title     = {Hypercolumns for object segmentation and fine-grained localization},
    author    = {Hariharan, Bharath and Arbel{\'a}ez, Pablo and Girshick, Ross and Malik, Jitendra},
    booktitle=CVPR,
    pages     = {447--456},
    year      = {2015}
}

@article{beamsearch,
  title={Speech understanding systems: Report of a steering committee},
  author={Medress, Mark F. and Cooper, Franklin S and Forgie, Jim W. and Green, CC and Klatt, Dennis H. and O'Malley, Michael H. and Neuburg, Edward P and Newell, Allen and Reddy, DR and Ritea, Barry and others},
  journal={Artificial Intelligence},
  volume={9},
  number={3},
  pages={307--316},
  year={1977}
}

@inproceedings{SwinTransformer,
  title={Swin transformer: Hierarchical vision transformer using shifted windows},
  author={Liu, Ze and Lin, Yutong and Cao, Yue and Hu, Han and Wei, Yixuan and Zhang, Zheng and Lin, Stephen and Guo, Baining},
  booktitle={ICCV},
  pages={10012--10022},
  year={2021}
}

@inproceedings{VAT,
  title={Cost aggregation with 4{D} convolutional swin transformer for few-shot segmentation},
  author={Hong, Sunghwan and Cho, Seokju and Nam, Jisu and Lin, Stephen and Kim, Seungryong},
  booktitle=ECCV,
  pages={108--126},
  year={2022}
}

@inproceedings{LiFT,
  title={LiFT: A Surprisingly Simple Lightweight Feature Transform for Dense ViT Descriptors},
  author={Suri, Saksham and Walmer, Matthew and Gupta, Kamal and Shrivastava, Abhinav},
  booktitle={ECCV},
  pages={110--128},
  year={2024}
}

@inproceedings{CATs,
  title={Cats: Cost aggregation transformers for visual correspondence},
  author={Cho, Seokju and Hong, Sunghwan and Jeon, Sangryul and Lee, Yunsung and Sohn, Kwanghoon and Kim, Seungryong},
  booktitle=NIPS,
  pages={9011--9023},
  year={2021}
}

@article{CATs++,
  title={Cats++: Boosting cost aggregation with convolutions and transformers},
  author={Cho, Seokju and Hong, Sunghwan and Kim, Seungryong},
  journal=TPAMI,
  volume={45},
  number={6},
  pages={7174--7194},
  year={2022}
}

@inproceedings{PMNC,
  title={Patchmatch-based neighborhood consensus for semantic correspondence},
  author={Lee, Jae Yong and DeGol, Joseph and Fragoso, Victor and Sinha, Sudipta N.},
  booktitle=CVPR,
  pages={13153--13163},
  year={2021}
}

@inproceedings{SemiMatch,
  title={Semi-supervised learning of semantic correspondence with pseudo-labels},
  author={Kim, Jiwon and Ryoo, Kwangrok and Seo, Junyoung and Lee, Gyuseong and Kim, Daehwan and Cho, Hansang and Kim, Seungryong},
  booktitle=CVPR,
  pages={19699--19709},
  year={2022}
}

@inproceedings{TransforMatcher,
  title={Transformatcher: Match-to-match attention for semantic correspondence},
  author={Kim, Seungwook and Min, Juhong and Cho, Minsu},
  booktitle=CVPR,
  pages={8697--8707},
  year={2022}
}

@article{SimSC,
  title={SimSC: A simple framework for semantic correspondence with temperature learning},
  author={Li, Xinghui and Han, Kai and Wan, Xingchen and Prisacariu, Victor Adrian},
  journal={arXiv preprint arXiv:2305.02385},
  year={2023}
}

@inproceedings{ResNet,
  title={Deep residual learning for image recognition},
  author={He, Kaiming and Zhang, Xiangyu and Ren, Shaoqing and Sun, Jian},
  booktitle=CVPR,
  pages={770--778},
  year={2016}
}

@inproceedings{SCorrSAN,
  title={Learning semantic correspondence with sparse annotations},
  author={Huang, Shuaiyi and Yang, Luyu and He, Bo and Zhang, Songyang and He, Xuming and Shrivastava, Abhinav},
  booktitle=ECCV,
  pages={267--284},
  year={2022},
  organization={Springer},
  year={2022}
}

@inproceedings{han2018co,
  title={Co-teaching: Robust training of deep neural networks with extremely noisy labels},
  author={Han, Bo and Yao, Quanming and Yu, Xingrui and Niu, Gang and Xu, Miao and Hu, Weihua and Tsang, Ivor and Sugiyama, Masashi},
  booktitle=NIPS,
  volume={31},
  year={2018}
}

@InProceedings{RTNs,
    title     = {Recurrent transformer networks for semantic correspondence},
    author    = {Kim, Seungryong and Lin, Stephen and Jeon, Sangryul and Min, Dongbo and Sohn, Kwanghoon},
    booktitle = NIPS,
    volume={31},
    year      = {2018}
}

@inproceedings{SFNet,
  title={SFNet: Learning object-aware semantic correspondence},
  author={Lee, Junghyup and Kim, Dohyung and Ponce, Jean and Ham, Bumsub},
  booktitle=CVPR,
  pages={2278--2287},
  year={2019}
}

@ARTICLE{SFNet2022,
  author={Lee, Junghyup and Kim, Dohyung and Lee, Wonkyung and Ponce, Jean and Ham, Bumsub},
  journal= TPAMI, 
  title={Learning Semantic Correspondence Exploiting an Object-Level Prior}, 
  year={2022},
  volume={44},
  number={3},
  pages={1399-1414},
  }

@InProceedings{PARN,
    title     = {PARN: Pyramidal Affine Regression Networks for Dense Semantic Correspondence},
    author    = {Jeon, Sangryul and Kim, Seungryong and Min, Dongbo and Sohn, Kwanghoon},
    booktitle = ECCV,
    pages     = {351-366},
    year      = {2018}
}

@InProceedings{paul2018attentive,
    title     = {Attentive Semantic Alignment with Offset-Aware Correlation Kernels},
    author    = {Paul Hongsuck Seo and Jongmin Lee and Deunsol Jung and Bohyung Han and Minsu Cho},
    booktitle = ECCV,
    pages     = {349--364},
    year      = {2018}
}

@inproceedings{SAM-Net,
  title={Semantic attribute matching networks},
  author={Kim, Seungryong and Min, Dongbo and Jeong, Somi and Kim, Sunok and Jeon, Sangryul and Sohn, Kwanghoon},
  booktitle= CVPR,
  pages={12339--12348},
  year={2019}
}

@inproceedings{SC-ImageNet,
  title={Weakly Supervised Learning of Semantic Correspondence through Cascaded Online Correspondence Refinement},
  author={Huang, Yiwen and Sun, Yixuan and Lai, Chenghang and Xu, Qing and Wang, Xiaomei and Shen, Xuli and Ge, Weifeng},
  booktitle=ICCV,
  pages={16254--16263},
  year={2023}
}

@inproceedings{DistillDIFT,
  author    = {Frank Fundel and Johannes Schusterbauer and Vincent Tao Hu and Björn Ommer},
  title    = {Distillation of Diffusion Features for Semantic Correspondence},
  booktitle  = {WACV},
  pages     = {6762--6774},
  year      = {2025}
}

@inproceedings{thai20243,
  title={3$\times$ 2: 3D Object Part Segmentation by 2D Semantic Correspondences},
  author={Thai, Anh and Wang, Weiyao and Tang, Hao and Stojanov, Stefan and Rehg, James M and Feiszli, Matt},
  booktitle={ECCV},
  pages={149-166},
  year={2024}
}

@article{UFC2024,
  title={Unifying Feature and Cost Aggregation with Transformers for Semantic and Visual Correspondence},
  author={Hong, Sunghwan and Cho, Seokju and Kim, Seungryong and Lin, Stephen},
  booktitle = ICLR,
  pages     = {4732--4746},
  year={2024}
}

@InProceedings{SCOT,
    title     = {Semantic Correspondence as an Optimal Transport Problem},
    author    = {Liu, Yanbin and Zhu, Linchao and Yamada, Makoto and Yang, Yi},
    booktitle=CVPR,
    pages     = { 4463--4472},
    year      = {2020}
}

@inproceedings{PMD,
  title={Probabilistic model distillation for semantic correspondence},
  author={Li, Xin and Fan, Deng-Ping and Yang, Fan and Luo, Ao and Cheng, Hong and Liu, Zicheng},
  booktitle= CVPR,
  pages={7505--7514},
  year={2021}
}

@InProceedings{GSF,
    title={Guided Semantic Flow},
    author={Sangryul Jeon and Dongbo Min and Seungryong Kim and Jihwan Choe and Kwanghoon Sohn},
    booktitle=ECCV,
    pages={631--648},
    year={2020}
}

@inproceedings{DCCNet,
  title={Dynamic context correspondence network for semantic alignment},
  author={Huang, Shuaiyi and Wang, Qiuyue and Zhang, Songyang and Yan, Shipeng and He, Xuming},
  booktitle= ICCV,
  pages={2010--2019},
  year={2019}
}

@InProceedings{DualRC2020,
  title={Dual-resolution correspondence networks},
  author={Li, Xinghui and Han, Kai and Li, Shuda and Prisacariu, Victor},
  booktitle=NIPS,
  pages={17346--17357},
  year={2020}
}

@article{DualRC-Net,
  title={DualRC: A Dual-Resolution Learning Framework With Neighbourhood Consensus for Visual Correspondences},
  author={Li, Xinghui and Han, Kai and Li, Shuda and Prisacariu, Victor},
  journal=TPAMI,
  year={2024},
  volume={46},
  number={1},
  pages={236-249},
  doi={10.1109/TPAMI.2023.3316770}
  }

@article{MatchMe,
  title={Match me if you can: Semantic Correspondence Learning with Unpaired Images},
  author={Kim, Jiwon and Heo, Byeongho and Yun, Sangdoo and Kim, Seungryong and Han, Dongyoon},
  journal={arXiv preprint arXiv:2311.18540},
  year={2023}
}

@article{IFCAT,
  title={Integrative feature and cost aggregation with transformers for dense correspondence},
  author={Hong, Sunghwan and Cho, Seokju and Kim, Seungryong and Lin, Stephen},
  journal={arXiv preprint arXiv:2209.08742},
  year={2022}
}

@inproceedings{MMNet,
  title={Multi-scale matching networks for semantic correspondence},
  author={Zhao, Dongyang and Song, Ziyang and Ji, Zhenghao and Zhao, Gangming and Ge, Weifeng and Yu, Yizhou},
  booktitle= ICCV,
  pages={3354--3364},
  year={2021}
}

@inproceedings{FPN,
  title={Feature pyramid networks for object detection},
  author={Lin, Tsung-Yi and Doll{\'a}r, Piotr and Girshick, Ross and He, Kaiming and Hariharan, Bharath and Belongie, Serge},
  booktitle=CVPR,
  pages={2117--2125},
  year={2017}
}

@inproceedings{ACTR,
  title={Correspondence transformers with asymmetric feature learning and matching flow super-resolution},
  author={Sun, Yixuan and Zhao, Dongyang and Yin, Zhangyue and Huang, Yiwen and Gui, Tao and Zhang, Wenqiang and Ge, Weifeng},
  booktitle=CVPR,
  pages={17787--17796},
  year={2023}
}

@inproceedings{laskar2019semantic,
  title={Semantic matching by weakly supervised 2d point set registration},
  author={Laskar, Zakaria and Tavakoli, Hamed Rezazadegan and Kannala, Juho},
  booktitle= WACV,
  pages={1061--1069},
  year={2019}
}

@article{ma2021image,
  title={Image matching from handcrafted to deep features: A survey},
  author={Ma, Jiayi and Jiang, Xingyu and Fan, Aoxiang and Jiang, Junjun and Yan, Junchi},
  journal={IJCV},
  volume={129},
  number={1},
  pages={23--79},
  year={2021},
  publisher={Springer}
}

@article{horn1981determining,
  title={Determining optical flow},
  author={Horn, Berthold KP and Schunck, Brian G},
  journal={Artificial intelligence},
  volume={17},
  number={1-3},
  pages={185--203},
  year={1981},
  publisher={Elsevier},
}

@article{marr1979computational,
  title={A computational theory of human stereo vision},
  author={Marr, David and Poggio, Tomaso},
  journal={Proceedings of the Royal Society of London. Series B. Biological Sciences},
  year={1979},
  volume={204},
  number={1156},
  pages={301--328},
  publisher={The Royal Society London}
}

@InProceedings{3D-guided,
    title     = {Learning dense correspondence via 3d-guided cycle consistency},
    author    = {Zhou, Tinghui and Krahenbuhl, Philipp and Aubry, Mathieu and Huang, Qixing and Efros, Alexei A},
    booktitle = {CVPR},
    pages     = {117--126},
    year      = {2016}
}

@inproceedings{WeakMatchNet,
  author = {Y.-C. Chen and P.-H. Huang and L.-Y. Yu and J.-B. Huang and M.-H. Yang and Y.-Y. Lin},
  booktitle = {Asian Conference on Computer Vision},
  title = {Deep Semantic Matching with Foreground Detection and Cycle-Consistency},
  pages = {347--362},
  year = {2018}
}

@inproceedings{CycleGAN,
  title={Unpaired image-to-image translation using cycle-consistent adversarial networks},
  author={Zhu, Jun-Yan and Park, Taesung and Isola, Phillip and Efros, Alexei A},
  booktitle={ECCV},
  pages={2223--2232},
  year={2017}
}

@InProceedings{CNNGeometric,
    author    = {Rocco, Ignacio and Arandjelovic, Relja and Sivic, Josef},
    title     = {Convolutional neural network architecture for geometric matching},
    booktitle=CVPR,
    pages     = {6148--6157},
    year      = {2017}
}

@InProceedings{WeakAlign,
    author    = {Rocco, Ignacio and Arandjelović, Relja and Sivic, Josef},
    title     = {End-to-end weakly-supervised semantic alignment},
    booktitle=CVPR,
    pages     = {6916--6925},
    year      = {2018}
}

@inproceedings{FMAP,
  title={Zero-shot image feature consensus with deep functional maps},
  author={Cheng, Xinle and Deng, Congyue and Harley, Adam W and Zhu, Yixin and Guibas, Leonidas},
  booktitle={ECCV},
  pages={277--293},
  year={2024}
}

@InProceedings{NC-Net,
    title     = {Neighbourhood consensus networks},
    author    = {Rocco, Ignacio and Cimpoi, Mircea and Arandjelovi{\'c}, Relja and Torii, Akihiko and Pajdla, Tomas and Sivic, Josef},
    booktitle = NIPS,
    volume={31},
    year      = {2018}
}

@inproceedings{ANC-Net,
  title={Correspondence networks with adaptive neighbourhood consensus},
  author={Li, Shuda and Han, Kai and Costain, Theo W and Howard-Jenkins, Henry and Prisacariu, Victor},
  booktitle= CVPR,
  pages={10196--10205},
  year={2020}
}

@InProceedings{TSS,
    title     = {Joint recovery of dense correspondence and cosegmentation in two images},
    author    = {Taniai, Tatsunori and Sinha, Sudipta N and Sato, Yoichi},
    booktitle=CVPR,
    pages     = {4246--4255},
    year      = {2016}
}

@InProceedings{FlowWeb,
    title     = {Flow{W}eb: Joint image set alignment by weaving consistent, pixel-wise correspondences},
    author    = {Zhou, Tinghui and Jae Lee, Yong and Yu, Stella X and Efros, Alyosha A},
    booktitle=CVPR,
    pages     = {1191--1200},
    year      = {2015}
}

@InProceedings{GLU-Net,
    title     = {{GLU-Net}: Global-Local Universal Network for dense flow and correspondences},
    author    = {Prune Truong and Martin Danelljan and Radu Timofte},
    booktitle=CVPR,
    pages     = {6258--6268},
    year      = {2020}
}

@inproceedings{CHM,
  title={Convolutional Hough Matching Networks},
  author={Min, Juhong and Cho, Minsu},
  booktitle=CVPR,
  pages={2940--2950},
  year={2021}
}

@article{CHMNet,
  title={Convolutional hough matching networks for robust and efficient visual correspondence},
  author={Min, Juhong and Kim, Seungwook and Cho, Minsu},
  journal= TPAMI,
  volume={45},
  number={7},
  pages={8159--8175},
  year={2023}
}

@article{SPair-71k,
   title   = {{SPair}-71k: A Large-scale Benchmark for Semantic Correspondence},
   author  = {Juhong Min and Jongmin Lee and Jean Ponce and Minsu Cho},
   journal = {arXiv prepreint arXiv:1908.10543},
   year    = {2019}
}

@inproceedings{WarpC,
  title={Warp consistency for unsupervised learning of dense correspondences},
  author={Truong, Prune and Danelljan, Martin and Yu, Fisher and Van Gool, Luc},
  booktitle=ICCV,
  pages={10346--10356},
  year={2021}
}

@inproceedings{PWarpC,
  title={Probabilistic warp consistency for weakly-supervised semantic correspondences},
  author={Truong, Prune and Danelljan, Martin and Yu, Fisher and Van Gool, Luc},
  booktitle=CVPR,
  pages={8708--8718},
  year={2022}
}

@inproceedings{shtedritski2023learning,
  title={Learning universal semantic correspondences with no supervision and automatic data curation},
  author={Shtedritski, Aleksandar and Vedaldi, Andrea and Rupprecht, Christian},
  booktitle=ICCV,
  pages={933--943},
  year={2023}
}

@InProceedings{DETR,
    title     = {End-to-end object detection with transformers},
    author    = {Carion, Nicolas and Massa, Francisco and Synnaeve, Gabriel and Usunier, Nicolas and Kirillov, Alexander and Zagoruyko, Sergey},
    booktitle = ECCV,
    pages     = {213--229},
    year      = {2020}
}

@article{DeformableDETR,
  title={Deformable DETR: Deformable Transformers for End-to-End Object Detection},
  author={Zhu, Xizhou and Su, Weijie and Lu, Lewei and Li, Bin and Wang, Xiaogang and Dai, Jifeng},
  booktitle = ICLR,
  pages     = {894--910},
  year={2021}
}

@InProceedings{ViT,
    title     = {An image is worth 16x16 words: Transformers for image recognition at scale},
    author    = {Dosovitskiy, Alexey and Beyer, Lucas and Kolesnikov, Alexander and Weissenborn, Dirk and Zhai, Xiaohua and Unterthiner, Thomas and Dehghani, Mostafa and Minderer, Matthias and Heigold, Georg and Gelly, Sylvain and Uszkoreit, Jakob and Houlsby, Neil},
    booktitle = ICLR,
    year      = {2021}
}

@InProceedings{DINO-ViT,
    title     = {Emerging properties in self-supervised vision transformers},
    author    = {Caron, Mathilde and Touvron, Hugo and Misra, Ishan and J\'egou, Herv\'e and Mairal, Julien and Bojanowski, Piotr and Joulin, Armand},
    booktitle = ICCV,
    pages     = {9650--9660},
    year      = {2021}
}

@article{Deep-vit-features,
	  author    = {Shir Amir and Yossi Gandelsman and Shai Bagon and Tali Dekel},
  	title     = {Deep ViT Features as Dense Visual Descriptors}, 
	  journal   = {ECCVW What is Motion For?},
	  year      = {2022},
  }

@inproceedings{CLIP,
  title={Learning transferable visual models from natural language supervision},
  author={Radford, Alec and Kim, Jong Wook and Hallacy, Chris and Ramesh, Aditya and Goh, Gabriel and Agarwal, Sandhini and Sastry, Girish and Askell, Amanda and Mishkin, Pamela and Clark, Jack and others},
  booktitle=ICML,
  pages={8748--8763},
  year={2021}
}

@article{iBOT,
  title={iBOT: Image BERT Pre-Training with Online Tokenizer},
  author={Zhou, Jinghao and Wei, Chen and Wang, Huiyu and Shen, Wei and Xie, Cihang and Yuille, Alan and Kong, Tao},
  journal=ICLR,
  year={2022}
}

@inproceedings{Imagenet,
  title={Imagenet: A large-scale hierarchical image database},
  author={Deng, Jia and Dong, Wei and Socher, Richard and Li, Li-Jia and Li, Kai and Fei-Fei, Li},
  booktitle=CVPR,
  pages={248--255},
  year={2009}
}

@inproceedings{A-Tale-of-Two-Features,
  title={A tale of two features: Stable diffusion complements dino for zero-shot semantic correspondence},
  author={Zhang, Junyi and Herrmann, Charles and Hur, Junhwa and Polania Cabrera, Luisa and Jampani, Varun and Sun, Deqing and Yang, Ming-Hsuan},
  booktitle=NIPS,
  pages={45533--45547},
  year={2023}
}

@article{DINOv2,
  title={DINOv2: Learning Robust Visual Features without Supervision},
  author={Oquab, Maxime and Darcet, Timoth{\'e}e and Moutakanni, Th{\'e}o and Vo, Huy V and Szafraniec, Marc and Khalidov, Vasil and Fernandez, Pierre and HAZIZA, Daniel and Massa, Francisco and El-Nouby, Alaaeldin and others},
  journal={Transactions on Machine Learning Research}
}

@inproceedings{Stable-Diffusion,
  title={High-resolution image synthesis with latent diffusion models},
  author={Rombach, Robin and Blattmann, Andreas and Lorenz, Dominik and Esser, Patrick and Ommer, Bj{\"o}rn},
  booktitle=CVPR,
  pages={10684--10695},
  year={2022}
}

@inproceedings{DIFT,
  title={Emergent correspondence from image diffusion},
  author={Tang, Luming and Jia, Menglin and Wang, Qianqian and Phoo, Cheng Perng and Hariharan, Bharath},
  booktitle=NIPS,
  pages={1363--1389},
  year={2023}
}

@inproceedings{SD4Match,
  title={SD4Match: Learning to Prompt Stable Diffusion Model for Semantic Matching},
  author={Li, Xinghui and Lu, Jingyi and Han, Kai and Prisacariu, Victor},
  booktitle= CVPR,
  pages={27558--27568},
  year={2024}
}

@inproceedings{LDM_correspondences,
  title={Unsupervised semantic correspondence using stable diffusion},
  author={Hedlin, Eric and Sharma, Gopal and Mahajan, Shweta and Isack, Hossam and Kar, Abhishek and Tagliasacchi, Andrea and Yi, Kwang Moo},
  booktitle=NIPS,
  pages={8266--8279},
  year={2024}
}

@inproceedings{Diffusion-Hyperfeatures,
  title={Diffusion hyperfeatures: Searching through time and space for semantic correspondence},
  author={Luo, Grace and Dunlap, Lisa and Park, Dong Huk and Holynski, Aleksander and Darrell, Trevor},
  booktitle=NIPS,
  pages={47500--47510},
  year={2023}
}

@inproceedings{GeoAware-SC,
  title={Telling left from right: Identifying geometry-aware semantic correspondence},
  author={Zhang, Junyi and Herrmann, Charles and Hur, Junhwa and Chen, Eric and Jampani, Varun and Sun, Deqing and Yang, Ming-Hsuan},
  booktitle= CVPR,
  pages={3076--3085},
  year={2024}
}

@inproceedings{Transformer,
  title={Attention is all you need},
  author={Vaswani, Ashish and Shazeer, Noam and Parmar, Niki and Uszkoreit, Jakob and Jones, Llion and Gomez, Aidan N and Kaiser, {\L}ukasz and Polosukhin, Illia},
  booktitle= NIPS,
  volume={30},
  year={2017}
}

@inproceedings{NeMF,
  title={Neural matching fields: Implicit representation of matching fields for visual correspondence},
  author={Hong, Sunghwan and Nam, Jisu and Cho, Seokju and Hong, Susung and Jeon, Sangryul and Min, Dongbo and Kim, Seungryong},
  booktitle=NIPS,
  pages={13512--13526},
  year={2022}
}

@inproceedings{Sparse-NCNet,
  title={Efficient neighbourhood consensus networks via submanifold sparse convolutions},
  author={Rocco, Ignacio and Arandjelovi{\'c}, Relja and Sivic, Josef},
  booktitle=ECCV,
  pages={605--621},
  year={2020}
}

@InProceedings{chen_cvpr14,
    author       = {Xianjie Chen and Roozbeh Mottaghi and Xiaobai Liu and Sanja Fidler and Raquel Urtasun and Alan Yuille},
    title        = {Detect What You Can: Detecting and Representing Objects using Holistic Models and Body Parts},
    booktitle    = CVPR,
    pages        = {1971-1978},
    year         = {2014},
}

@article{ham2018proposal,
    title   = {Proposal flow: Semantic correspondences from object proposals},
    author  = {Ham, Bumsub and Cho, Minsu and Schmid, Cordelia and Ponce, Jean},
    journal = TPAMI,
    volume  = {40},
    number  = {7},
    pages   = {1711--1725},
    year    = {2018},
    publisher = {IEEE}
}

@InProceedings{ufer2017deep,
    title     = {Deep semantic feature matching},
    author    = {Ufer, Nikolai and Ommer, Bj{\"o}rn},
    booktitle=CVPR,
    pages     = {5929--5938},
    year      = {2017}
}

@InProceedings{taniai2016joint,
    title     = {Joint recovery of dense correspondence and cosegmentation in two images},
    author    = {Taniai, Tatsunori and Sinha, Sudipta N and Sato, Yoichi},
    booktitle=CVPR,
    pages     = {4246--4255},
    year      = {2016}
}

@InProceedings{DSP2013,
    title     = {Deformable spatial pyramid matching for fast dense correspondences},
    author    = {Kim, Jaechul and Liu, Ce and Sha, Fei and Grauman, Kristen},
    booktitle=CVPR,
    pages     = {2307--2314},
    year      = {2013}
}

@inproceedings{fei2004learning,
  title={Learning generative visual models from few training examples: An incremental bayesian approach tested on 101 object categories},
  author={Fei-Fei, Li and Fergus, Rob and Perona, Pietro},
  booktitle={CVPRW},
  pages={178--178},
  year={2004}
}

@article{li2006one,
    title   = {One-shot learning of object categories},
    author  = {Li, Fei-Fei and Fergus, Rob and Perona, Pietro},
    journal = TPAMI,
    volume  = {28},
    number  = {4},
    pages   = {594--611},
    year    = {2006},
    publisher = {IEEE}
}

@InProceedings{ham2016proposal,
    title     = {Proposal flow},
    author    = {Ham, Bumsub and Cho, Minsu and Schmid, Cordelia and Ponce, Jean},
    booktitle=CVPR,
    pages     = {3475--3484},
    year      = {2016}
}

@Article{everingham2015pascal,
    title   = {The Pascal Visual Object Classes Challenge: A Retrospective},
    author  = {Everingham, Mark and Eslami, S. M. Ali and Van Gool, Luc and Williams, Christopher K. I. and Winn, John and Zisserman, Andrew},
    journal = IJCV,
    year    = {2015},
    month   = {Jan},
    day     = {01},
    volume  = {111},
    number  = {1},
    pages   = {98--136}
}

@inproceedings{lin2014jointly,
  title={Jointly optimizing 3{D} model fitting and fine-grained classification},
  author={Lin, Yen-Liang and Morariu, Vlad I and Hsu, Winston and Davis, Larry S.},
  booktitle=ECCV,
  pages={466--480},
  year={2014}
}

@inproceedings{rubinstein2013unsupervised,
  title={Unsupervised joint object discovery and segmentation in internet images},
  author={Rubinstein, Michael and Joulin, Armand and Kopf, Johannes and Liu, Ce},
  booktitle=CVPR,
  pages={1939--1946},
  year={2013}
}

@inproceedings{hariharan2011semantic,
  title={Semantic contours from inverse detectors},
  author={Hariharan, Bharath and Arbel{\'a}ez, Pablo and Bourdev, Lubomir and Maji, Subhransu and Malik, Jitendra},
  booktitle=ICCV,
  pages={991--998},
  year={2011}
}

@inproceedings{Xiang2014BeyondPA,
    title   = {Beyond PASCAL: A benchmark for 3D object detection in the wild},
    author  = {Yu Xiang and Roozbeh Mottaghi and Silvio Savarese},
    booktitle = WACV,
    pages   = {75-82},
    year    = {2014}
}

@InProceedings{cho2013learning,
    title     = {Learning graphs to match},
    author    = {Cho, Minsu and Alahari, Karteek and Ponce, Jean},
    booktitle = ICCV,
    pages     = {25-32},
    year      = {2013}
}

@techreport{griffin2007caltech,
  title={Caltech-256 object category dataset},
  author={Griffin, Gregory and Holub, Alex and Perona, Pietro and others},
  year={2007},
  institution={Technical Report 7694, California Institute of Technology Pasadena}
}

@inproceedings{AP-10K,
  title={Ap-10k: A benchmark for animal pose estimation in the wild},
  author={Yu, Hang and Xu, Yufei and Zhang, Jing and Zhao, Wei and Guan, Ziyu and Tao, Dacheng},
  booktitle=NIPS,
  year={2021}
}

@article{xian2018zero,
  title={Zero-shot learning—a comprehensive evaluation of the good, the bad and the ugly},
  author={Xian, Yongqin and Lampert, Christoph H and Schiele, Bernt and Akata, Zeynep},
  journal=TPAMI,
  volume={41},
  number={9},
  pages={2251--2265},
  year={2018},
  publisher={IEEE},
  journal=TPAMI
}

@misc{africanWildLife,
  author = {Bianca Ferreira},
  title = {African WildLife},
  year = {2021},
  publisher = {Kaggle},
  url = {https://www.kaggle.com/biancaferreira/african-wildlife}
}

@misc{wcats,
  author = {Enis Šahović},
  title = {wild cats},
  year = {2020},
  publisher = {Kaggle},
  url  ={https://www.kaggle.com/enisahovi/cats-projekat-4}
}

@misc{animal5,
  author = {Yagnik Vinodkumar Trivedi},
  title = {Animals 5},
  year = {2020},
  publisher = {Kaggle},
  url ={https://www.kaggle.com/ytrivedi1/animals-5}
  }

@misc{animalDCP,
  author = {Ashish Saxena},
  title = {Animal Image Dataset(DOG, CAT and PANDA)},
  year = {2019},
  publisher = {Kaggle},
  url = {https://www.kaggle.com/ashishsaxena2209/animal-image-datasetdog-cat-and-panda}
}

@misc{animals10,
  author = {Corrado Alessio},
  title = {Animals 10},
  year = {2019},
  publisher = {Kaggle},
  url ={https://www.kaggle.com/alessiocorrado99/animals10}}

@misc{IUCN,
  author = {antoreepjana},
  title = {IUCN Animals Dataset},
  year = {2021},
  publisher = {Kaggle},
  url = {https://www.kaggle.com/antoreepjana/iucn-animals-dataset}
}

@misc{endangeredanimals,
  author = {Sonain Jamil},
  title = {Endangered Animals},
  year = {2020},
  publisher = {Kaggle},
  url ={https://www.kaggle.com/sonain/endangered-animals}
  }

@article{textual_inversion,
  title={An image is worth one word: Personalizing text-to-image generation using textual inversion},
  author={Gal, Rinon and Alaluf, Yuval and Atzmon, Yuval and Patashnik, Or and Bermano, Amit H and Chechik, Gal and Cohen-Or, Daniel},
  booktitle = ICLR,
  pages = {611--632},
  year={2021}
}

@inproceedings{bian2017gms,
  title={GMS: Grid-based motion statistics for fast, ultra-robust feature correspondence},
  author={Bian, JiaWang and Lin, Wen-Yan and Matsushita, Yasuyuki and Yeung, Sai-Kit and Nguyen, Tan-Dat and Cheng, Ming-Ming},
  booktitle=CVPR,
  pages={4529--4538},
  year={2017}
}

@inproceedings{sattler2009improving,
  title={Improving RANSAC's Efficiency with a Spatial Consistency Filter},
  author={Sattler, T and Leibe, B and Kobbelt, L},
  booktitle=ICCV,
  pages={2090--2097},
  year={2009}
}

@inproceedings{schaffalitzky2002automated,
  title={Automated scene matching in movies},
  author={Schaffalitzky, Frederik and Zisserman, Andrew},
  booktitle={Image and Video Retrieval: International Conference},
  pages={186--197},
  year={2002}
}

@article{DINOv3,
  title={DINOv3},
  author={Sim{\'e}oni, Oriane and Vo, Huy V and Seitzer, Maximilian and Baldassarre, Federico and Oquab, Maxime and Jose, Cijo and Khalidov, Vasil and Szafraniec, Marc and Yi, Seungeun and Ramamonjisoa, Micha{\"e}l and others},
  journal={arXiv preprint arXiv:2508.10104},
  year={2025}
}

@inproceedings{CleanDIFT,
    title={CleanDIFT: Diffusion Features without Noise}, 
    author={Nick Stracke and Stefan Andreas Baumann and Kolja Bauer and Frank Fundel and Björn Ommer},
    booktitle=CVPR,
    pages={117-127},
    year={2025} 
}

@inproceedings{SemAlign3D,
  title     = {SemAlign3D: Semantic Correspondence between RGB-Images through Aligning 3D Object-Class Representations},
  author    = {Wandel, Krispin and Wang, Hesheng},
  booktitle = CVPR,
  pages={1138-1147},
  year      = {2025},
}

@article{SCAC,
  title={Towards Robust Semantic Correspondence: A Benchmark and Insights},
  author={Chong, Wenyue},
  journal={arXiv preprint arXiv:2508.00272},
  year={2025}
}

@article{DiTF,
  title={Unleashing Diffusion Transformers for Visual Correspondence by Modulating Massive Activations},
  author={Gan, Chaofan and Tu, Yuanpeng and Chen, Xi and Chen, Tieyuan and Li, Yuxi and Harandi, Mehrtash and Lin, Weiyao},
  journal={arXiv preprint arXiv:2505.18584},
  year={2025}
}

@article{mariotti2025jamais,
  title={Jamais Vu: Exposing the Generalization Gap in Supervised Semantic Correspondence},
  author={Mariotti, Octave and Du, Zhipeng and Bhalgat, Yash and Mac Aodha, Oisin and Bilen, Hakan},
  journal={arXiv preprint arXiv:2506.08220},
  year={2025}
}

@article{DIY-SC,
    title = {Do It Yourself: Learning Semantic Correspondence from Pseudo-Labels},
    author = {D{\"u}nkel, Olaf and Wimmer, Thomas and Theobalt, Christian and Rupprecht, Christian and Kortylewski, Adam},
    journal = {arXiv preprint arXiv:2506.05312},
    year = {2025}
  }

@inproceedings{qian2025bridging,
  title={Bridging Viewpoint Gaps: Geometric Reasoning Boosts Semantic Correspondence},
  author={Qian, Qiyang and Chen, Hansheng and Tomizuka, Masayoshi and Keutzer, Kurt and Wang, Qianqian and Xu, Chenfeng},
  booktitle= CVPR,
  pages={11579--11589},
  year={2025}
}

@article{SURF,
  title={Speeded-up robust features (SURF)},
  author={Bay, Herbert and Ess, Andreas and Tuytelaars, Tinne and Van Gool, Luc},
  journal= CVIU,
  volume={110},
  number={3},
  pages={346--359},
  year={2008}
}

@inproceedings{ORB,
  title={ORB: An efficient alternative to SIFT or SURF},
  author={Rublee, Ethan and Rabaud, Vincent and Konolige, Kurt and Bradski, Gary},
  booktitle= ICCV,
  pages={2564--2571},
  year={2011}
}
  }
\end{footnotesize}

% \bibliographystyle{IEEEtran}
% \normalem
% \bibliography{ref}

% \input{Biography}

\clearpage
\begin{appendices}
\setcounter{table}{0} 
\setcounter{figure}{0} 
\setcounter{page}{1}

\textbf{Overview.}
In this appendix, we present detailed experimental evaluations and qualitative analysis to support our main findings.
In Appendix A, we expand the experiments of feature enhancement in \Cref{sec:feature_enhancement} and matching refinement in \Cref{sec:cost_aggregation} with more resolutions, evaluating feature enhancement modules by comparing CNN and Transformer approaches (Appendix A-A) and assessing several cost aggregation methods, including the Neighbourhood Consensus, CATs Aggregator, and Match-to-Match attention module (Appendix A-B).
\rev{
In Appendix B, we qualitatively compare the robustness of our method against the state-of-the-art approach GeoAware-SC~\cite{GeoAware-SC} through a qualitative analysis. 
\rev{In Appendix C, we analyze the newly released DINOv3 \cite{DINOv3}, with its zero-shot and fine-tuned performance.}
In Appendix D, we present a comprehensive efficiency analysis of the benchmarks, evaluating methods in terms of GFLOPs, peak GPU memory usage, and inference time across different resolutions.
In Appendix E, we provide several challenging cases, discuss the limitations of the proposed method, and suggest potential directions for future work. }

\section{Feature Enhancement and Matching Refinement Evaluation}
\renewcommand{\thetable}{\Alph{section}\arabic{table}}
\renewcommand{\thefigure}{\Alph{section}\arabic{figure}} 
\begin{table*}[b]
    \begin{center}
    \caption{Evaluation of different bottlenecks on the SPair-71k dataset. 
    The methods are categorized into two types: frozen DINOv2 backbone and joint training of the last layer of DINOv2 and bottleneck.
    The highest PCK values are highlighted in bold, while baseline methods (without bottleneck) are indicated in gray. 
    FT.: Fine-tune, Reso.: Image Resolution, Feature Map Size.}
    \vspace{-10pt}
    \label{tab:appendix_bottleneck}
    \renewcommand{\arraystretch}{1.2} % Row spacing
    \scalebox{1.0}{
    \begin{tabular}{c|c|c|c|cccc|c}
        \toprule
        \multirow{3}{*}{Backbone} & 
        \multirow{3}{*}{FT.} & 
        \multirow{3}{*}{Bottleneck} & 
        \multirow{3}{*}{Reso.} & 
        \multicolumn{4}{c}{SPair-71k} &  \\ 
        & & & & \multicolumn{4}{c}{PCK @ $\alpha_{\text{bbox}}$} & \\ 
        & & & & 0.01 & 0.05 & 0.1 & 0.15 & $\Delta$PCK$\alpha$ 0.1 \\ 
        \midrule
        \textcolor{gray}{DINOv2} & \textcolor{gray}{$\times$} & \textcolor{gray}{-} & \textcolor{gray}{$840, 60$} & \textcolor{gray}{7.3} & \textcolor{gray}{40.0} & \textcolor{gray}{54.4} & \textcolor{gray}{62.8} & \textcolor{gray}{--} \\ 
        DINOv2 & $\times$ & ResNet Bottleneck Block & $840, 60$ & 10.3 & 60.2 & 75.2 & 81.3 & +20.8 \\ 
        DINOv2 & $\times$ & Self-Attention & $840, 60$ & 11.8 & 60.8 & 76.5 & 83.1 & +22.1 \\ 
        DINOv2 & $\times$ & Self+Cross Attention & $840, 60$ & 11.5 & 60.6 & 76.3 & 83.2 & +21.9 \\ 
        \midrule
        \textcolor{gray}{DINOv2} & \textcolor{gray}{$\times$} & \textcolor{gray}{-} & \textcolor{gray}{$448, 32$} & \textcolor{gray}{3.7} & \textcolor{gray}{35.4} & \textcolor{gray}{52.9} & \textcolor{gray}{62.2} & \textcolor{gray}{--} \\ 
        DINOv2 & $\times$ & ResNet Bottleneck Block & $448, 32$ & 4.3 & 46.9 & 69.5 & 78.7 & +16.6 \\ 
        DINOv2 & $\times$ & Self-Attention & $448, 32$ & 5.3 & 49.2 & 70.2 & 78.7 & +17.3 \\ 
        DINOv2 & $\times$ & Self+Cross Attention & $448, 32$ & 5.7 & 49.6 & 69.8 & 78.3 & +16.9 \\ 
        \midrule
        \textcolor{gray}{DINOv2} & \textcolor{gray}{$\times$} & \textcolor{gray}{-} & \textcolor{gray}{$224, 16$} & \textcolor{gray}{1.1} & \textcolor{gray}{20.9} & \textcolor{gray}{43.1} & \textcolor{gray}{56.2} & \textcolor{gray}{--} \\ 
        DINOv2 & $\times$ & ResNet Bottleneck Block & $224, 16$ & 1.3 & 23.8 & 50.9 & 66.3 & +7.8 \\ 
        DINOv2 & $\times$ & Self-Attention & $224, 16$ & 1.1 & 20.5 & 44.8 & 59.6 & +1.7 \\ 
        DINOv2 & $\times$ & Self+Cross Attention & $224, 16$ & 1.6 & 25.3 & 50.2 & 63.8 & +7.1 \\ 
        \midrule
        \textcolor{gray}{DINOv2} & \textcolor{gray}{\checkmark} & \textcolor{gray}{-} & \textcolor{gray}{$840, 60$} & \textcolor{gray}{15.0} & \textcolor{gray}{67.4} & \textcolor{gray}{81.7} & \textcolor{gray}{87.1} & \textcolor{gray}{--} \\ 
        DINOv2 & \checkmark & ResNet Bottleneck Block & $840, 60$ & \textbf{15.1} & \textbf{71.1} & \textbf{84.2} & \textbf{88.9} & +2.5 \\ 
        DINOv2 & \checkmark & Self-Attention & $840, 60$ & 13.9 & \underline{66.4} & \underline{80.9} & \underline{87.1} & -0.8 \\ 
        DINOv2 & \checkmark & Self+Cross Attention & $840, 60$ & \underline{14.5} & 66.3 & \underline{80.9} & 87.0 & -0.8 \\ 
        \midrule
        \textcolor{gray}{DINOv2} & \textcolor{gray}{\checkmark} & \textcolor{gray}{-} & \textcolor{gray}{$448, 32$} & \textcolor{gray}{7.1} & \textcolor{gray}{56.2} & \textcolor{gray}{76.6} & \textcolor{gray}{84.3} & \textcolor{gray}{--} \\ 
        DINOv2 & \checkmark & ResNet Bottleneck Block & $448, 32$ & 6.3 & 57.9 & 79.8 & 86.8 & +3.2 \\ 
        DINOv2 & \checkmark & Self-Attention & $448, 32$ & 7.6 & 57.2 & 77.2 & 85.1 & +0.6 \\ 
        DINOv2 & \checkmark & Self+Cross Attention & $448, 32$ & 7.8 & 58.2 & 78.4 & 86.0 & +1.8 \\ 
        \midrule
        \textcolor{gray}{DINOv2} & \textcolor{gray}{\checkmark} & \textcolor{gray}{-} & \textcolor{gray}{$224, 16$} & \textcolor{gray}{1.9} & \textcolor{gray}{30.5} & \textcolor{gray}{58.2} & \textcolor{gray}{72.1} & \textcolor{gray}{--} \\ 
        DINOv2 & \checkmark & ResNet Bottleneck Block & $224, 16$ & 1.7 & 29.8 & 60.4 & 75.2 & +2.2 \\ 
        DINOv2 & \checkmark & Self-Attention & $224, 16$ & 2.1 & 32.0 & 60.8 & 74.6 & +2.6 \\ 
        DINOv2 & \checkmark & Self+Cross Attention & $224, 16$ & 2.2 & 33.0 & 61.0 & 74.5 & +2.8 \\
        \bottomrule
    \end{tabular}
    }
    \end{center}
    \vspace{-15pt}
\end{table*}

\begin{table*}[t]
    \begin{center}
    \caption{Evaluation of neighbourhood consensus module on the SPair-71k dataset. FT.: Fine-tune.
    The methods are categorized into two types: frozen DINOv2 backbone and joint training of the last layer of DINOv2 and the cost aggregator.
    The baseline methods are indicated in gray.
    NeighConsensus (Ori.) denotes the original implementation of NeighConsensus in NC-Net \cite{NC-Net}.
    }
    \vspace{-10pt}
    \label{tab:appendix_cost}
    \renewcommand{\arraystretch}{1.2} % Row spacing
    \scalebox{1.0}{
    \begin{tabular}{c|c|c
        |>{\centering\arraybackslash}p{1 cm}
        |cccc|c}
        \toprule
        \multirow{3}{*}{Backbone} & 
        \multirow{3}{*}{FT.} & 
        \multirow{3}{*}{Cost Aggregator} & 
        \multirow{3}{*}{Reso.} & 
        \multicolumn{4}{c}{SPair-71k} &  \\ 
        & & & & \multicolumn{4}{c}{PCK @ $\alpha_{\text{bbox}}$} & \\ 
        & & & & 0.01 & 0.05 & 0.1 & 0.15 & $\Delta$PCK$\alpha$ 0.1 \\ 
        \midrule
        \textcolor{gray}{DINOv2} & \textcolor{gray}{$\times$} & \textcolor{gray}{-} & \textcolor{gray}{840, 60} & \textcolor{gray}{7.3} & \textcolor{gray}{40.0} & \textcolor{gray}{54.4} & \textcolor{gray}{62.8} & \textcolor{gray}{--} \\ 
        DINOv2 & $\times$ & Match2Match & 840, 60 & 6.8 & 39.5 & 54.1 & 62.6 & -0.3 \\ 
        DINOv2 & $\times$ & NeighConsensus & 840, 60 & 6.2 & 45.8 & 61.9 & 69.3 & +7.5 \\ 
        \midrule
        \textcolor{gray}{DINOv2} & \textcolor{gray}{$\times$} & \textcolor{gray}{-} & \textcolor{gray}{$448,32$} & \textcolor{gray}{3.7} & \textcolor{gray}{35.4} & \textcolor{gray}{52.9} & \textcolor{gray}{62.2} & \textcolor{gray}{--} \\ 
        DINOv2 & $\times$ & CATs & 448, 32 & 2.6 & 30.8 & 49.5 & 59.7 & -3.4 \\ 
        DINOv2 & $\times$ & Match2Match & 448, 32 & 3.6 & 34.8 & 52.2 & 61.7 & -0.7 \\ 
        DINOv2 & $\times$ & NeighConsensus & 448, 32 & 2.8 & 33.4 & 56.5 & 67.7 & +3.6 \\ 
        \midrule
        \textcolor{gray}{DINOv2} & \textcolor{gray}{$\times$} & \textcolor{gray}{-} & \textcolor{gray}{224, 16} & \textcolor{gray}{1.1} & \textcolor{gray}{20.9} & \textcolor{gray}{43.1} & \textcolor{gray}{56.2} & \textcolor{gray}{--} \\ 
        DINOv2 & $\times$ & CATs & 224, 16 & 1.0 & 18.4 & 40.2 & 53.4 & -2.9 \\ 
        DINOv2 & $\times$ & Match2Match & 224, 16 & 1.1 & 20.8 & 42.9 & 55.4 & -0.2 \\ 
        DINOv2 & $\times$ & NeighConsensus & 224, 16 & 0.9 & 18.0 & 43.4 & 60.5 & +0.3 \\ 
        DINOv2 & $\times$ & NeighConsensus (Ori.) & 224, 16 & 1.1 & 20.4 & 47.1 & 63.3 & +4.0 \\ 
        \midrule
        \textcolor{gray}{DINOv2} & \textcolor{gray}{\checkmark} & \textcolor{gray}{-} & \textcolor{gray}{840, 60} & \textcolor{gray}{15.0} & \textcolor{gray}{67.4} & \textcolor{gray}{81.7} & \textcolor{gray}{87.1} & \textcolor{gray}{--} \\ 
        DINOv2 & \checkmark & Match2Match & 840, 60 & \textbf{13.9} & \underline{68.3} & \underline{83.3} & \underline{88.6} & +1.6 \\ 
        DINOv2 & \checkmark & NeighConsensus & 840, 60 & \underline{12.7} & \textbf{70.2} & \textbf{85.2} & \textbf{89.6} & +3.5 \\ 
        \midrule
        \textcolor{gray}{DINOv2} & \textcolor{gray}{\checkmark} & \textcolor{gray}{-} &\textcolor{gray}{448, 32} & \textcolor{gray}{7.1} & \textcolor{gray}{56.2} & \textcolor{gray}{76.6} & \textcolor{gray}{84.3} & \textcolor{gray}{--} \\ 
        DINOv2 & \checkmark & CATs & 448, 32 & 3.2 & 42.6 & 71.4 & 82.4 & -5.2 \\ 
        DINOv2 & \checkmark & Match2Match & 448, 32 & 6.3 & 55.4 & 77.8 & 85.9 & +1.2 \\ 
        DINOv2 & \checkmark & NeighConsensus & 448, 32 & 5.0 & 54.7 & 79.4 & 87.3 & +2.8 \\ 
        \midrule
        \textcolor{gray}{DINOv2} & \textcolor{gray}{\checkmark} & \textcolor{gray}{-} & \textcolor{gray}{224, 16} & \textcolor{gray}{1.9} & \textcolor{gray}{30.5} & \textcolor{gray}{58.2} & \textcolor{gray}{72.1} & \textcolor{gray}{--} \\ 
        DINOv2 & \checkmark & CATs & 224, 16 & 1.3 & 23.1 & 52.1 & 68.8 & -6.1 \\ 
        DINOv2 & \checkmark & Match2Match & 224, 16 & 1.8 & 30.7 & 59.8 & 73.9 & +1.6 \\ 
        DINOv2 & \checkmark & NeighConsensus & 224, 16 & 1.9 & 31.5 & 62.4 & 77.1 & +4.2 \\ 
        DINOv2 & \checkmark & NeighConsensus (Ori.) & 224, 16 & 2.5 & 36.4 & 66.1 & 79.2 & +7.9 \\
        \bottomrule
    \end{tabular}
    }
    \end{center}
    \vspace{-17pt}
\end{table*}
\label{sec:appd_exp}
\rev{In \Cref{sec:feature_enhancement} and \Cref{sec:cost_aggregation}, we conduct experiments to explore how feature enhancement techniques and matching refinement modules may further improve matching performance. 
In this section, we expand these experiments with more input resolutions.}

\subsection{Feature Enhancement Evaluation}
\label{sec:appd_feature_enhancement}
We compare the performance of CNN and Transformer feature enhancement modules across different resolutions, experimental results are provided in~\Cref{tab:appendix_bottleneck}.

All feature enhancement methods generally improve the performance compared to the baseline (no feature enhancement) across different resolutions. 
The improvements are more significant in the frozen backbone setting compared to the fine-tuned setting.
\rev{The absolute performance improvement from all enhancement modules is largest at the highest resolution (840, 60) and progressively decreases as resolution drops. This highlights the importance of detailed input features for the enhancement process.}

\subsection{Cost Aggregation Evaluation}
\label{sec:appd_cost_aggregation}

We compare the performance of the Neighbourhood Consensus, CATs Aggregator, and Match-to-Match attention module. 
The detailed evaluation results are provided in~\Cref{tab:appendix_cost}.
Due to the computational cost of high-resolution inputs, the original Neighbourhood Consensus setting is applied only at the 224 resolution. 
For the 448 and 840 resolutions, we use three layers of $3 \times 3 \times 3 \times 3$ filters while reducing the number of intermediate channels to half of the original configuration.\\
\rev{The experimental results demonstrate that NeighConsensus consistently outperforms other cost aggregators across different resolutions, and the original implementation of NeighConsensus (Ori.) also shows competitive performance, especially at lower resolutions.
Regarding the CATs cost aggregator, it exhibits specific limitations: 
First, it cannot be applied to high-resolution inputs (840$\times$840) due to computational constraints. 
Second, at 448$\times$448 resolution, it performs substantially worse than both NeighConsensus and Match2Match, with PCK@0.1 dropping by 3.4\% and 5.2\% in frozen and fine-tuned settings respectively, making it the least effective among the three aggregators.}

\vspace{-10pt}
\section{\rev{Qualitative Comparison with GeoAware-SC}}
\label{sec:appd_qualitative}
\begin{figure*}[htbp]
\centering
% Add column subtitles
\makebox[0.24\textwidth]{GeoAware-SC \cite{GeoAware-SC}}
\makebox[0.24\textwidth]{Ours}
\makebox[0.24\textwidth]{GeoAware-SC \cite{GeoAware-SC}}
\makebox[0.24\textwidth]{Ours}\\[5pt]

\includegraphics[width=0.24\textwidth]{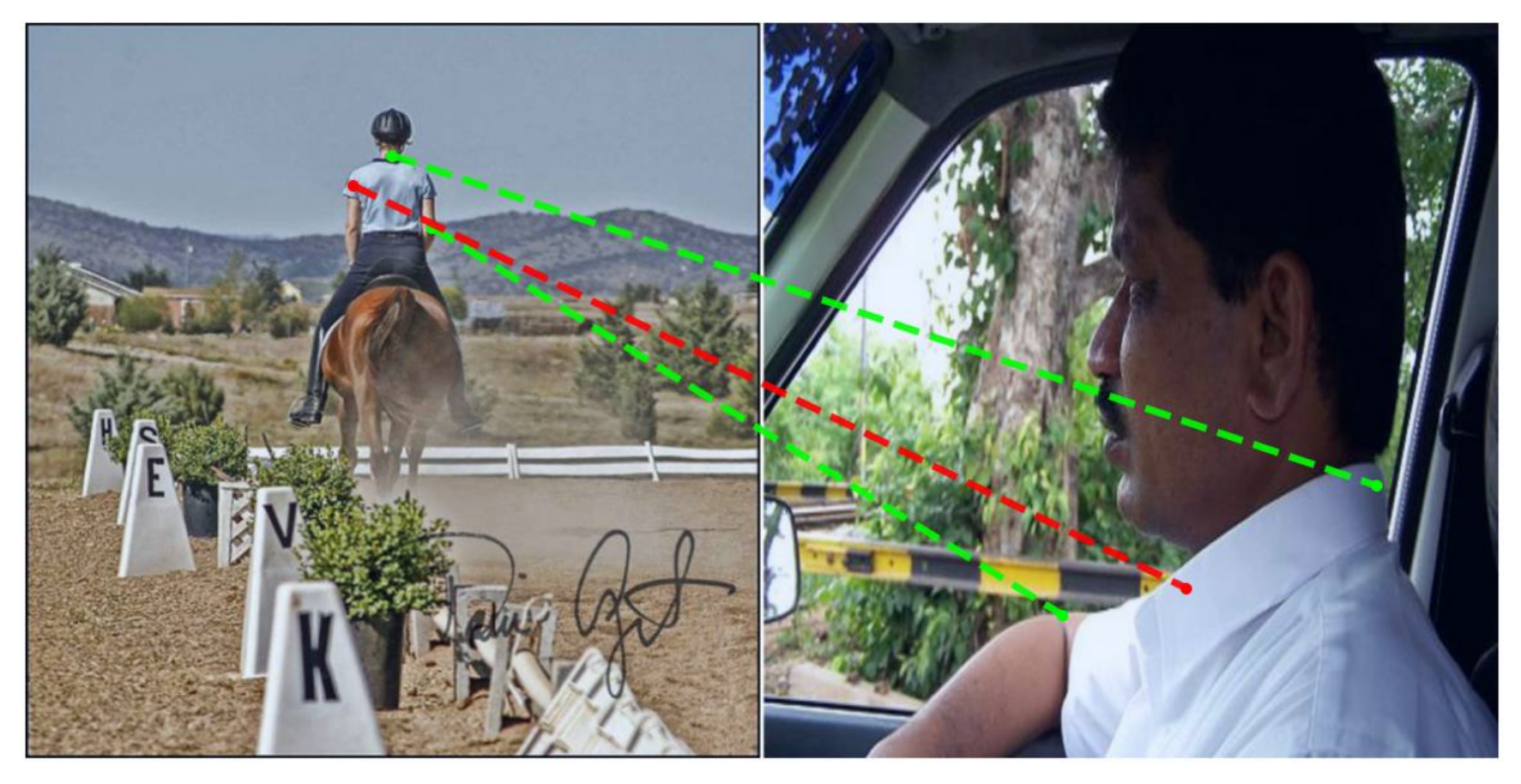}
\includegraphics[width=0.24\textwidth]{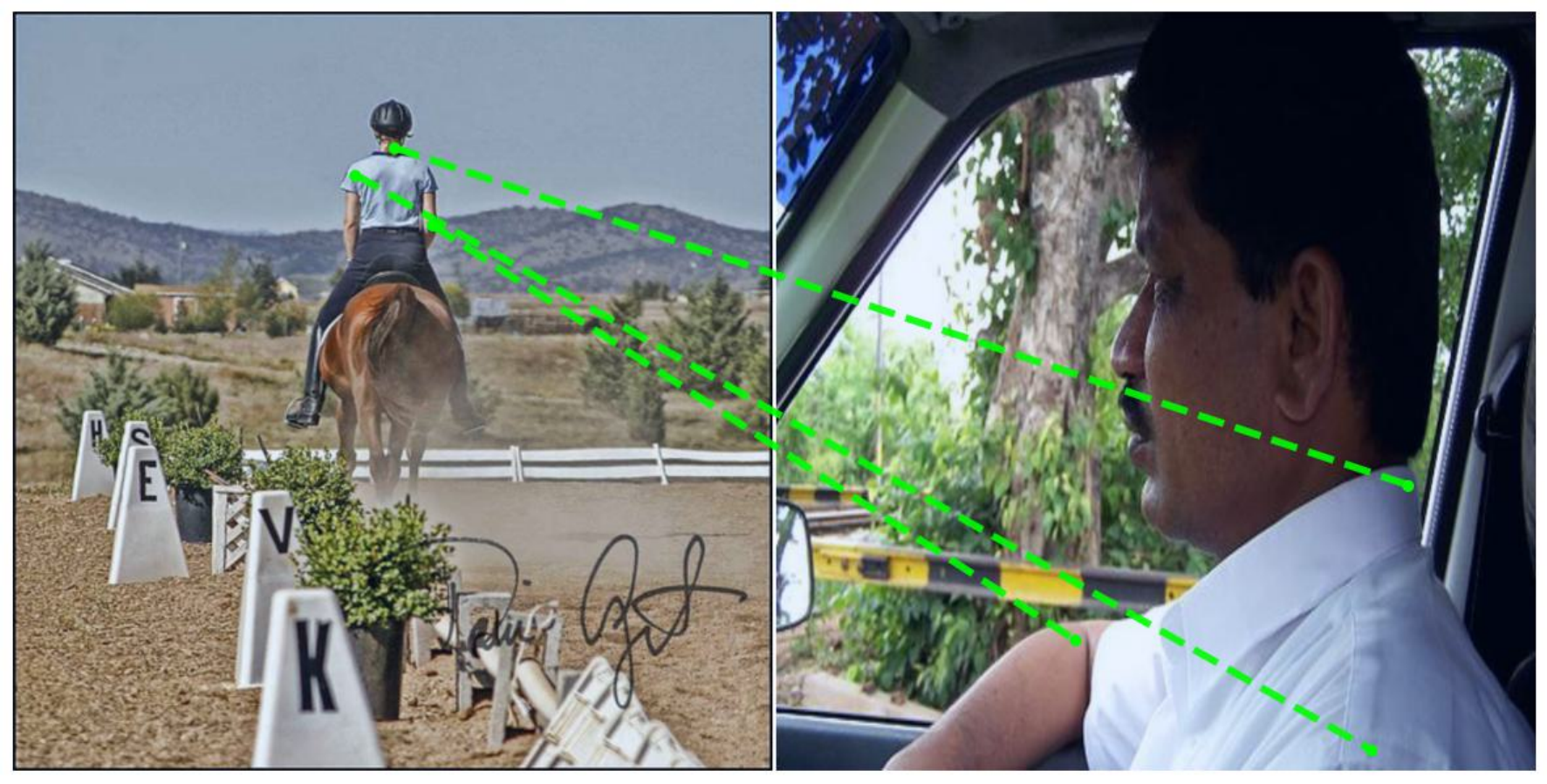}
\includegraphics[width=0.24\textwidth]{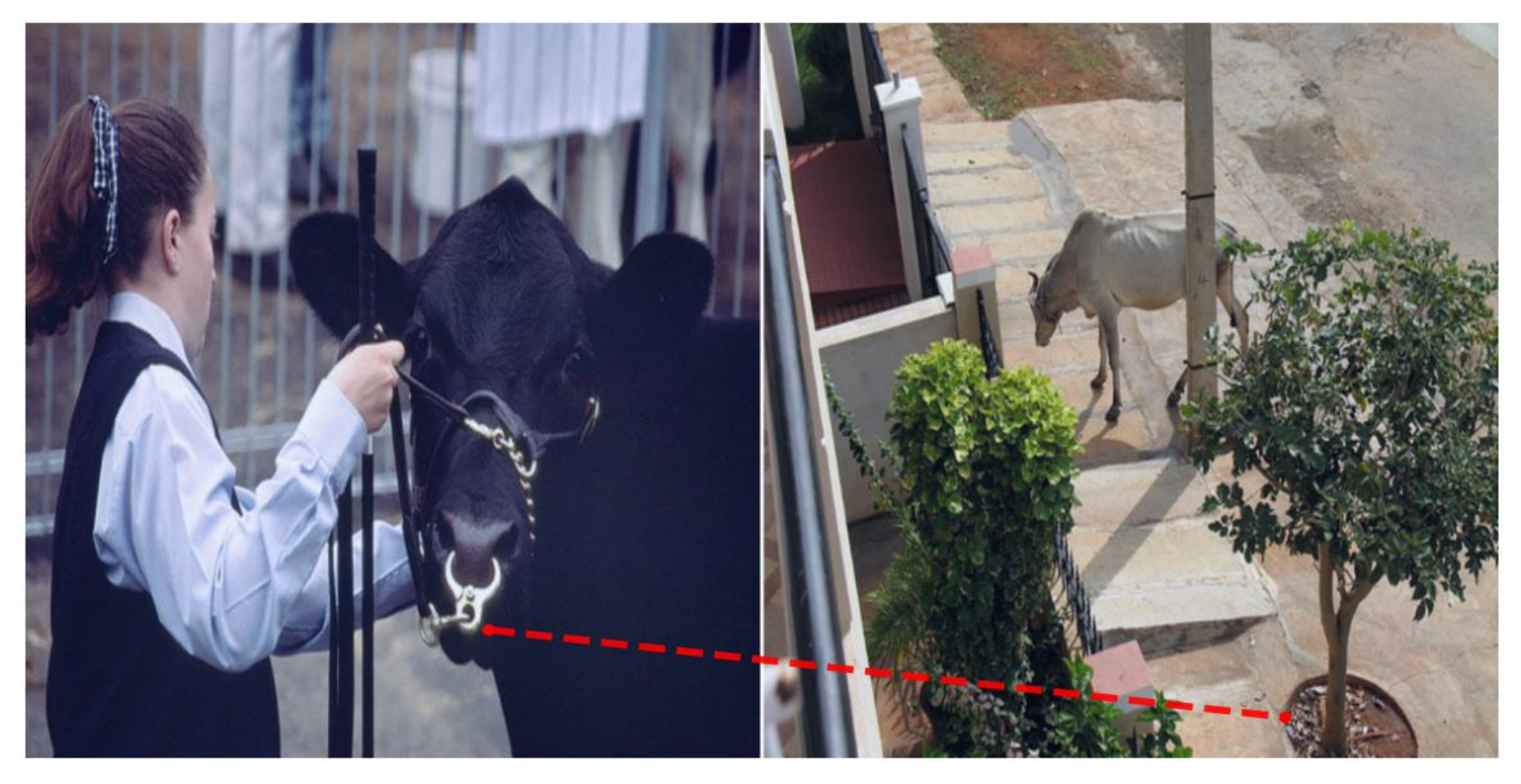}
\includegraphics[width=0.24\textwidth]{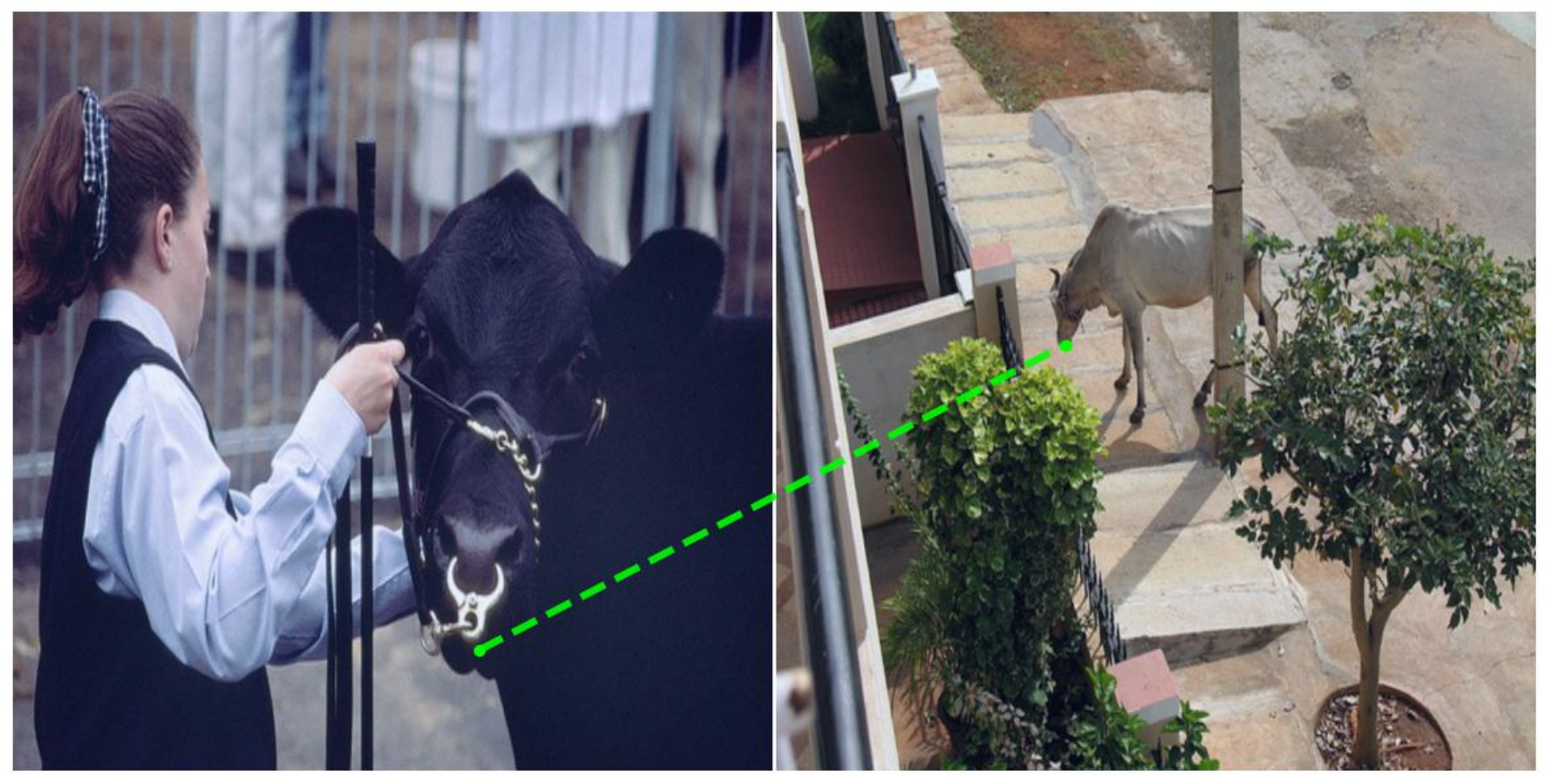}\\
\includegraphics[width=0.24\textwidth]{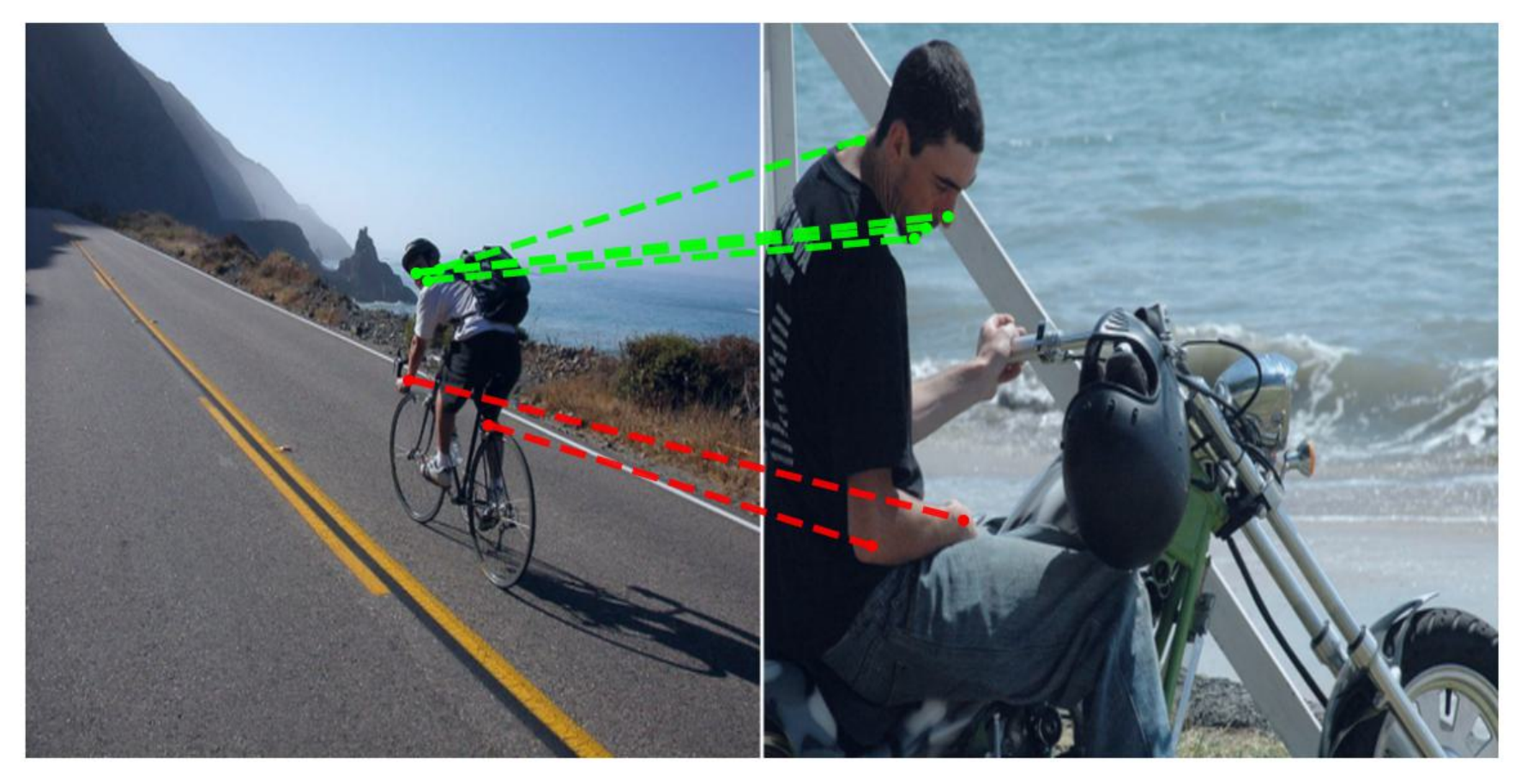}
\includegraphics[width=0.24\textwidth]{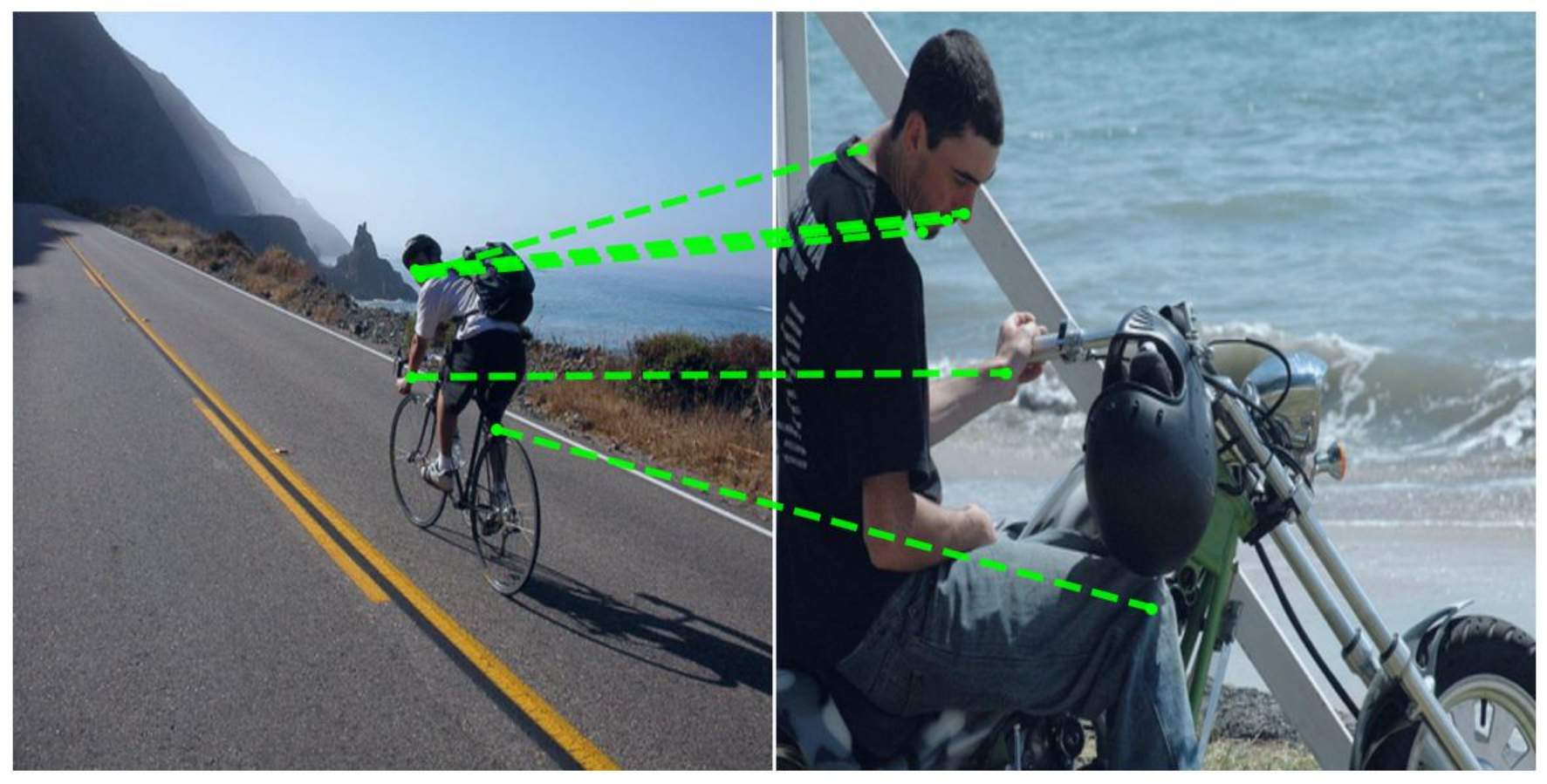}
\includegraphics[width=0.24\textwidth]{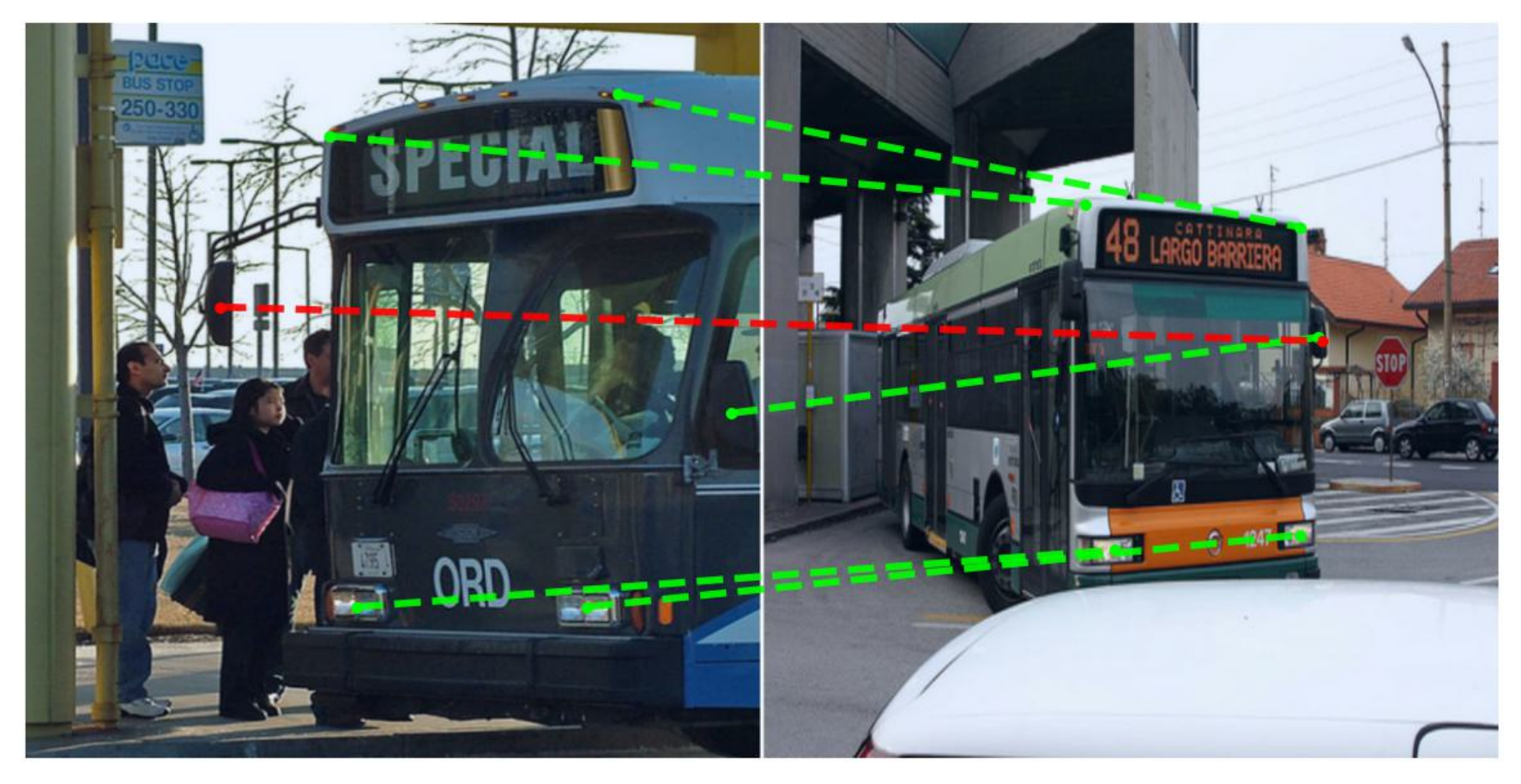}
\includegraphics[width=0.24\textwidth]{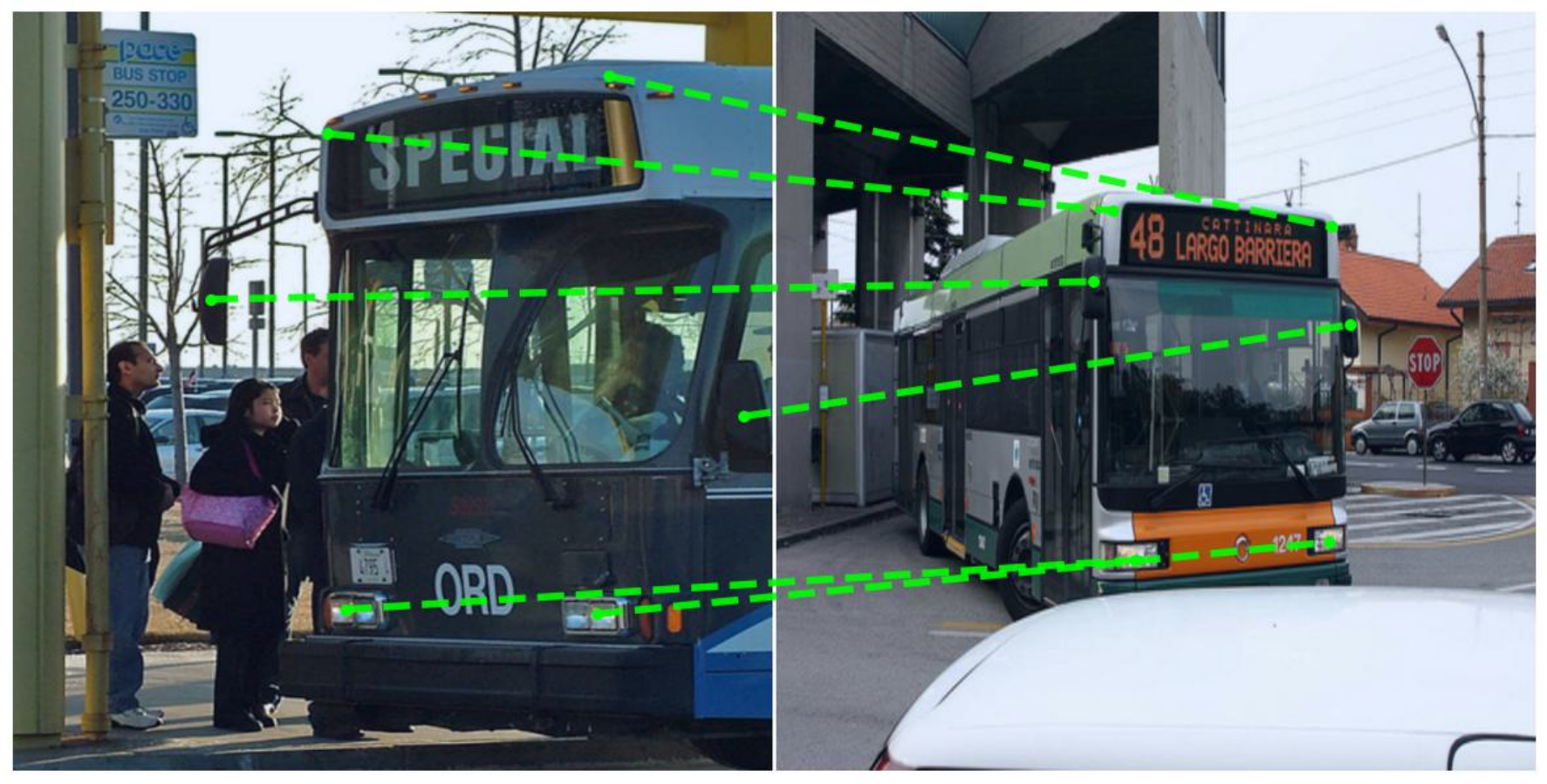}\\
\includegraphics[width=0.24\textwidth]{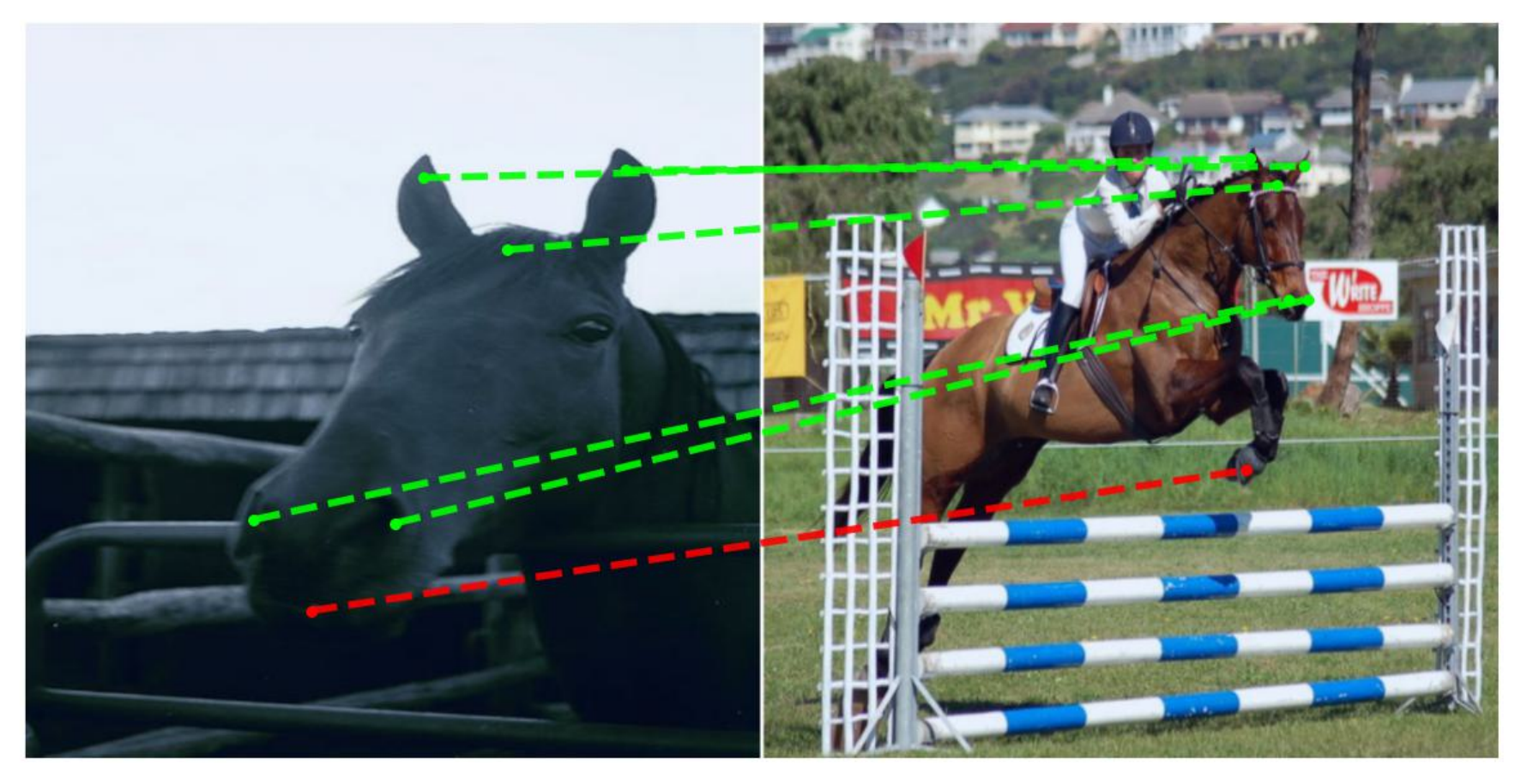}
\includegraphics[width=0.24\textwidth]{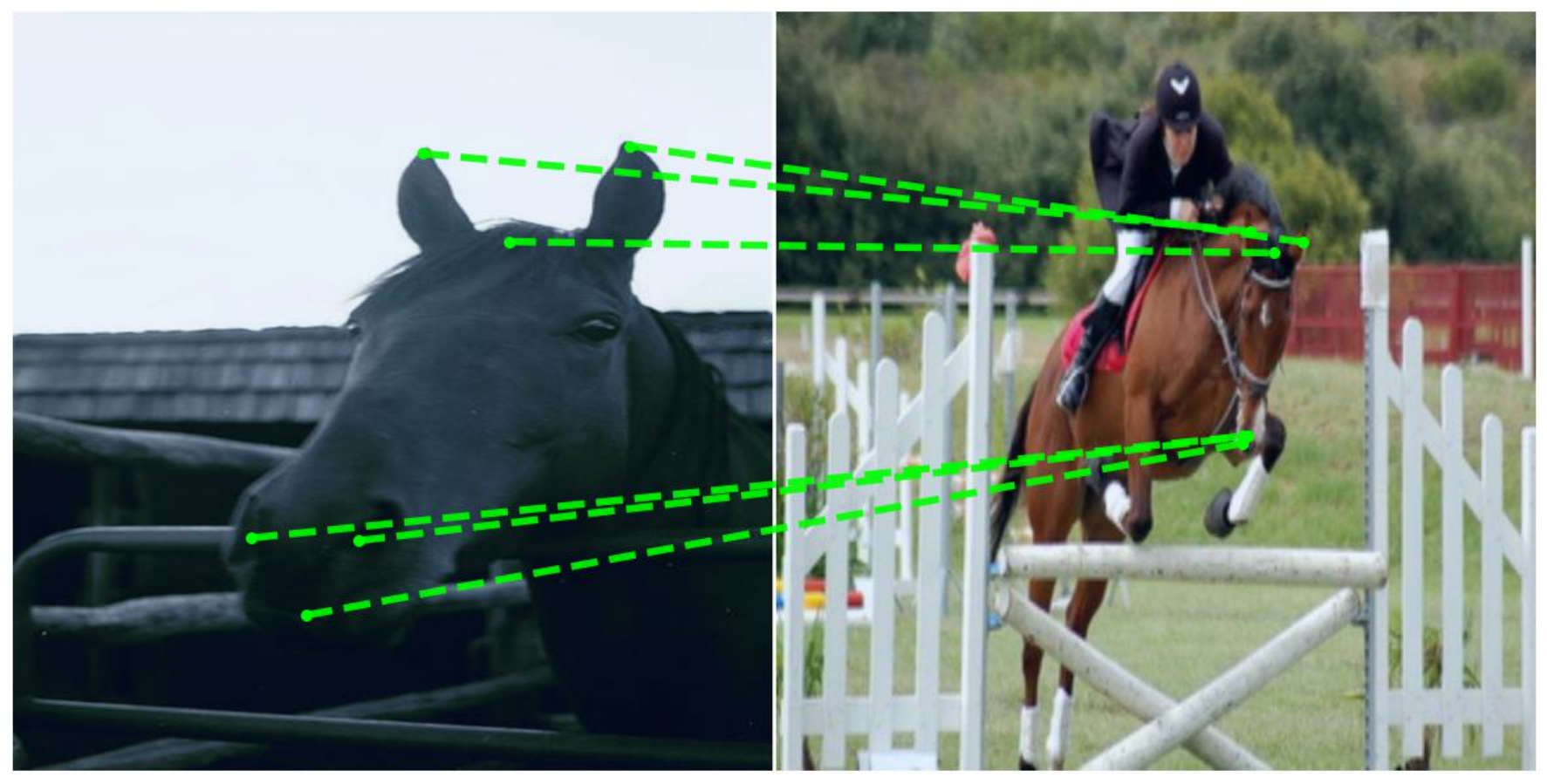}
\includegraphics[width=0.24\textwidth]{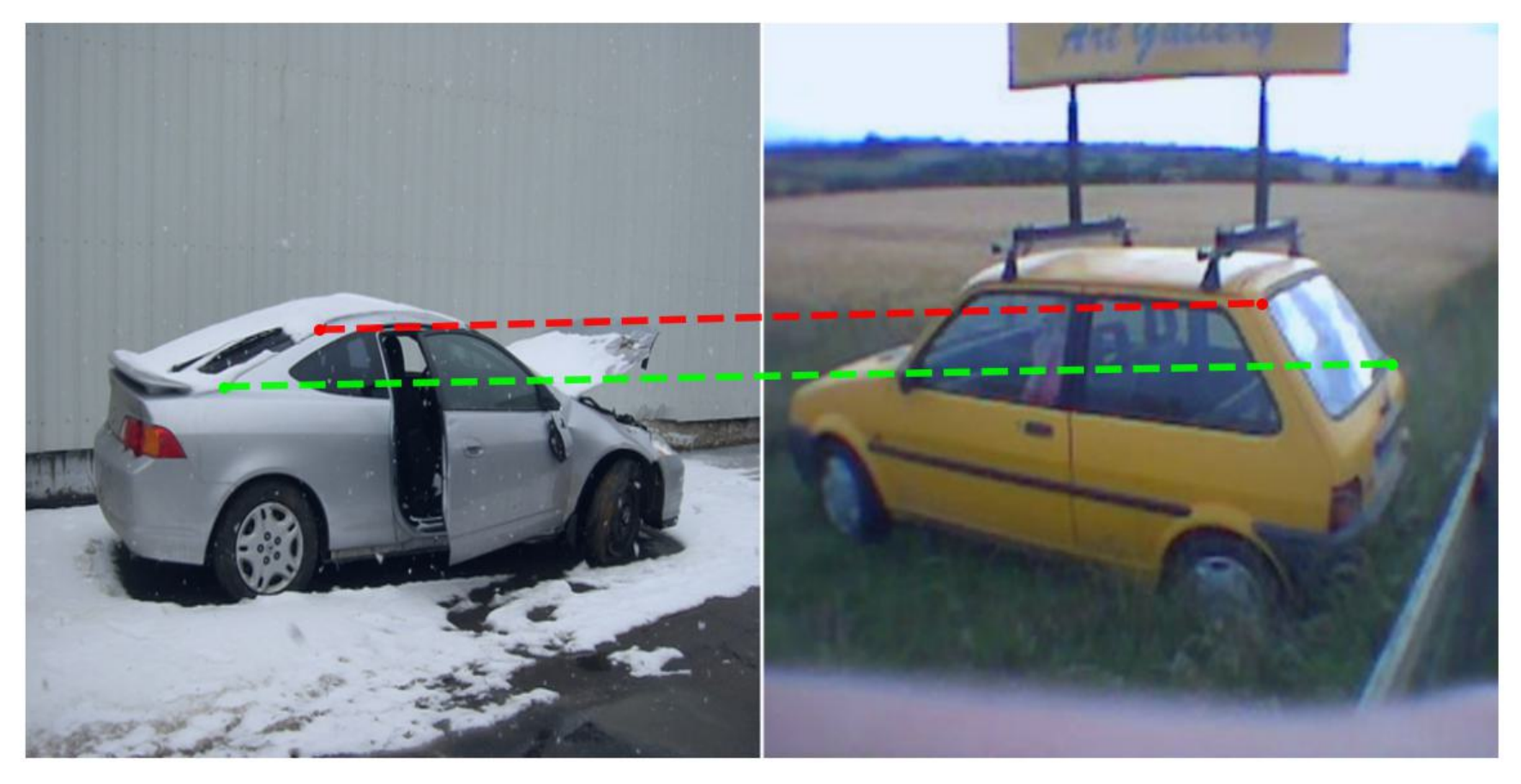}
\includegraphics[width=0.24\textwidth]{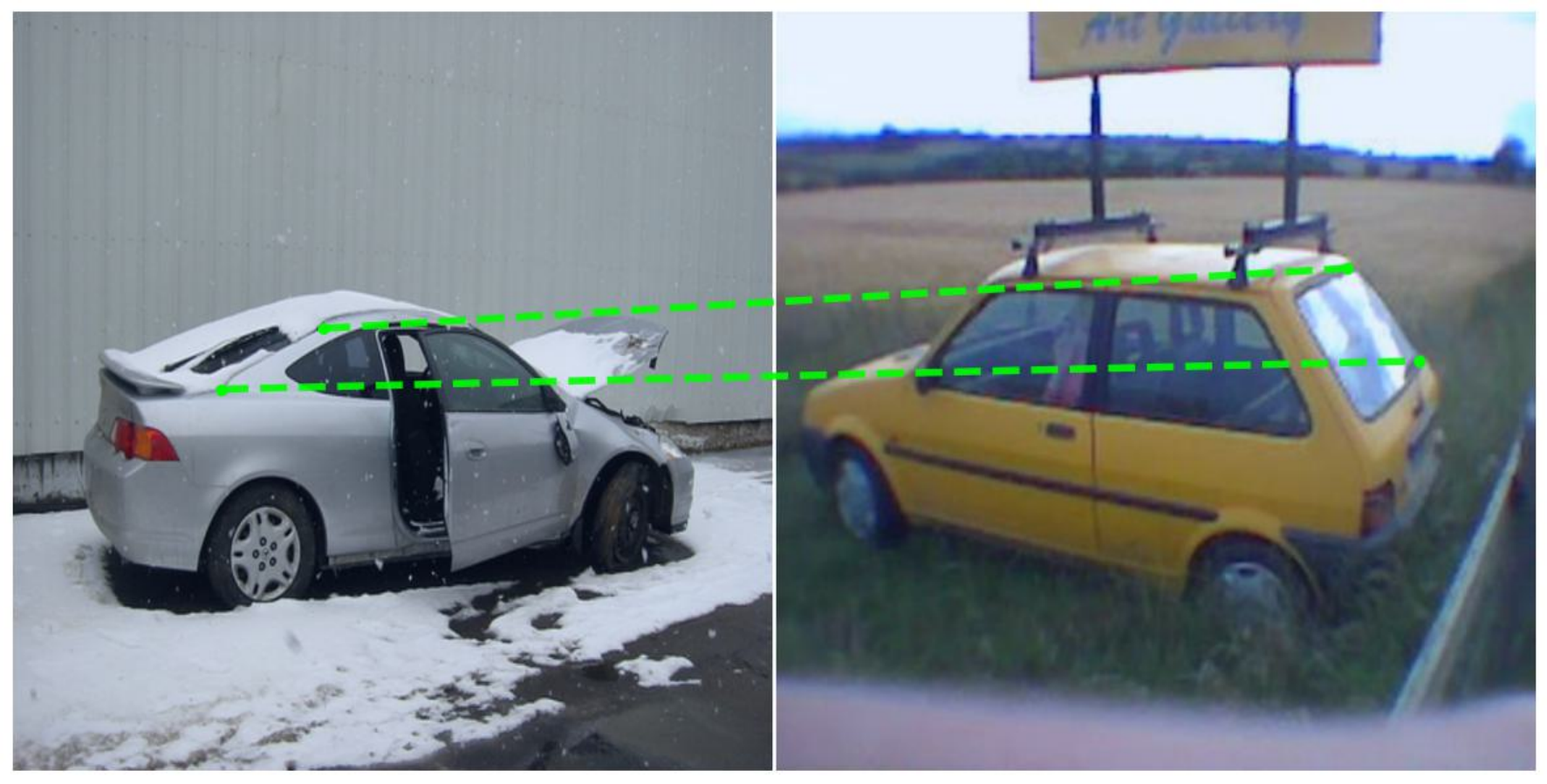}\\
\includegraphics[width=0.24\textwidth]{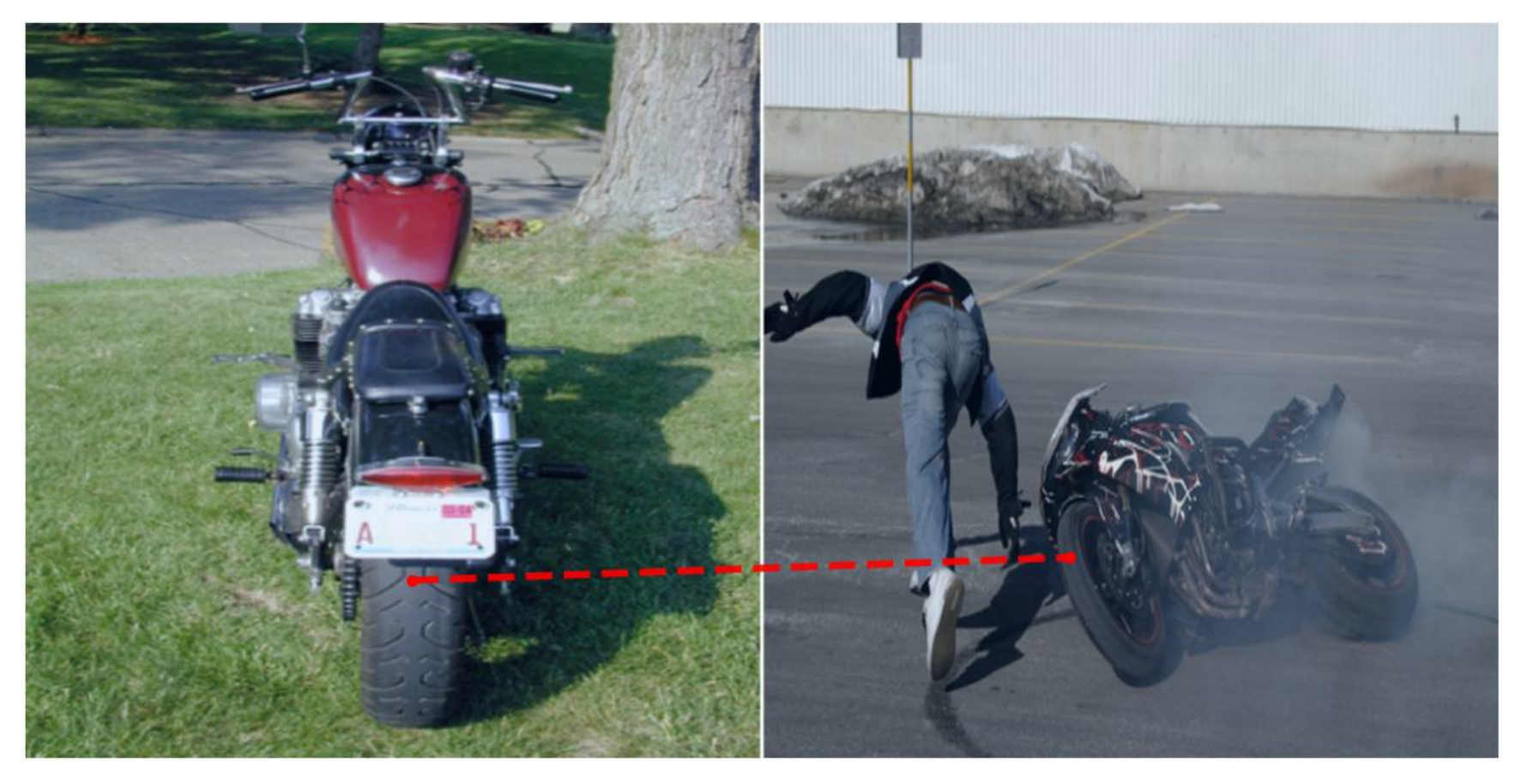}
\includegraphics[width=0.24\textwidth]{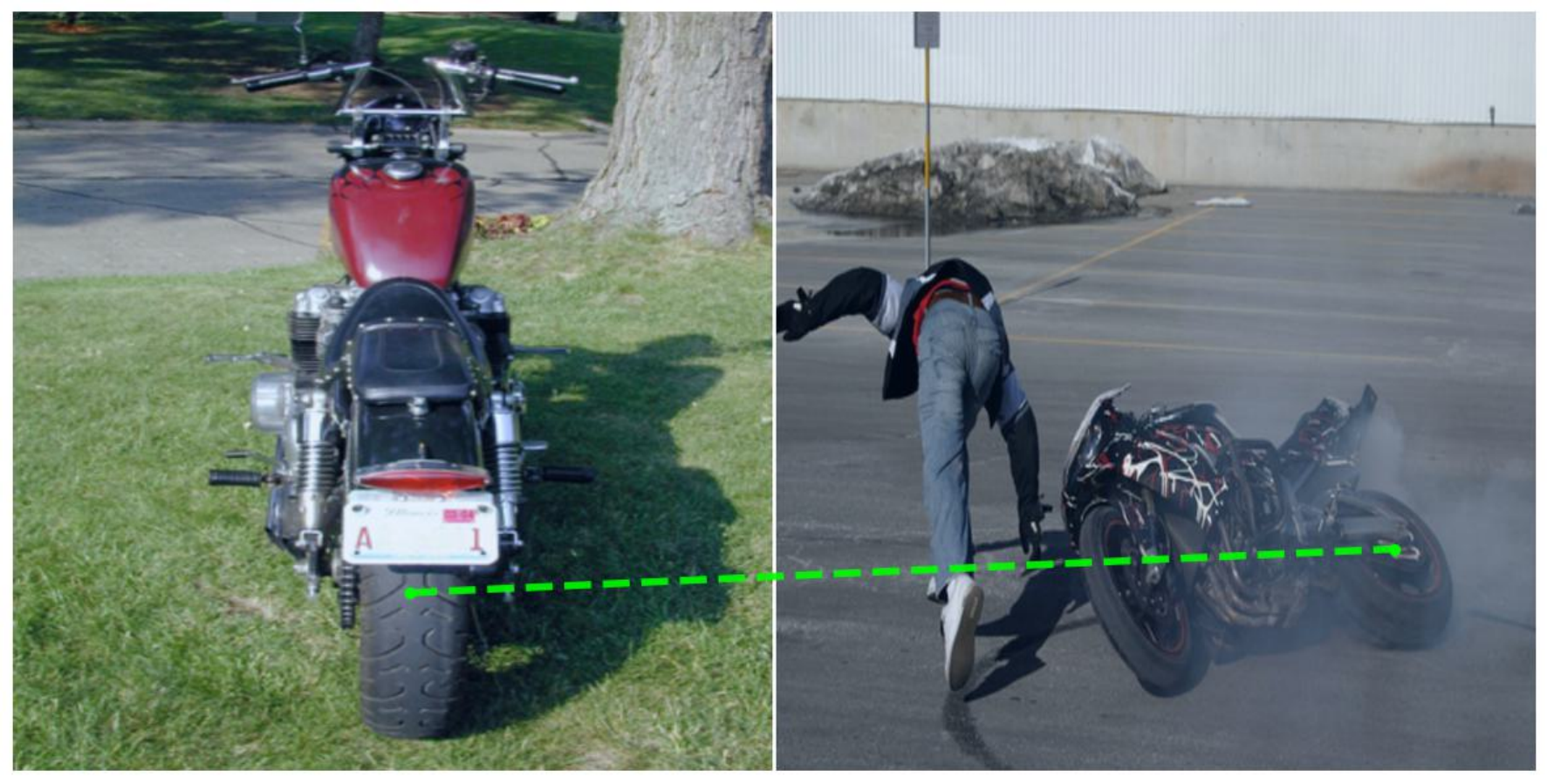}
\includegraphics[width=0.24\textwidth]{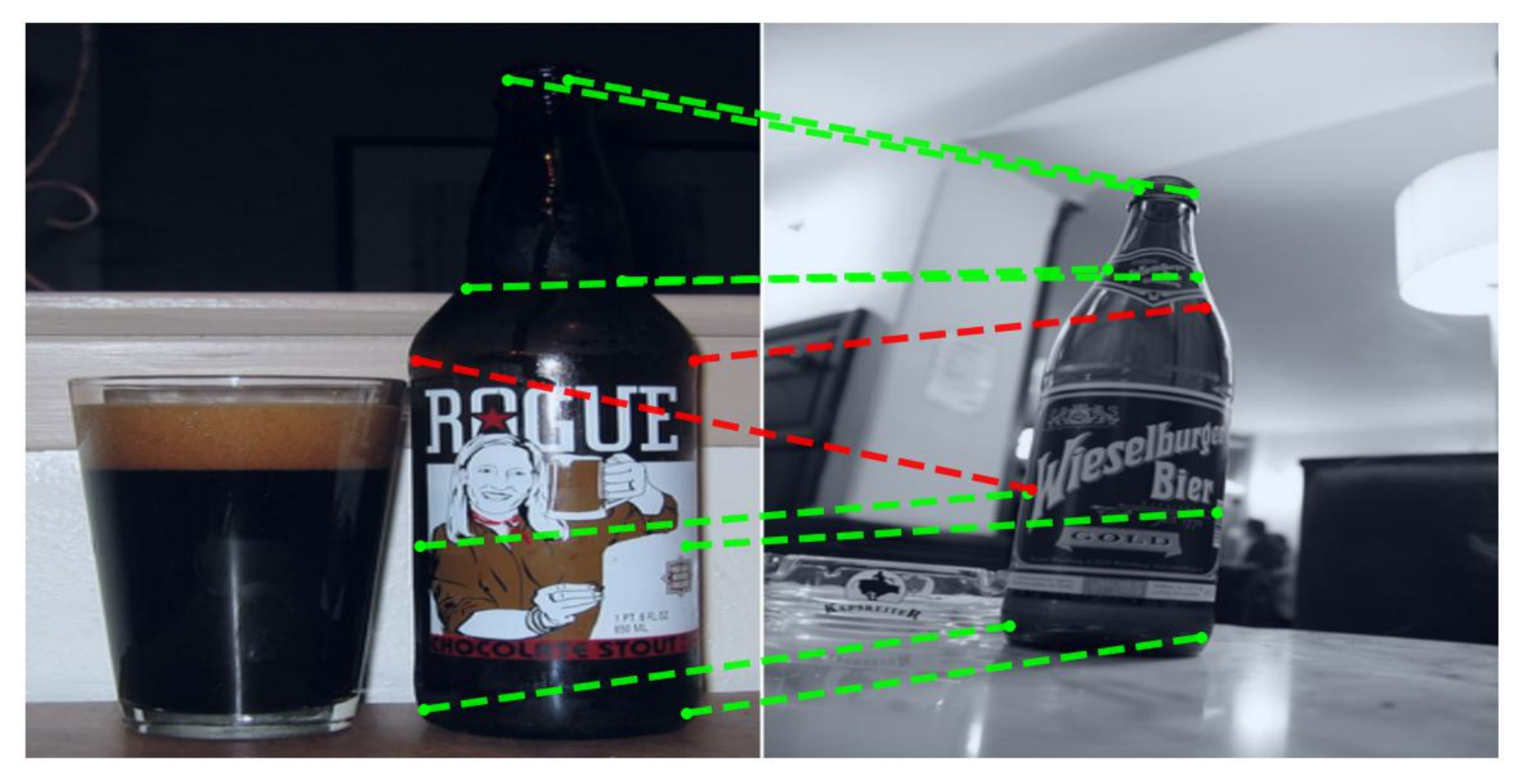}
\includegraphics[width=0.24\textwidth]{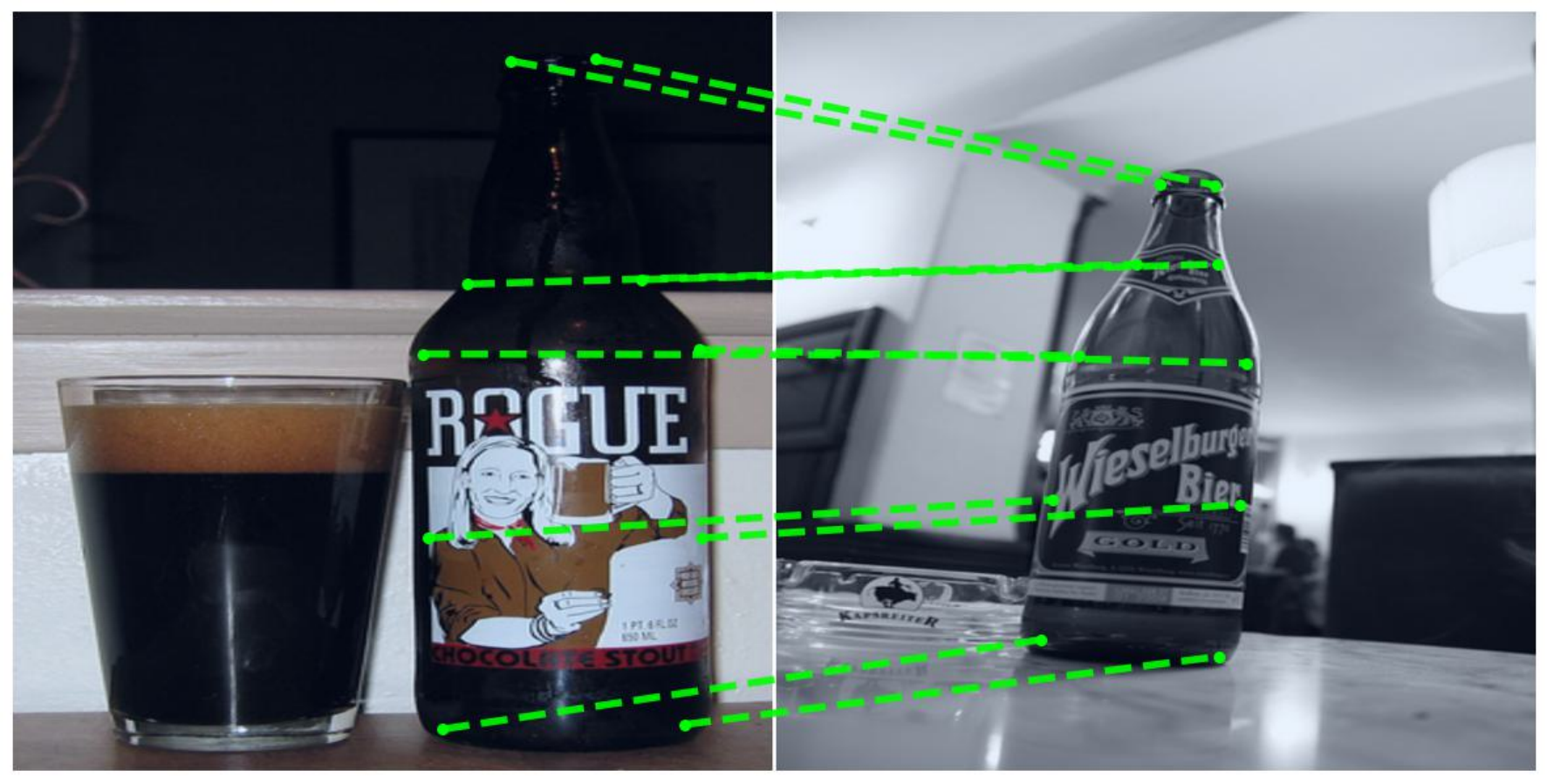}
\caption{Visualization of feature matching results across diverse scenarios. Our proposed baseline successfully handles significant appearance variations and geometric deformations while maintaining reliable correspondences across various object categories.}
\label{fig:qualitative}
\end{figure*}
\rev{In this section, we qualitatively compare our proposed method with the state-of-the-art method GeoAware-SC \cite{GeoAware-SC}.} Our proposed method demonstrates superior robustness across a wide range of real-world scenarios. As shown in \Cref{fig:qualitative}, we visualize the matching results on a wide range of categories including animals (horses, dogs, birds), vehicles (motorcycles, buses, aircraft), and everyday objects. The method successfully handles challenging cases with dramatic changes in viewpoint, illumination, and appearance variations. Particularly noteworthy is its ability to establish reliable correspondences despite significant geometric deformations, such as in the cases of moving animals and vehicles captured from different angles.

\section{Performance Evaluation of DINOv3}
\label{sec:appd_dinov3}
\rev{
DINOv3 \cite{DINOv3} is a vision foundation model trained through large-scale self-supervised learning, requiring no manual annotations, and capable of powerful dense and global feature representations.
For our evaluation, we select the ViT-L/16 distilled pretrained on LVD-1689M dataset as the DINOv3 backbone for evaluation.
We conduct zero-shot evaluation, with results in \Cref{tab:appendix_zero_backbone}. 
DINOv3 surpasses DINOv2 across all resolutions, achieving a PCK@0.1 of 58.4\% at 960$\times$960 resolution.
Interestingly, while DINOv3 is stronger than DINOv2 alone, the SD2-1+DINOv3 fusion (58.1\%) performs slightly worse than the SD2-1+DINOv2 fusion (60.4\%), indicating that the fusion of these features is not complementary, which is worth future research. 
}

\rev{
Subsequently, we fine-tune the final four blocks of the DINOv3 backbone. This straightforward adaptation establishes a new state-of-the-art on both the SPair-71k and AP-10K datasets (in \Cref{tab:appendix_finetune_backbone} and \Cref{tab:appendix_ap10k}), outperforming all prior methods.
}
\begin{table}[htbp]
    \begin{center}
    \caption{Evaluation of zero-shot feature backbones on the SPair-71k dataset. Reso.: Image Resolution, Feature Map Size.}
    \vspace{-7pt}
    \label{tab:appendix_zero_backbone}
    \renewcommand{\arraystretch}{1.2} % Row spacing
    \scalebox{1.0}{
    \begin{tabular}{c
        |>{\centering\arraybackslash}p{1.7cm}
        |cccc}
        \toprule
        \multirow{3}{*}{Backbone} & 
        \multirow{3}{*}{Reso.} & 
        \multicolumn{4}{c}{SPair-71k} \\
        & & \multicolumn{4}{c}{PCK @ $\alpha_{\text{bbox}}$} \\
        & & 0.01 & 0.05 & 0.1 & 0.15 \\
        \midrule
        DINOv2 & 840, 60 & \underline{7.3} & 40.0 & 54.4 & 62.8 \\
        DINOv2 & 448, 32 & 3.7 & 35.4 & 52.9 & 62.2 \\
        DINOv2 & 224, 16 & 1.1 & 20.9 & 43.1 & 56.2 \\
        \midrule
        SD2-1+DINOv2 & (960,840), 60 & \textbf{9.7} & \textbf{47.9} & \textbf{60.4} & \underline{67.0} \\
        SD2-1+DINOv2 & (512,448), 32 & 4.7 & 40.3 & 56.5 & 64.1 \\
        SD2-1+DINOv2 & (256,224), 16 & 1.3 & 21.4 & 41.9 & 53.1 \\
        \midrule
        DINOv3 & 960, 60 & 6.9 & \underline{41.7} & \underline{58.4} & \underline{67.0} \\
        DINOv3 & 512, 32 & 3.6 & 35.5 & 54.9 & 64.4 \\
        DINOv3 & 256, 16 & 1.2 & 21.3 & 45.3 & 59.3 \\
        \midrule
        SD2-1+DINOv3 & 960, 60 & 6.3 & 41.4 & 58.1 & \textbf{67.4} \\
        SD2-1+DINOv3 & 512, 32 & 3.7 & 35.9 & 55.5 & 64.9 \\
        SD2-1+DINOv3 & 256, 16 & 1.2 & 21.5 & 45.6 & 59.4 \\
        \bottomrule
    \end{tabular}
    }
    \end{center}
    \vspace{-15pt}
\end{table}

\begin{table}[!t]
    \begin{center}
    \caption{Evaluation of fine-tuning feature backbones on the SPair-71k dataset. Reso.: Image Resolution, Feature Map Size.}
    \vspace{-10pt}
    \label{tab:appendix_finetune_backbone}
    \renewcommand{\arraystretch}{1.2} % Row spacing
    \scalebox{1.0}{
    \begin{tabular}{c|c|cccc}
        \toprule
        \multirow{3}{*}{Backbone} & 
        \multirow{3}{*}{Reso.} & 
        \multicolumn{4}{c}{SPair-71k} \\
        & & \multicolumn{4}{c}{PCK @ $\alpha_{\text{bbox}}$} \\
        & & 0.01 & 0.05 & 0.1 & 0.15 \\
        \midrule
        DINOv2 & 840, 60 & \underline{15.0} & 67.4 & 81.7 & 87.1 \\
        DINOv2 & 448, 32 & 7.1 & 56.2 & 76.6 & 84.3 \\
        DINOv2 & 224, 16 & 1.9 & 30.5 & 58.2 & 72.1 \\
        \midrule
        DINOv3 & 960, 60 & \textbf{21.7} & \textbf{81.3} & \textbf{90.6} & \textbf{93.5} \\
        DINOv3 & 512, 32 & 10.6 & \underline{71.1} & \underline{86.8} & \underline{91.3} \\
        DINOv3 & 256, 16 & 3.2 & 44.4 & 74.3 & 84.2 \\
        \bottomrule
    \end{tabular}
    }
    \end{center}
    \vspace{-10pt}
\end{table}
\begin{table}[!t]
    \begin{center}
    \caption{Performance comparison of supervised methods on the AP-10K Dataset. Reso.: Image Resolution}
    \vspace{-5pt}
    \label{tab:appendix_ap10k}
    \renewcommand{\arraystretch}{1.2} % Row spacing
    \scalebox{1.0}{
    \begin{tabular}{c|c
        |cccc}
        \toprule
        \multirow{3}{*}{Methods} & 
        \multirow{3}{*}{Reso.} & 
        \multicolumn{4}{c}{AP-10K} \\
        & & \multicolumn{4}{c}{PCK @ $\alpha_{\text{bbox}}$} \\
        & &  0.01 & 0.05 & 0.1 & 0.15 \\
        \midrule
        GeoAware-SC~\cite{GeoAware-SC} & 960,840 & 23.1 & 73.0 & \underline{87.5} & - \\
        \midrule
        \textbf{Ours(DINOv2)}  & 840 & \underline{24.5} & \underline{74.3} & 87.4 & 92.2 \\
        \textbf{Ours(DINOv2+ResNet)} & 840 & 24.0 & 73.1 & 86.7 & 92.3 \\
        \textbf{Ours(DINOv2+NC)} & 840 & 21.1 & 72.8 & 87.4 & \underline{92.6} \\
        \midrule
        \textbf{Ours(DINOv3)} & 960 & \textbf{26.5} & \textbf{79.3} & \textbf{90.3} & \textbf{94.0} \\
        \bottomrule
    \end{tabular}
    }
    \end{center}
    \vspace{-15pt}
\end{table}

\section{Efficiency Analysis}
\label{sec:appd_efficiency}
\begin{table*}[t]
    \centering
    \caption{Comparison of GFLOPs, memory usage, and inference time for various semantic matching methods. The total inference time is listed before the parentheses. The time of the forward pass of the module is in parentheses (·). 
    F.B. denotes feature backbone, F.E. denotes feature enhancement, and M.R. denotes matching refinement.}
    \label{tab:benchmark_efficiency}
    \renewcommand{\arraystretch}{1.2}
    \scalebox{0.94}{
    \begin{tabular}{cc|c|ccc|ccc|ccc}
        \toprule
        & \multirow{3}{*}{Methods}& 
        \multirow{3}{*}{Params} & 
        \multicolumn{3}{c|}{Low Reso. (256/224, 16)} & 
        \multicolumn{3}{c|}{Medium Reso. (512/448, 32)} & 
        \multicolumn{3}{c}{High Reso. (960/840, 60)} \\
        & & & GFLOPs & Mem   & Time & GFLOPs & Mem   & Time & GFLOPs & Mem & Time \\
        & &(M) & & (MB) & (ms) & & (MB) & (ms) & & (MB) & (ms) \\
        \midrule
        \textbf{F.B.} & CLIP & 303.2 & 155.6 & 616.9 & 37.1 (36.0) & - &-&-& -& -&- \\
        & ResNet-101 & 27.5 & 18.4 & 63.5 & 42.2 (41.1) & 73.7 & 91.1 & 38.8 (37.8) & 259.1 & 185.5 & 41.5 (40.4) \\
        & iBOT & 85.8 & 44.0 & 180.3 & 18.4 (17.3) & 175.5 & 263.5 & 26.8 (25.7) & 616.5 & 1134.4 & 149.1 (148.2) \\
        & DINOv1 & 71.6 & 36.7 & 151.7 & 15.4 (14.3) & 146.4 & 235.5 & 24.2 (23.2) & 514.4 & 1106.5 & 125.3 (124.2) \\
        & DINOv2 & 86.6 & 43.9 & 186.9 & 20.5 (19.5) & 175.2 & 306.7 & 36.4 (35.4) & 615.5 & 1493.0 & 157.6 (156.5) \\
        & DINOv3 & 303.2 & 118.7 & 1180.4 & 55.5 (54.3) & 468.1 & 1224.5 & 123.5 (122.0) & 1640.1 & 1368.2 & 610.6 (609.0) \\
        & SD2-1 & 1148.8 & 849.5 & 2302.7 & 128.3 (127.1) & 4230.3 & 2729.6 & 282.3 (280.1) & 14698.0 & 4058.5 & 1001.2 (998.9) \\
        & SD2-1+DINOv2 & 1235.4 & 1153.4 & 2506.2 & 169.9 (167.4) & 4405.5 & 2910.1 & 319.6 (316.5) & 15313.5 & 4260.0 & 1182.8 (1179.6) \\
        & SD2-1+DINOv3 & 1452.0 & 1228.2 & 5800.9 & 212.0 (210.2) & 4698.4 & 6514.1 & 688.5 (686.7) & 16338.1 & 8891.2 & 3480.7 (3478.6) \\
        \midrule
        \textbf{F.E.} & \textcolor{gray}{DINOv2 (Baseline)} & \textcolor{gray}{86.6} & \textcolor{gray}{43.9} & \textcolor{gray}{186.9} & \textcolor{gray}{20.5 (19.5)} & \textcolor{gray}{175.2} & \textcolor{gray}{306.7} & \textcolor{gray}{36.4 (35.4)} & \textcolor{gray}{615.5} & \textcolor{gray}{1493.0} & \textcolor{gray}{157.6 (156.5)} \\
        & ResNet Bottleneck & 87.2 & 43.9 & 200.8 & +3.3 (+3.1) & 175.2 & 331.1 & +7.6 (+7.3) & 615.5 & 1557.6 & +23.3 (+23.0) \\
        & Self-Attention & 98.4 & 50.0 & 223.0 & +5.5 (+5.3) & 199.4 & 357.2 & +9.4 (+9.0) & 700.5 & 1589.2 & +28.8 (+28.6) \\
        & Self+Cross Attention & 98.4 & 50.0 & 223.0 & +6.9 (+6.7) & 199.4 & 357.2 & +9.4 (+9.0) & 700.5 & 1589.2 & +28.8 (+28.5) \\
        \midrule
        \textbf{M.R.} & \textcolor{gray}{DINOv2 (Baseline)} & \textcolor{gray}{86.6} & \textcolor{gray}{43.9} & \textcolor{gray}{186.9} & \textcolor{gray}{20.5 (19.5)} & \textcolor{gray}{175.2} & \textcolor{gray}{306.7} & \textcolor{gray}{36.4 (35.4)} & \textcolor{gray}{615.5} & \textcolor{gray}{1493.0} & \textcolor{gray}{157.6 (156.5)} \\
        & Match2Match & 86.6 & 43.9 & 424.9 & +0.1 (+0.1) & 175.5 & 1543.7 & +65.3 (+65.0) & 618.8 & 14986.3 & +626.1 (+625.9) \\
        & CATs & 103.8 & 44.9 & 378.5 & +0.4 (+0.3) & 199.6 & 564.1 & +22.8 (+22.6) & - & - & - \\ 
        & NeighConsensus & 86.6 & 43.9 & 363.1 & +24.6 (+24.4) & 175.2 & 527.1 & +73.9 (+73.6) & 615.5 & 2489.8 & +577.7 (+577.5) \\
        & NeighConsensus (Ori.) & 86.6 & 43.9 & 365.7 & +46.2 (+45.9) & 175.2 & 696.8 & +199.3 (+199.1) & 615.5 & 4530.5 & +1734.8 (+1734.5) \\
        \bottomrule
    \end{tabular}
    }
\end{table*}
\rev{In this section, we evaluate the computational efficiency of our methods by measuring their GFLOPs, peak GPU memory usage, and inference time across various resolutions. 
Based on the nature of the methods, we categorize them into three groups: feature backbone-based (F.B.), feature enhancement-based (F.E.), and matching refinement-based (M.R.), and report their results in \Cref{tab:benchmark_efficiency}.
In addition to the total running time for obtaining the final correspondences, we also report the time of the forward pass of the module in parentheses (·).
For both feature enhancement and matching refinement methods, we select DINOv2 as the backbone, and report their overhead over the backbone. 
All experiments were performed on a workstation equipped with an Intel(R) Xeon(R) Silver 4314 CPU @ 2.40GHz and a NVIDIA GeForce RTX 3090 GPU.
}

\rev{  
% Feature backbone
According to \Cref{tab:benchmark_efficiency}, the feature backbone primarily determines a method's overall computational cost and efficiency, as seen in the pronounced divergence in efficiency across all metrics between the lightweight ResNet-101 and the heavyweight SD2-1. 
% ResNet-101
ResNet-101, being the most lightweight backbone, achieves the lowest GFLOPs and memory usage across all resolutions, and maintains a fast inference time of around 40 ms even at the highest resolution.
% SD2-1
On the other hand, using generative models as backbones introduces significant computational challenges. Models based on SD2-1, for instance, far exceed all others in parameters, GFLOPs, memory, and inference time, posing a severe challenge for hardware deployment due to this immense feature extraction cost.
% DINOv2
This highlights the importance of finding an effective trade-off. 
Our proposed DINOv2 baseline achieves a rapid inference time of just 20 ms at $224 \times 224$ resolution. Even at $840 \times 840$ resolution, it maintains an efficient 150 ms runtime while delivering a high PCK@0.1 score of 85.1\%.
This well-rounded profile of high efficiency and strong performance makes it highly suitable for real-time applications.
}
\rev{
% Feature enhancement
Regarding the feature enhancement modules, their cost is relatively moderate, with different modules showing similar resource usage. 
This suggests that they can be conceptualized as ``optional plug-ins'' for improving feature quality, which users can choose to add or omit based on specific resource constraints.
% matching refinement
In contrast, integrating a matching refinement module into the pipeline imposes a substantial computational burden, especially with large image resolutions, due to $O(n^4)$ complexity. 
These modules not only cause a significant increase in peak GPU memory, but their inference time also scales poorly with the increase of resolution. 
This indicates that the matching refinement module can become a bottleneck in terms of both memory and speed, which severely constrains the method's viability for practical applications. }

\section{Discussion of Challenges}
\begin{figure*}[!htbp]
\centering
% Add column subtitles
\makebox[0.24\textwidth]{Extreme Intra-class Variation}
\makebox[0.24\textwidth]{Left-right Confusion}
\makebox[0.24\textwidth]{Cross-Category Confusion}
\makebox[0.24\textwidth]{Incomplete View}
\\[5pt]
\includegraphics[width=0.24\textwidth]{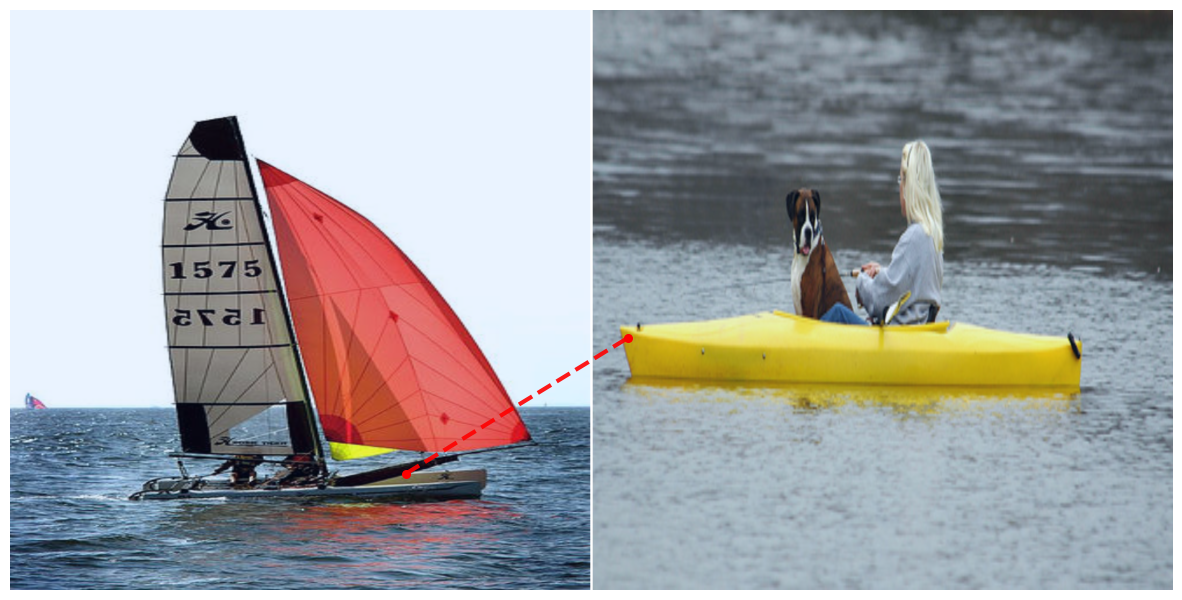}
\includegraphics[width=0.24\textwidth]{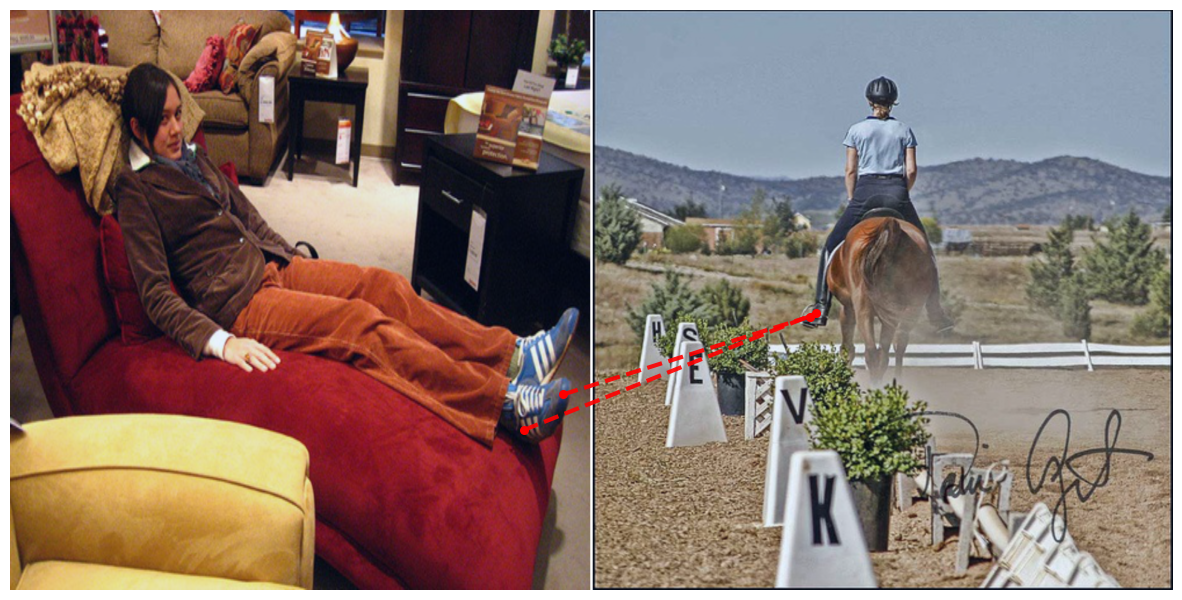}
\includegraphics[width=0.24\textwidth]{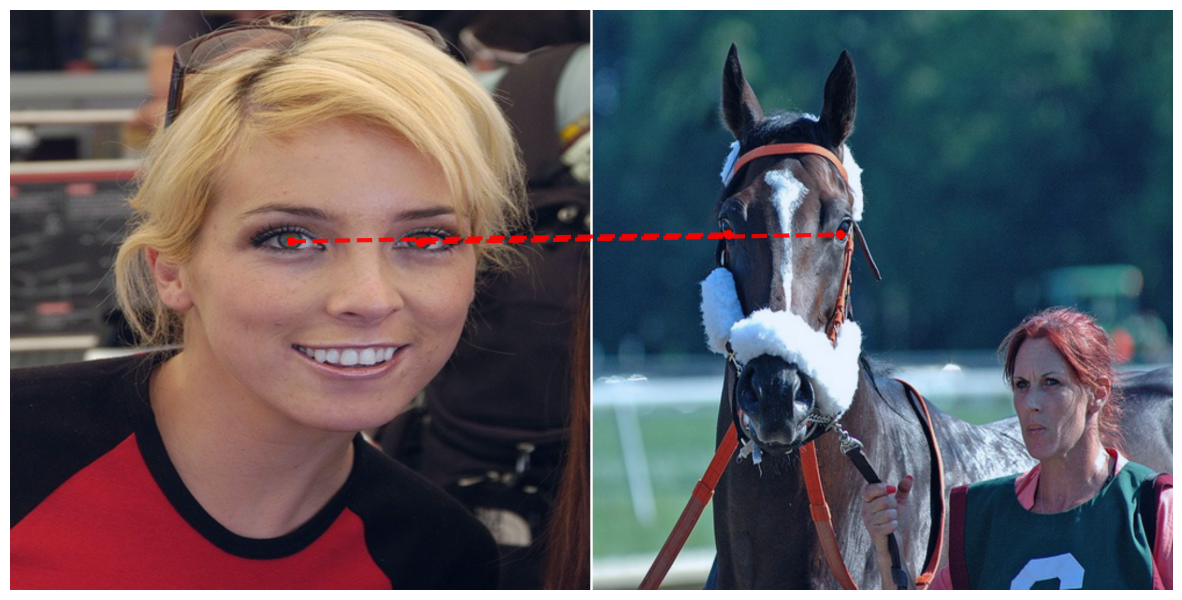}
\includegraphics[width=0.24\textwidth]{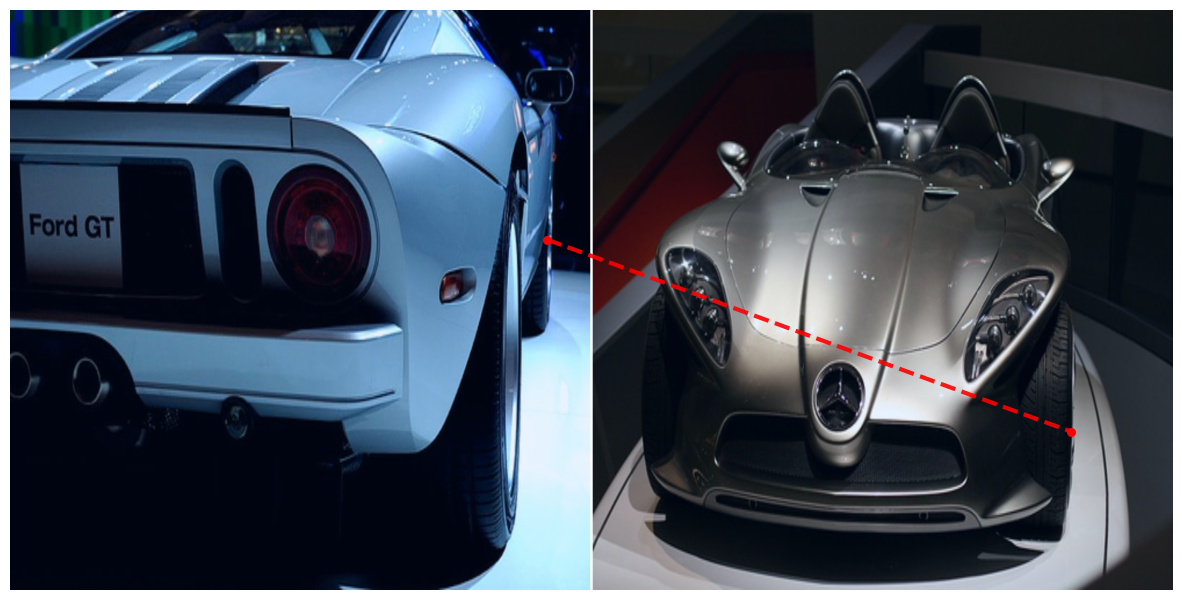}
\\
\includegraphics[width=0.24\textwidth]{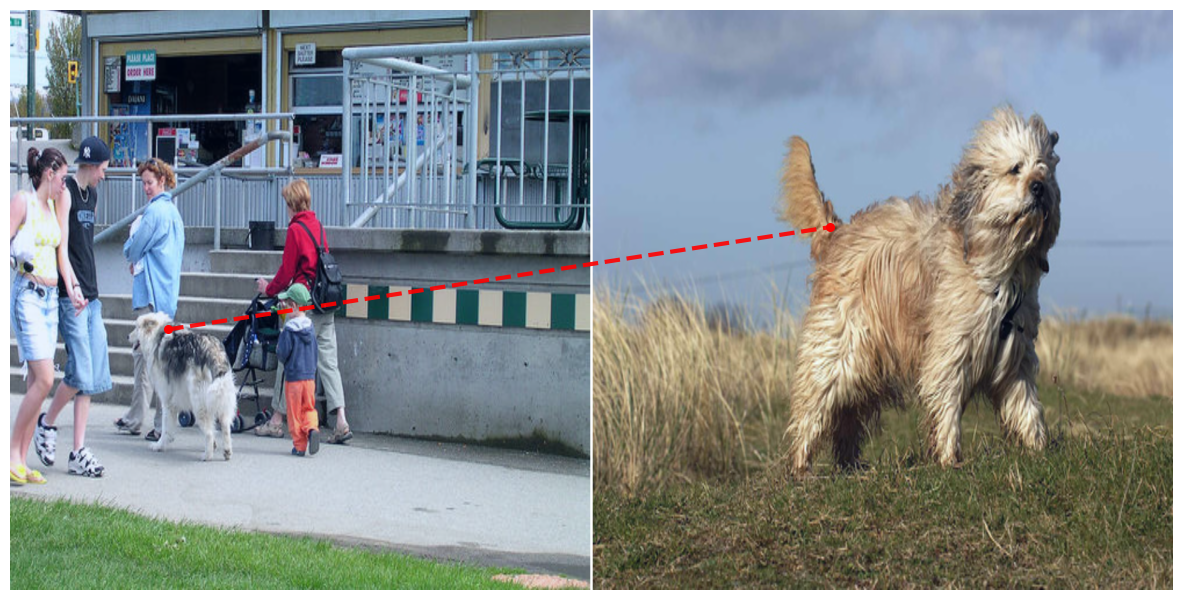}
\includegraphics[width=0.24\textwidth]{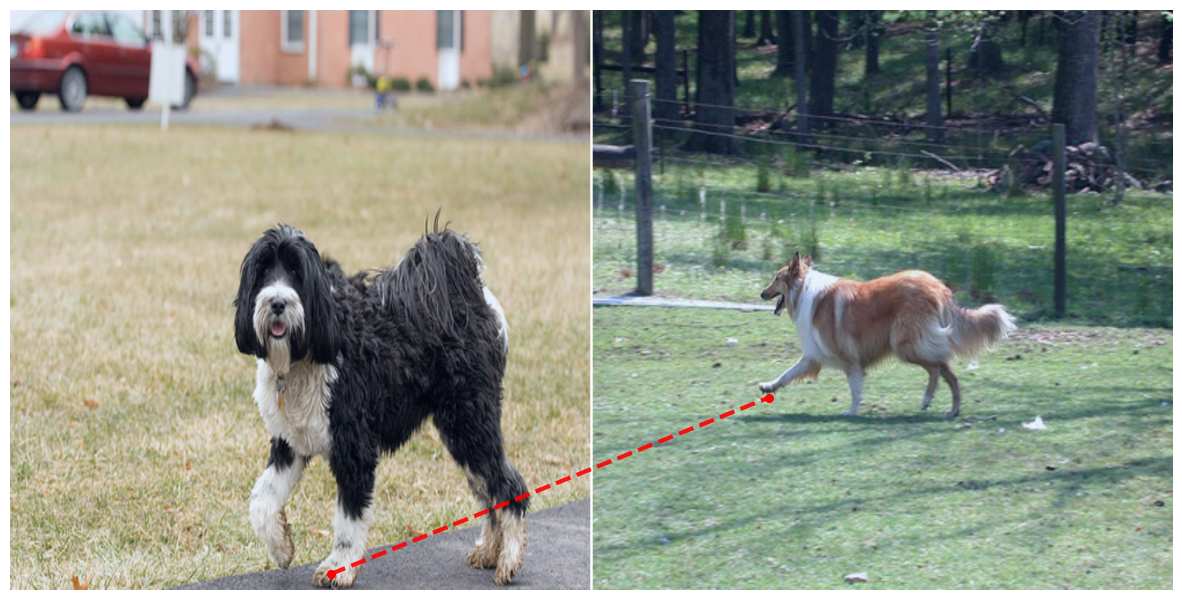}
\includegraphics[width=0.24\textwidth]{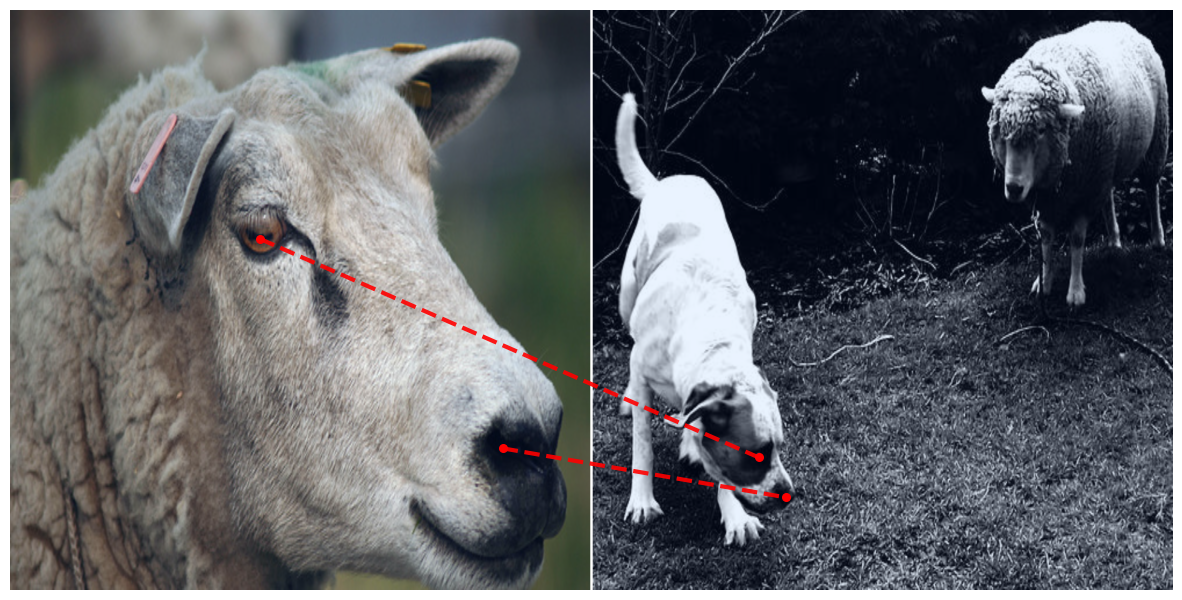}
\includegraphics[width=0.24\textwidth]{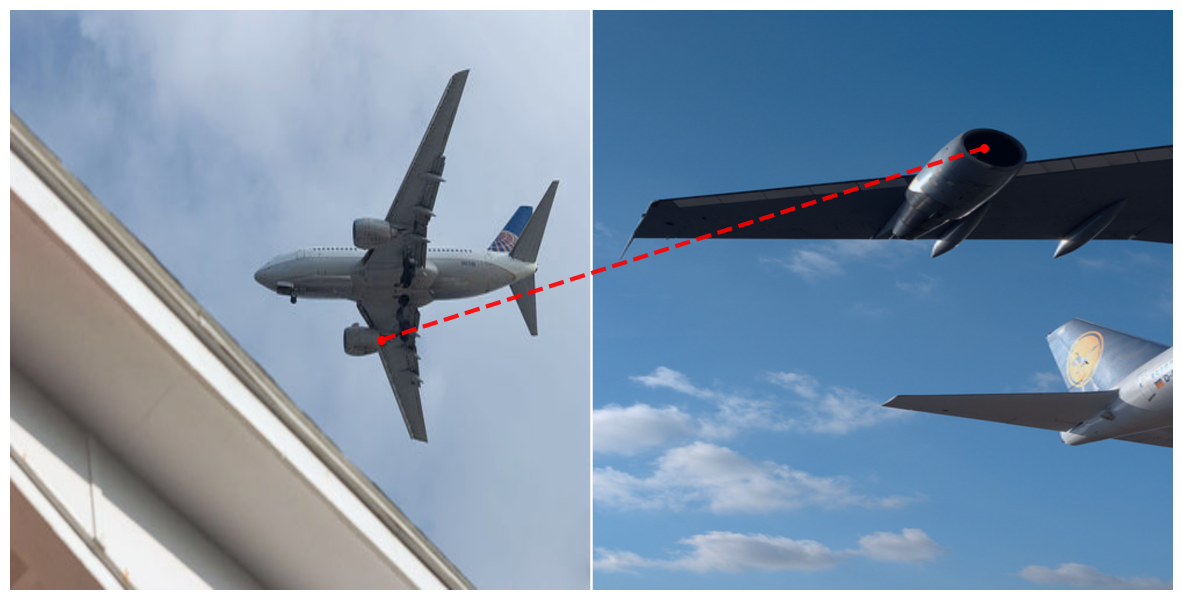}
% \\
% \includegraphics[width=0.24\textwidth]{fig/failure/31.png}
% \includegraphics[width=0.24\textwidth]{fig/failure/32.png}
% \includegraphics[width=0.24\textwidth]{fig/failure/33.png}
% \includegraphics[width=0.24\textwidth]{fig/failure/34.png}
% \\
% \includegraphics[width=0.24\textwidth]{fig/failure/41.png}
% \includegraphics[width=0.24\textwidth]{fig/failure/42.png}
% \includegraphics[width=0.24\textwidth]{fig/failure/43.png}
% \includegraphics[width=0.24\textwidth]{fig/failure/44.png}
\caption{Visualization of failure cases for the proposed method. The examples highlight challenges in handling (a) extreme intra-class variation, (b) left-right confusion, (c) cross-category confusion, and (d) incomplete view caused by large viewpoint changes and truncation.}
\label{fig:failure}
\end{figure*}

\rev{In this section, we discuss several challenging scenarios shared by our and existing methods. We identify four primary challenging cases and illustrate them in \Cref{fig:failure}.}

\rev{
%  extreme intra-class
First, we found that current methods still struggle with extreme intra-class variation. 
When faced with significant appearance differences within the same category (e.g., boats or animals), these methods have difficulty in identifying their semantically corresponding parts. 
This represents a widespread challenge in the semantic correspondence task.
A possible solution is to augment the training data with more diverse examples. 
Recent work by Chong \cite{SCAC} shows that even state-of-the-art methods remain vulnerable to common image corruptions, highlighting a pressing need for more diverse training data to improve model robustness.
Alternatively, model training could focus on learning more invariant features. Promising directions include disentangling style from content to match based on stable geometry rather than volatile appearance, or mapping instances to a shared canonical space where correspondence is established in an abstract domain immune to diverse 2D appearances.
}

\rev{
% left-right confusion
Another challenge arises from left-right confusion, where many methods underperform in distinguishing between bilaterally symmetric parts that share similar color and texture, such as a person's feet or an animal's legs.
% This limitation is not unique to our approach and is also observed in recent work such as GeoAware-SC~\cite{GeoAware-SC}. 
This issue stems from a broader lack of spatial and structural awareness, which prevents the model from understanding an object's composition beyond local pixel patterns. 
To mitigate this, one could incorporate stronger positional encodings or introduce explicit structural modeling to help the model understand object composition. 
A straightforward implementation is to add an orientation-aware module that provides the model with spatial priors to distinguish these symmetric parts.
}

\rev{
% Cross-Category Confusion
The third challenge is category confusion, where the model incorrectly matches objects from different classes (a human vs. a horse). This is likely because the model has learned generic patterns based on ambiguous local features, rather than the key discriminative features required to differentiate between species.
This category confusion can likely be attributed to an objective mismatch between DINOv2/v3's training and the semantic correspondence task. DINOv2/v3's training lacks explicit class supervision to ensure that representations of different classes are well-separated. 
Consequently, visually similar yet semantically distinct classes may be mapped to proximate regions in the feature space, causing the model to produce incorrect cross-category matches. 
This limitation may be addressed at both the fine-tuning and pre-training stages. 
For example, we may improve the fine-tuning by introducing an auxiliary classification loss. In this way, the model is jointly trained on both correspondence and image classification, compelling it to learn more class-aware features. 
We may also improve pre-training of the feature backbone by leveraging existing labeled datasets to explicitly encourage representations of objects within the same super-category to be closer than those from different categories. 
Ultimately, both strategies aim to shape a more semantically structured feature space where inter-class distances are maximized, thereby reducing category confusion and enhancing the reliability of correspondence.}

\rev{
% Incomplete object
Finally, we observe that current methods are vulnerable to an incomplete view caused by large changes in viewpoint and scale. 
We hypothesize this is because the model, being inherently 2D, lacks an intrinsic understanding of the 3D world. Consequently, it does not comprehend that different 2D images can be projections of the same 3D object, and it struggles to handle related phenomena such as occlusion and truncation.
A promising direction to address this is to explicitly incorporate 3D geometric information. 
Several attempts have been made in this direction.
% SemAlign3D
For instance, SemAlign3D \cite{SemAlign3D} has shown that augmenting 2D features with class-level 3D object representations—built from monocular depth and aligned at inference—can enhance robustness under extreme viewpoint changes.
% Jamais Vu
Mariotti~\etal~\cite{mariotti2025jamais} demonstrate that lifting sparse 2D keypoints into a learned category-specific 3D canonical manifold during training generates dense correspondences that generalize better to unseen keypoints.
% Bridging Viewpoint Gaps
Alternatively, robustness can be enhanced by strengthening geometric reasoning directly in the 2D domain.
Qian \etal~\cite{qian2025bridging} demonstrate that combining geometric reasoning with semantic refinement significantly boosts robustness under extreme viewpoint and appearance variations, achieving up to 19.6\% gains in zero-shot settings.
% Future
Future work could explore incorporating lightweight 3D information or leveraging multi-view learning to enhance the model's understanding of spatial relationships.
}
\end{appendices}

\vfill
\end{document}